\documentclass[preprints,article,submit,pdftex,moreauthors]{Definitions/mdpi}

\firstpage{1}
\pubvolume{1}
\issuenum{1}
\articlenumber{0}
\pubyear{2023}
\copyrightyear{2023}
\datereceived{ }
\daterevised{ } % Comment out if no revised date
\dateaccepted{ }
\datepublished{ }
\hreflink{https://doi.org/} % If needed use \linebreak
\pdfoutput=1 % Uncommented for upload to arXiv.org

\usepackage{graphicx}
\usepackage{subcaption}
\usepackage{mathtools}
\usepackage{makecell}

\Title{Deep Convolutional Neural Network for Plume Rise Measurements in Industrial Environments}

\TitleCitation{Deep Convolutional Neural Network for Plume Rise Measurements in Industrial Environments}

\Author{Mohammad Koushafar, Gunho Sohn * and Mark Gordon}

\AuthorNames{Mohammad Koushafar, Gunho Sohn and Mark Gordon}

\AuthorCitation{Koushafar, M.; Sohn, G.; Gordon, M.}
\address[1]{%
Department of Earth and Space Science and Engineering, York University, 4700 Keele Street, Toronto, ON M3J 1P3, Canada}

\corres{Correspondence: gsohn@yorku.ca}

\abstract{Estimating Plume Cloud (PC) height is essential for various applications, such as global climate models. Smokestack Plume Rise (PR) is the constant height at which the PC is carried downwind as its momentum dissipates and the PC and the ambient temperatures equalize. Although different parameterizations are used in most air-quality models to predict PR, they have yet to be verified thoroughly. This paper proposes a low-cost measurement technology to monitor smokestack PCs and make long-term, real-time measurements of PR. For this purpose, a two-stage method is developed based on Deep Convolutional Neural Networks (DCNNs). In the first stage, an improved Mask R-CNN, called Deep Plume Rise Network (DPRNet), is applied to recognize the PC. Here, image processing analyses and least squares, respectively, are used to detect PC boundaries and fit an asymptotic model into the boundaries centerline. The y-component coordinate of this model's critical point is considered PR. In the second stage, a geometric transformation phase converts image measurements into real-life ones. A wide range of images with different atmospheric conditions, including day, night, and cloudy/foggy, have been selected for the DPRNet training algorithm. Obtained results show that the proposed method outperforms widely-used networks in smoke border detection and recognition.}


\begin{document}

%%%%%%%%%%%%%%%%%%%%%%%%%%%%%%%%%%%%%%%%%%
%\setcounter{section}{-1} %% Remove this when starting to work on the template.
\section{Introduction} \label{section.Introduction}
Smokestack Plume Cloud (PC) rises due to momentum and buoyancy. Finally, the PC dissipates and is carried downwind at a constant height. This height is called plume rise height or Plume Rise (PR). PR calculation is not straightforward, and it is a substantial problem in predicting the dispersion of different harmful effluents into the air \cite{briggs1982plume}. PR contributes to 1) the distance pollutants carried downwind, 2) their concentration at the surface, where they are deposited in the green environment or inhaled by people, and 3) the amounts of greenhouse gases mixed into the upper troposphere. Therefore, accurate measurement of the PR is of concern for research and operational applications such as air-quality transport models, local environment assessment cases and global climate models \cite{ashrafi2017direct}.

The parameterizations of PR prediction were developed in the 1960s by Briggs \cite{briggs1969plume, briggs1975plume}. Dimensional analysis was used to estimate the PR based on smokestack parameters and meteorological measurements in different atmospheric conditions. Early observations of PR were used to test and rectify the parameterizations developed using dimensional analysis \cite{bieser2011vertical}. Wind tunnel studies and field observations using technologies including film photography, theodolites, and cloud-height searchlights \cite{bringfelt1968plume} were several calibration techniques utilized in this domain. There are also three-dimensional air-quality models, using parameterizations equations including GEM-MACH \cite{makar2015feedbacks}, CAMx \cite{emery2010implementation}, and CMAQ \cite{byun1999science}.

Some studies tested the parameterizations of PR prediction in the 1970s and 1980s by comparing them to actual observations and demonstrated that the Briggs equations overestimate the PR \cite{rittmann1982application, england1976measurement, hamilton1967paper, moore1974comparison}. In 1993, an aircraft-based measurement was done to measure $\mathrm{SO_{2}}$ emissions of a power plant which indicated an overestimation of about 400 m \cite{sharf1993plume}. Although these earlier studies showed some degree of overestimation, in 2002, Webster et al. \cite{webster2002validation} performed surface measurements and concluded that the Briggs parameterizations tend to underestimate PR. In 2013, as part of the Canada-Alberta Joint Oil Sands Monitoring (JOSM) Plan, an aerial measurement study was done in the Athabasca oil sands region of northern Alberta to study dispersion, and chemical processing of emitted pollutants \cite{gordon2015determining, gordon2018comparison, akingunola2018chemical}. The project consisted of 84 flight hours of an instrumented Convair aircraft over 21 flights designed to measure pollutants emissions, study the transformation of chemicals downwind of the industry, and verify satellite measurements of pollutants and greenhouse gases in the region. Using aircraft-based measurements and reported smokestack parameters and meteorological data, it was demonstrated that the Briggs equations significantly underestimate PR at this location.

Given the results of \cite{gordon2015determining, gordon2018comparison, akingunola2018chemical} and the gap of more than 30 years since the Briggs equations were developed and previously tested, there is a need for further testing and possible modification of the Briggs equations based on modern observation techniques. In recent decades, there have been many significant advancements in environmental monitoring activities over industrial regions for safety and pollution prevention \cite{isikdogan2017surface, isikdogan2017rivamap, gu2018recurrent, gu2018highly}. Moreover, several smoke border detection and recognition models have been introduced recently using digital image analysis, such as wavelet and support vector machines \cite{gubbi2009smoke}, LBP and LBPV pyramids \cite{yuan2011video}, multi-scale partitions with AdaBoost \cite{yuan2012double}, and high-order local ternary patterns \cite{yuan2016high} which are well-performed and impressive. These improvements have led to the development of instrumentation which can be deployed near any smokestack to give information on pollutant dispersion and potential exposure to people downwind. This information will be based on actual real-time observation, i.e. digital images, as opposed to potentially erroneous and decades-old parameterizations. Due to the similarity of our work to smoke recognition on the one hand, and on the other, the unavailability of plume cloud recognition research, smoke recognition studies will be reviewed in the following.

To find the smoke within an image or a video frame, either a rough location of smoke is identified using bounding boxes called smoke border detection \cite{yuan2012double, yuan2015real}, or pixels are identified and classified in detail, called smoke recognition \cite{yuan2019deep, khan2021deepsmoke}. Due to the translucent edges of smoke clouds, the recognition task needs far more accuracy than border detection. Traditional smoke recognition methods utilize manual features, which lead to low accuracy recognition results due to a large variety of smoke appearances. These low-level features consist of motion characteristic analysis of the smoke \cite{yu2013real, shi2015study}, smoke colour \cite{garcia2018segmentation, yuan2019learning, garg2018smoke}, and smoke shape \cite{filonenko2017fast}. In another research, \cite{wu2008smoke, zen2013dangerous} took advantage of the Gaussian Mixture Model (GMM) to detect the motion region of the smoke and \cite{wang2019smoke} combined rough set and region growing methods as a smoke recognition algorithm which seems to be a time-consuming algorithm due to the computational burden of the region growing process. Since using colour information is less effective due to the similarity of smoke colour to its surrounding environment, the combination of motion and colour characteristics is considered for smoke recognition \cite{jian2018smoke, yuan2018smoke, zhao2019smoke}. Some algorithms utilize infrared images, and video frames in their experiments \cite{ajith2019unsupervised}, which are not easily accessible and can increase the project's costs. Moreover, using digital images makes the algorithm more flexible as it can be used with more hardware. On the other hand, some smokes are too close to the background temperature to be captured by the near red-channel wavelength. A higher-order dynamical system was introduced in 2017, which used particle swarm optimization for smoke pattern analysis \cite{dimitropoulos2016higher}. However, this approach had a low border detection rate and high computational complexity.

In recent years, deep learning-based methods, especially Convolutional Neural Network (CNN) based methods, have led to significant results in semantic segmentation \cite{wang2017gated, li2017fully, wang2017video, pham2022new} and object recognition \cite{ronneberger2015u, caelles2017one, hou2017deeply, dai2016instance, shi2022automatic}. Similarly, these methods are widely used in smoke border detection, and recognition \cite{muhammad2019edge, yin2017deep, hu2018real} with different architectures such as three-layer CNN \cite{liu2019simple}, generative adversarial network (GAN) \cite{jia2019automatic} and two-path Fully Convolutional Network (FCN) \cite{yuan2019deep}. Recently, a count prior embedding method was proposed for smoke recognition to extract information about the counts of different pixels (smoke and non-smoke) \cite{yuan2022cubic}. Experimental results showed an improvement in the recognition performance of these studies. However, the high computational complexities of these huge models are an obstacle to their use in PR real-time observations.

We have proposed a novel method using DCNN algorithms to measure PR. Our approach comprises two parts, 1) PC border detection and recognition based on an improved Mask R-CNN \cite{he2017mask}, and 2) geometric transformation \cite{luhmann2006close} to migrate from image scale measurements to real-life. 

This method accurately recognizes the PC and measures PR in real-time. Here, we reinforce the bounding box loss function in Region Proposal Network (RPN) \cite{zheng2020distance, hwang2022automatic, girshick2015fast} through engaging a new regularization to the loss function. This regularizer restricts the search domain of RPN to the smokestack exit. In other words, it minimizes the distance between the proposed bounding boxes and the desired smokestack exit, which is called smokestack exit loss ($L_{sse}$). The proposed method is also computationally economical because it generates only a limited number of anchor boxes swarmed across the desired smokestack exit. Consequently, the main contributions of this paper can be summarized as follows:
\begin{itemize}
  \item Proposing "DPRNet" (a deep learning framework for PR measurements) by incorporating PC recognition and image processing-based measurements. We have provided a relatively low-cost and reproducible algorithm to accurately recognize plume clouds using an automated digital image capturing software adaptable to utilize RGB widescreen images.
  %\item We propose a plume rise vision-based measurement technique that will improve plume-rise parameterizations, increasing air-quality models' accuracy. Furthermore, the industry might use this real-time visual system to track and survey the released pollutants online and obtain information to update related criteria, such as live air-quality health index (AQHI) updates.
  \item A pixel-level recognition dataset, Deep Plume Rise Dataset (DPRD),  containing 2500 fine annotations, is presented. As is expected, the DPRD dataset includes one class, namely PC. Widely-used DCNN-based smoke recognition methods are employed to evaluate our dataset. Furthermore, this newly generated dataset was used for PR measurements.
  %\item Unlike most recent studies in the smoke recognition field, we perform both the plume cloud's recognition and border detection tasks based on an improved state-of-the-art CNN model with fewer computational complexities. This has led to the building a comprehensive operational system to observe better the plume clouds behaviours within an industrial area with instant plume cloud measurement announcements. A developed system would directly interest industries looking to confirm or establish emissions reporting and environmental assessments.
\end{itemize}

This paper is organized as follows—section \ref{section.Theoretical background} briefly explains the theoretical information used in our proposed method. In section \ref{section.Methodology}, we describe our proposed framework for the PR measurement of a desired smokestack. Then, section \ref{section.Experimental results and discussion} presents our dataset collection procedure, under-study site, experimental results of the proposed method and evaluation results using different metrics, and calculations of the PR. Finally, this research's conclusions, findings, and future studies are described in section \ref{section.Conclusion}.

\section{Theoretical background}\label{section.Theoretical background}
\subsection{Briggs PR prediction} \label{subsection.Plume rise theory}
The PR calculation is an ill-posed problem used to predict the dispersion of harmful effluents in atmospheric science \cite{briggs1982plume}. PR is affected by two phenomena, buoyancy and momentum. Typically, the PCs are buoyant, which means they are hotter than the ambient air. Therefore, they rise since they are less dense than the surrounding air. Also, the PCs have a vertical velocity and momentum when they exit the smokestack, which again causes them to rise. PCs can also fall due to the gravitational force when cold and dense and when some surrounding obstacles cause them to move downwind \cite{de2013air}. In 1975 Briggs proposed an equation for the maximum distance of the PR, which was practically suitable for calculating PR. Considering both momentum and buoyancy in calculations, PR ($\Delta{z}$) in horizontal distance \emph{x} from the smokestack exit can be obtained as \cite{briggs1975plume},

\begin{linenomath}\label{Eq.xff5}
    \begin{equation}
    \Delta{z}= \Biggl(\frac{3\,F_{m}x}{0.6^{2}\,\bar{u}^{2}}+\frac{3\,F_{b}x^2}{{2}\times0.6^2\,\bar{u}^{3}}\Biggl)^{1/3}
    \end{equation}
\end{linenomath}

\noindent where $0.6$ is an entrainment rate (the mean rate of increase of PC in the wind direction) and $\bar{u}$ is the mean horizontal wind speed. Also, Momentum flux parameter ($F_{m}$) and Buoyancy flux parameter ($F_{b}$) are defined as below,

\begin{linenomath}\label{Eq.MFP}
    \begin{equation}
    F_{m} = [\frac{\bar{\rho_{s}}}{\bar{\rho}}] \, r^{2}_{s}\bar{w_{s}}^{2}
    \end{equation}
\end{linenomath}

\begin{linenomath}\label{Eq.BFP}
    \begin{equation}
    F_{b} = [1-\frac{\bar{\rho_{s}}}{\bar{\rho}}] \, gr^{2}_{s}\bar{w_{s}}^{2}
    \end{equation}
\end{linenomath}

\noindent where $\bar{\rho_{s}}$ is smokestack gas density, $\bar{\rho}$ is atmospheric air density, $r_{s}$ is smokestack radius, $\bar{w_{s}}$ is vertical velocity of the smokestack gas, and $g$ is acceleration due to gravity.

It should be noted that if we evaluate the wind speed at the local PC height and not the source height, the calculations should be operated iteratively \cite{cimorelli2005aermod, turner2007atmospheric}. PC buoyancy and horizontal momentum movements and consequently PR are strongly affected by the wind \cite{briggs1982plume}. Moreover, in stable conditions with low turbulence, PR is unaffected by wind speed fluctuations, making the measurements difficult. However, significant PR variations in unstable conditions have been witnessed at a fixed distance downwind.

\subsection{CNN and convolutional layer} \label{subsection.Convolutional Neural Network and Convolutional Layer}
CNNs are special types of neural networks suitable for processing grid data, such as image data with a two-dimensional or three-dimensional mesh structure of pixels. The name given to a CNN is derived from the convolutional layers used in it. Each convolutional layer contains several kernels and biases that are locally applied to the input and produces a feature map or an activation map according to the number of filters. If this convolutional layer is applied to the input image in a two-dimensional manner, its $j^{th}$ feature map $O_{j}$, which is obtained by applying the $j^{th}$ kernel, is calculated at the position $(x,y)$ as \cite{ji20123d},

\begin{linenomath}\label{Eq.Kerrnel}
    \begin{equation}
    O_{j}^{xy} = B_{j} + \sum_{k} \sum_{m=0}^{M-1}\sum_{n=0}^{N-1} w_{j}^{mn} z_{k}^{(x+m)(y+n)}
    \end{equation}
\end{linenomath}

\noindent where $k$ moves along the depth dimension of input $z \in \mathbb{R}^{p \times q \times r}$ and $w_{j}^{mn}$ is the two-dimensional kernel weight $W \in \mathbb{R}^{M \times N}$ at position $(m,n)$. $B_{j}$ is the bias matrix.

\subsection{Mask R-CNN}\label{subsection.Mask R-CNN}
Mask R-CNN is a region-based CNN family member, proposed in \cite{he2017mask} and is used widely in different identification tasks, including COCO dataset recognition. Firstly, RCNN was presented in \cite{albawi2017understanding}, where computer vision techniques generated region proposals. Then in \cite{girshick2015fast}, Fast RCNN was introduced with a CNN before region proposals to reduce running time. In 2017, Faster RCNN continued this evolution and offered the RPN to propose regions of interest (ROIs) \cite{chen2017implementation}. Finally, Mask R-CNN, as an extension of Faster RCNN, added a CNN for pixel-level recognition of the border-detected objects. Mask R-CNN is a relatively simple model and easy to generalize to other similar tasks \cite{he2017mask}. Therefore, Mask R-CNN can create pixel-level masks for the objects besides object localization and classification tasks.

\subsubsection{RPN}\label{subsubsection.Region proposal network}
\begin{figure}[h!]
\includegraphics[width=11 cm]{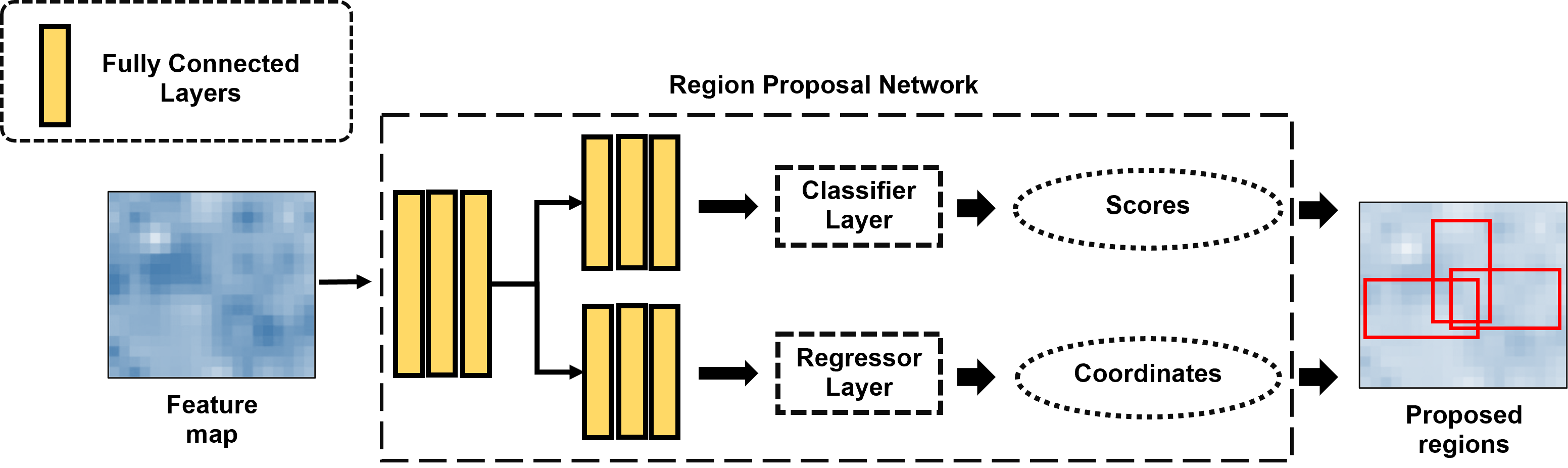}
\caption{Region proposal network.\label{fig.Region proposal network concept.}}
\end{figure}

An RPN is a deep FCN that proposes regions and is crucial to the Mask R-CNN. RPN helps selectively focus on valuable aspects within the input images. This network takes an image and gives a set of region proposals beside their scores for being an object. It slides a window over the convolutional feature map (backbone output) and maps it to a lower-dimension feature. The generated feature is then fed into two fully-connected layers to obtain the proposed regions class (object vs. non-object), and four corresponding coordinates \cite{chen2017implementation}. For each sliding window location, there are a maximum of \emph{k} possible proposals which are parameterized relative to \emph{k} reference boxes or anchors. RPN is trained end-to-end by back-propagation and stochastic gradient descent. Figure \ref{fig.Region proposal network concept.} depicts a scheme of RPN in which the proposed regions are generated as the module outputs.

\unskip

\subsubsection{Loss function}\label{subsubsection.Loss function}
Mask R-CNN loss function is a weighted summation of other losses related to different sections of this comprehensive model. As a definition based on \cite{he2017mask}, a multi-task loss function is proposed on each sampled ROI as,

\begin{linenomath}\label{Eq.Loss11}
    \begin{equation}
    L = L_{cls} + L_{reg} + L_{msk}
    \end{equation}
\end{linenomath}

\noindent where $L_{cls}$ recognizes the class type of each object, while $L_{reg}$ attempts to find the optimum anchor box for each object. Note that, in this study, we have one class, PC. $L_{msk}$ tries to recognize the optimum object's segment in each bounding box.

\section{Methodology}\label{section.Methodology}
\begin{figure}[h!]
\centering
\includegraphics[width=11 cm]{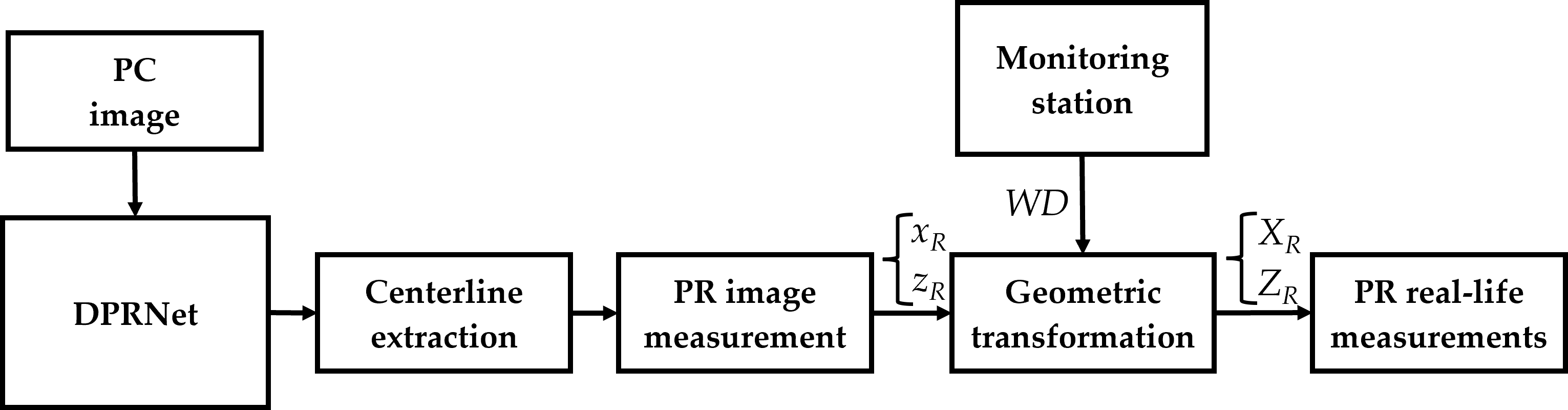}
\caption{PR measurement system framework. $x_{R}$ and $z_{R}$ are PR distance and PR in the image scale. Similarly, $X_{R}$ and $Z_{R}$ represent PR distance and PR, respectively, in real-life scale, and $WD$ shows the wind direction.\label{fig.Plume rise measurement system framework.}}
\end{figure}

The proposed method for PR measurement is represented in Figure \ref{fig.Plume rise measurement system framework.}. The images containing PC(s) are fed to DPRNet for PC border detection and recognition. PR is then measured on DPRNet's output based on estimating the critical point ($R$) \cite{ge2023unsupervised}. For this purpose, an asymptotic function \cite{ge2023unsupervised} is fitted into the PC centerline, extracted by image processing analysis (e.g. morphological operators) \cite{gonzales1987digital}. The measured image coordinates of $R$ are combined with the wind direction information to be processed by geometric transformation calculations. The main output of the system will reveal the PR as a physical height and the distance downwind at which the PR occurs.

\unskip

\subsection{Structure of DPRNet}\label{subsection.Structure of DeepRise}
\begin{figure}[th]
\centering
\includegraphics[width=11 cm]{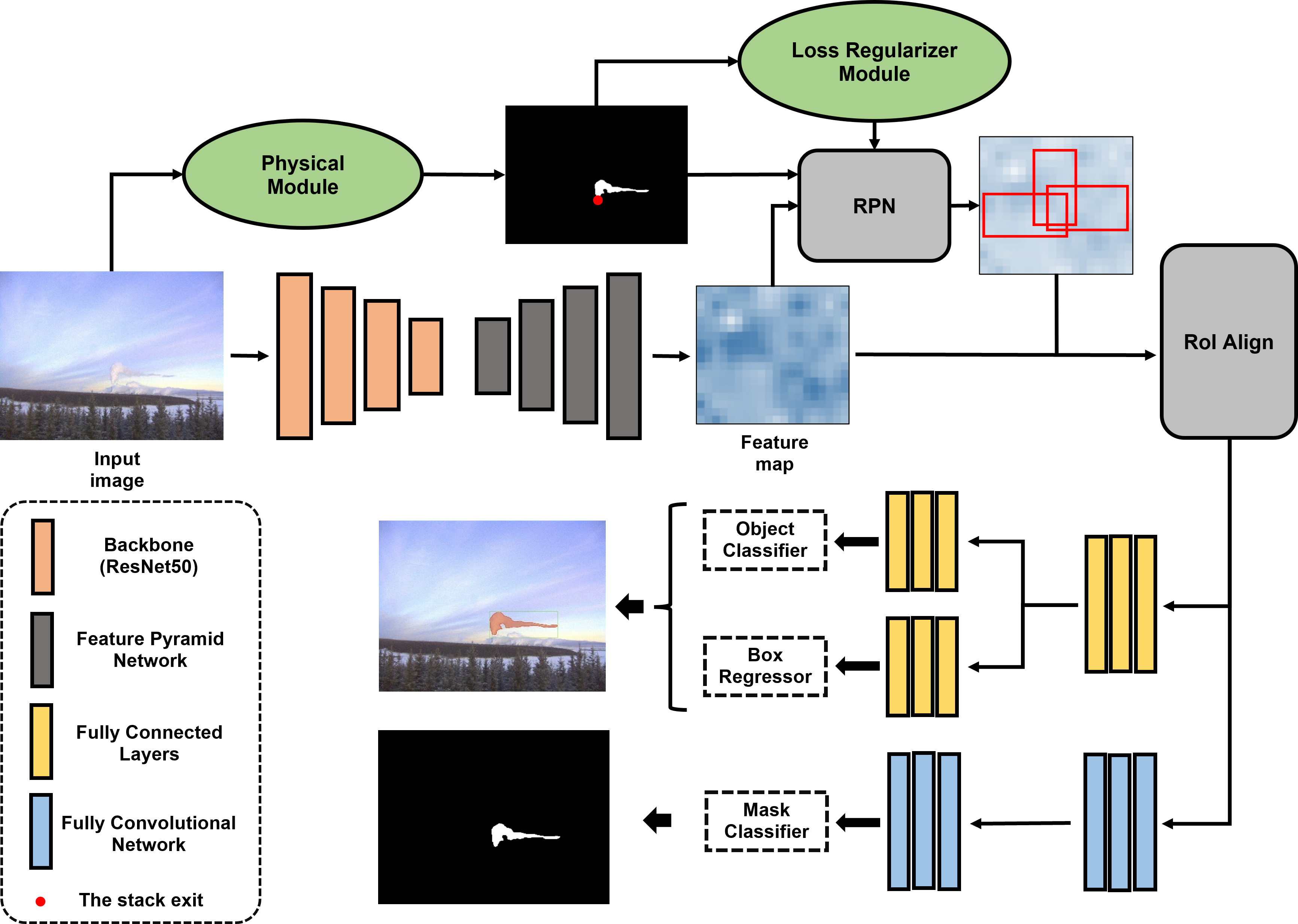}
\caption{DPRNet architecture.}
\label{DeepRise architecture.}
\end{figure}

This research aims to precisely recognize the PC of the desired smokestack from a wide range of image datasets captured from the study area. DPRNet is an adapted Mask R-CNN version with two novel smokestack PR measurement modules. These modules are 1) the physical module and 2) the loss regularizer module. These modules can improve RPN performance in locating the most probable proposal PCs. Mask R-CNN is the base of our proposed method, one of the widespread border detection and recognition methods. This robust framework can consider the irregular shapes of the PC, its translucent edges, and similar pixel values of the PC to its background \cite{he2017mask}. As seen from Figure \ref{DeepRise architecture.}, DPRNet is an application-oriented version of Mask R-CNN to which two new modules have been added.

In this architecture, ResNet50 \cite{he2016deep} is used as the backbone network to extract feature maps of the input images. Feature Pyramid Network (FPN) uses all these feature maps to generate multi-scale feature maps, which carry more helpful information than the regular feature pyramid. Then, RPN detects the PC by sliding a window over these feature maps to predict whether there is a PC and locate the existing PC by creating bounding boxes. Therefore, we have a set of PC proposals from RPN and the generated feature map by the backbone network. The ROI Align module works to scale the proposals to the feature map level and prevent misalignment by standardizing the aspect ratios of the proposals. Finally, these refined feature maps are sent to three different outputs. The first is a classification block that decides whether the ROI is the foreground (PC). The second one is a regression block which predicts the bounding boxes based on the provided ground truth. And the last block indicates a recognition mask for the detected PC using an FCN \cite{long2015fully}.

Two modules are added to Mask R-CNN to improve its efficiency and reduce the computational burden. The first module, a simple image processing one, approximates smokestack. The second module attempts to improve the loss related to $L_{reg}$ by adding a regularizer loss, elaborated in Section \ref{subsubsection.Loss function modification}. These modules are explained in detail in the following subsections.

\unskip

\subsubsection{Physical module}\label{subsubsection.Mask entering module}
Given either an estimated binary image (during inference time) or ground truth binary image (during training time), the smokestack exit can be detected by image processing techniques \cite{gonzales1987digital}. It stands to reason that the smokestack exit is the feasible region of the plume rise. As a result, proposed regions can be considered around this point (Figure \ref{DeepRise architecture.}). Thanks to this module, the method does not detect small PC pieces, sometimes seen in different parts of images other than the smokestack exit.

\subsubsection{Loss regularizer module}\label{subsubsection.Loss function modification}
Based on Section \ref{subsubsection.Loss function}, it is a crucial problem to set an efficient loss function, which can make the model stable as it can get. In this regard, a new regularizer is added to the loss function, which dictates the coordinates of the most attainable PC regions. Indeed, we try to minimize the distance of proposed bounding boxes by RPN and the smokestack exit. If a box with coordinates of ($x,y,w,h$) is defined here, the regression loss related to the smokestack exit can be defined as,
\begin{linenomath}\label{Eq.Loss896}
    \begin{equation}
    L_{sse} = R(u-{u^{*}}),
    \end{equation}
\end{linenomath}

\noindent in which,
\begin{linenomath}\label{Eq.Loss13}
    \begin{equation}
    \begin{aligned}
    u_{x} = \frac{x-x_{a}}{w_{a}},\;\;u_{y} = \frac{y-y_{a}}{h_{a}},\;\;u_{w} = \log{(\frac{w}{w_{a}})},\;\;u_{h} = \log{(\frac{h}{h_{a}})}, \\
    u^{*}_{x} = \frac{x^{*}-x_{a}}{w_{a}},\;\;u^{*}_{y} = \frac{y^{*}-y_{a}}{h_{a}},\;\;u^{*}_{w} = \log{(\frac{w^{*}}{w_{a}})},\;\;u^{*}_{h} = \log{(\frac{h^{*}}{h_{a}})}
    \end{aligned}
    \end{equation}
\end{linenomath}

\noindent where $u$ and $u^{*}$ represent the coordinates of our predicted and ground truth smokestack exit, and $R$ is the robust loss function. Note that variables with subscript $a$ and superscript $*$ represent the anchor coordinate and the ground truth coordinates, respectively, while the rest are defined as the predicted coordinates. The point $(x,y)$ indicates the position of the bounding box's top-left edge and the parameters $w$ and $h$ are, respectively, the width and height of the bounding box.

Unlike the Mask R-CNN model, in DPRNet, the loss regularizer module ($L_{sse}$) minimizes the recognition task errors of a specific PC, which copes with the main problems of this model, such as missing the desired smokestack exit and multi-box proposal for a single PC. $L_{sse}$ helps us avoid spanning the whole image pixels, causing high training time and computational complexities.

\subsection{Geometric transformation}\label{subsection.Triangulation}
To convert the image measurement results to real-world ones, we need to perform some calculations to transform the PR ($\Delta{z}$) and the PR distance ($X_{max}$) on an image to real-life measurement using wind direction.

The PR distance can be defined as the horizontal distance between the smokestack exit and the point $R$, discussed in Section \ref{section.Methodology}. Figure \ref{fig.Definitions of plume rise, plume rise distance, and the point $R$ on a sample captured image.} shows PR and PR distance definitions on a sample PC image.

\begin{figure}[th]
\begin{subfigure}{.5\textwidth}
  \centering
  \includegraphics[width=6 cm]{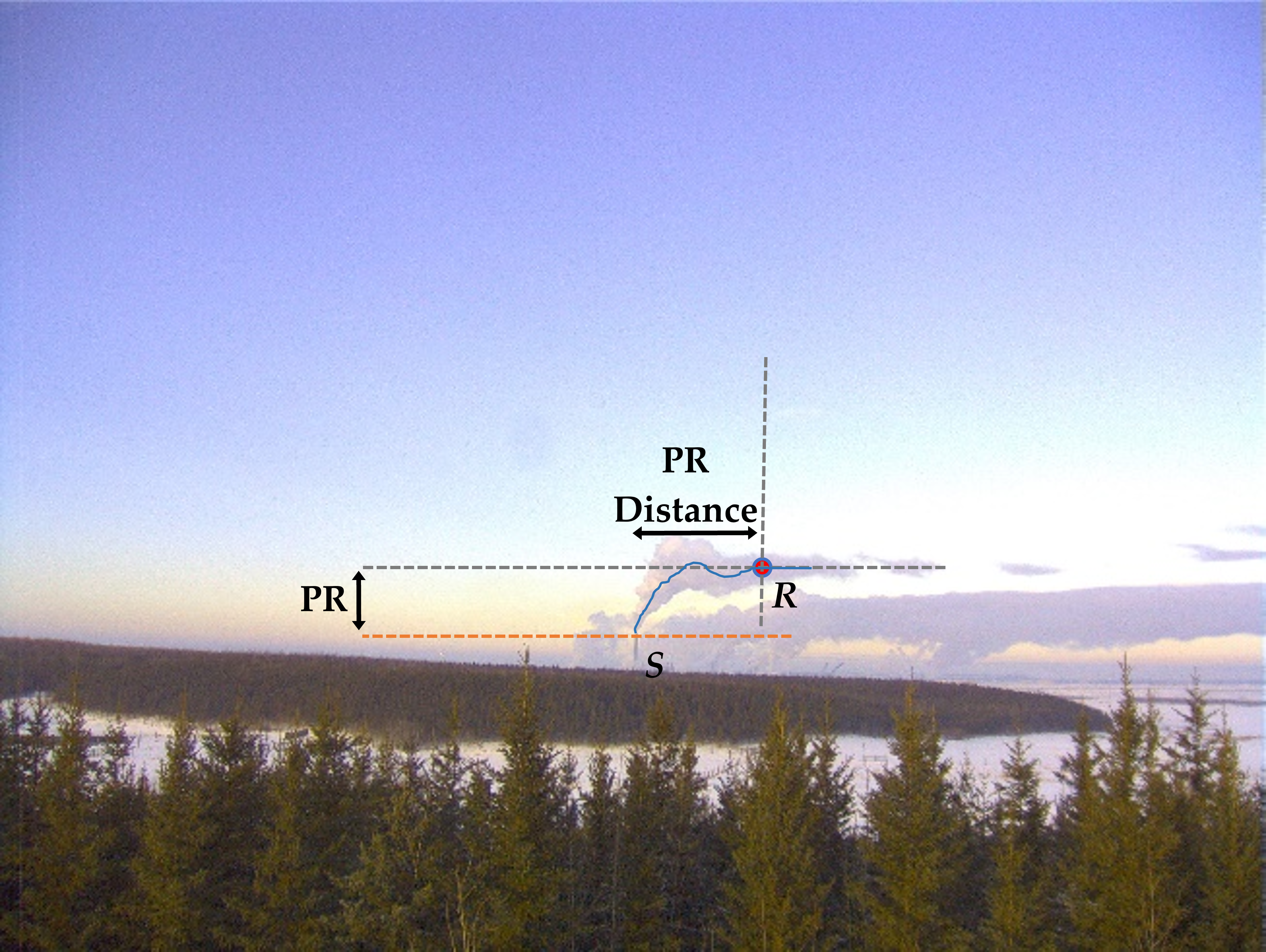}
  \caption{Camera view of the PC.}
  \label{subfig.Camera view scheme.}
\end{subfigure}
\begin{subfigure}{.5\textwidth}
  \centering
  \includegraphics[width=6 cm]{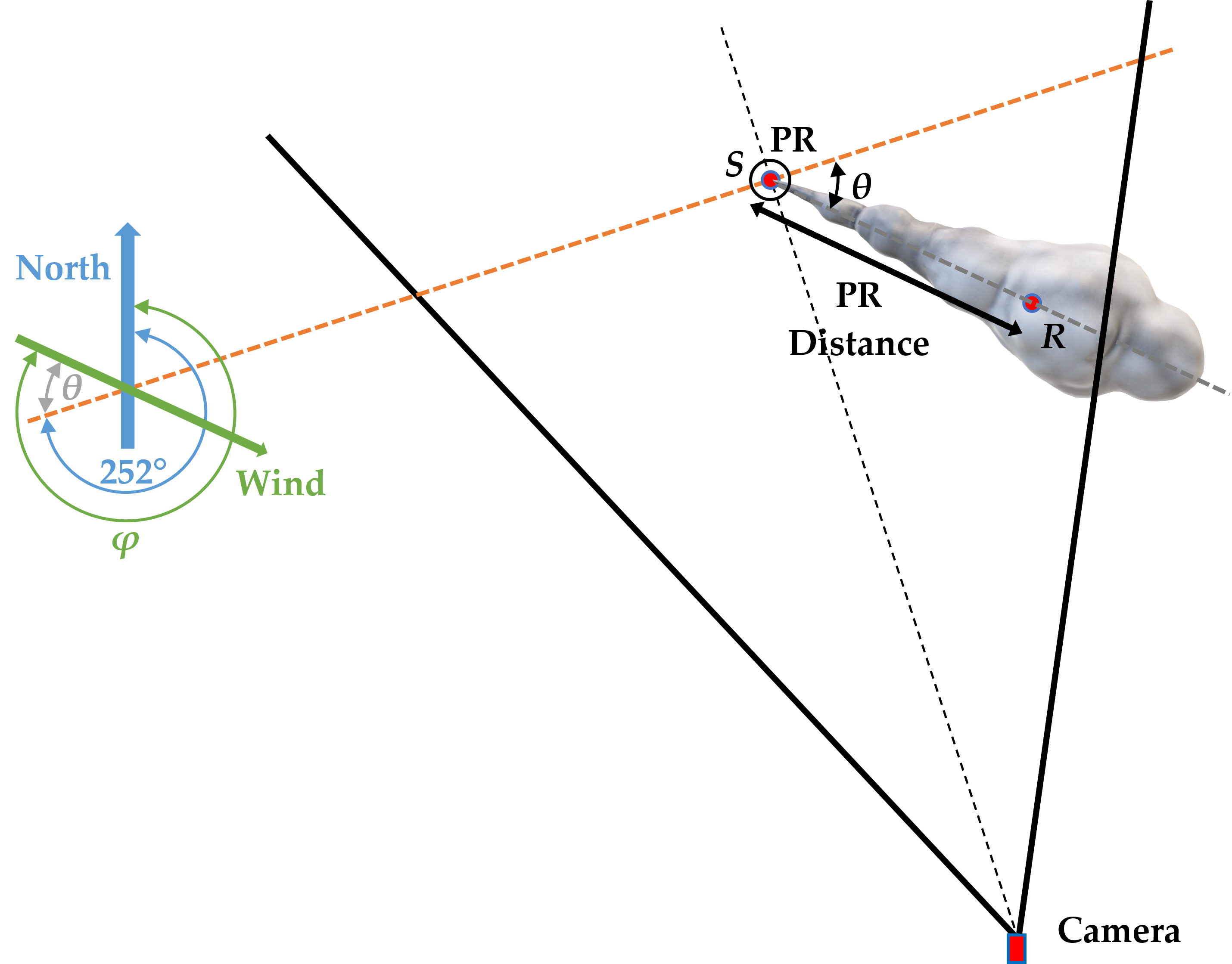}
  \caption{Top view of the PC.}
  \label{subfig.Top view scheme.}
\end{subfigure}
\caption{PR, PR distance, and the point $R$ on a sample image. $\theta$ represents the PC deviation due to the wind, and $S$ shows the smokestack position.}
\label{fig.Definitions of plume rise, plume rise distance, and the point $R$ on a sample captured image.}
\end{figure}
\unskip

\begin{figure}[th]
\begin{center}
  \centering
  \includegraphics[width=11 cm]{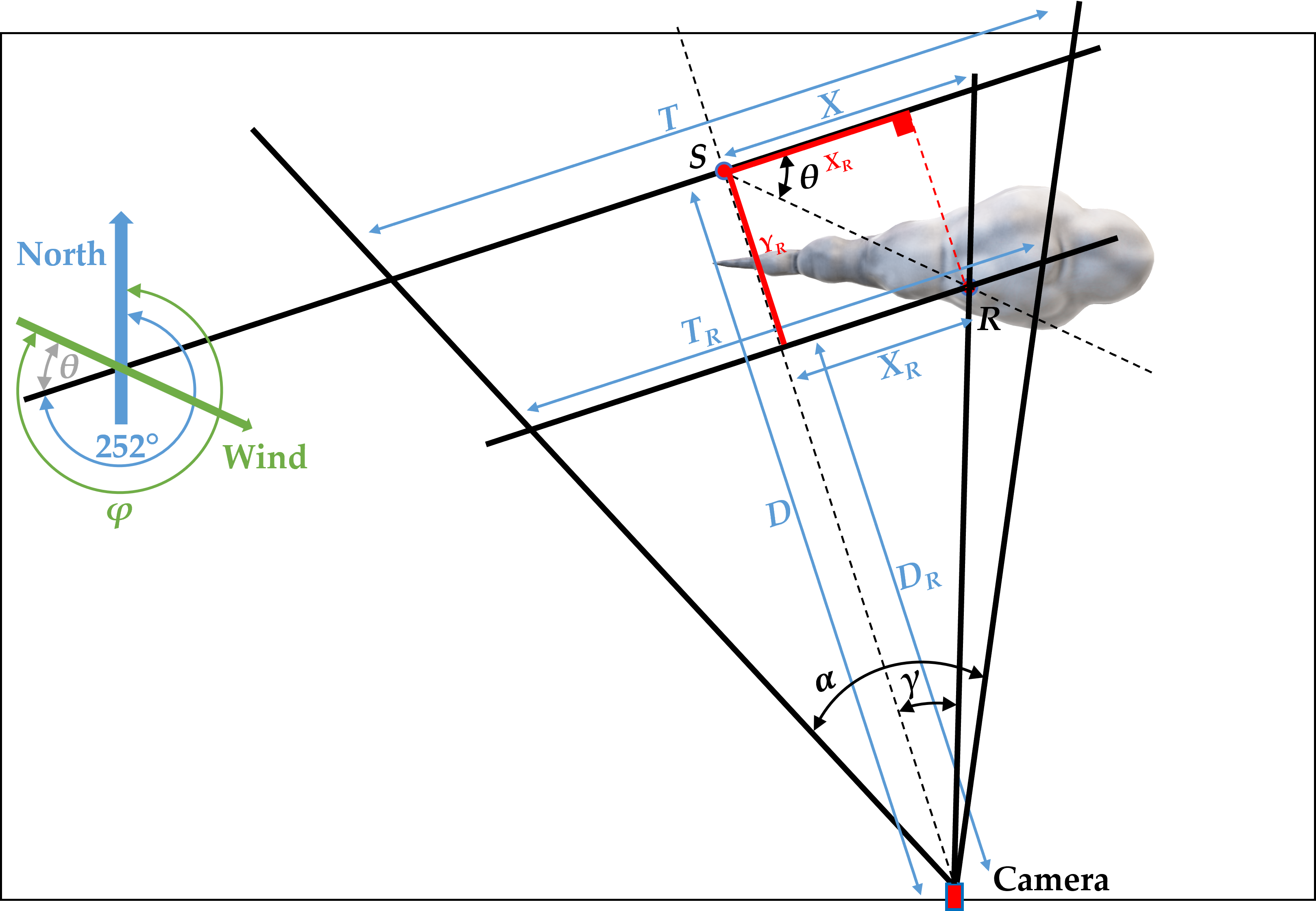}
  \caption{Schematic top view of the region.}
  \label{fig.Schematic top view of the region.}
\end{center}
\end{figure}
\unskip

\begin{figure}[th]
\begin{subfigure}{.5\textwidth}
  \centering
  \includegraphics[width=6 cm]{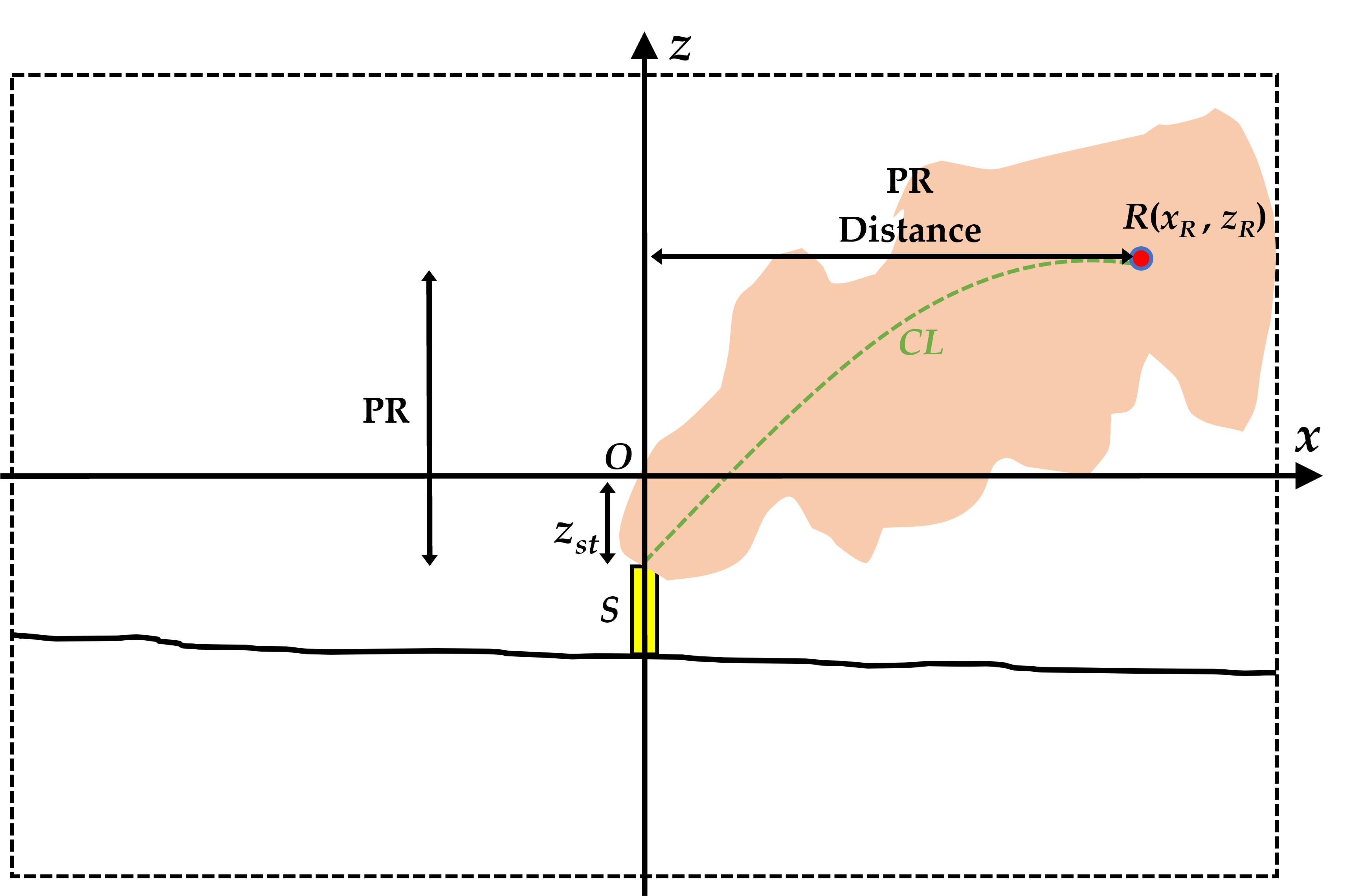}
  \caption{Camera view scheme.}
  \label{subfig.Camera view scheme.}
\end{subfigure}
\begin{subfigure}{.5\textwidth}
  \centering
  \includegraphics[width=6 cm]{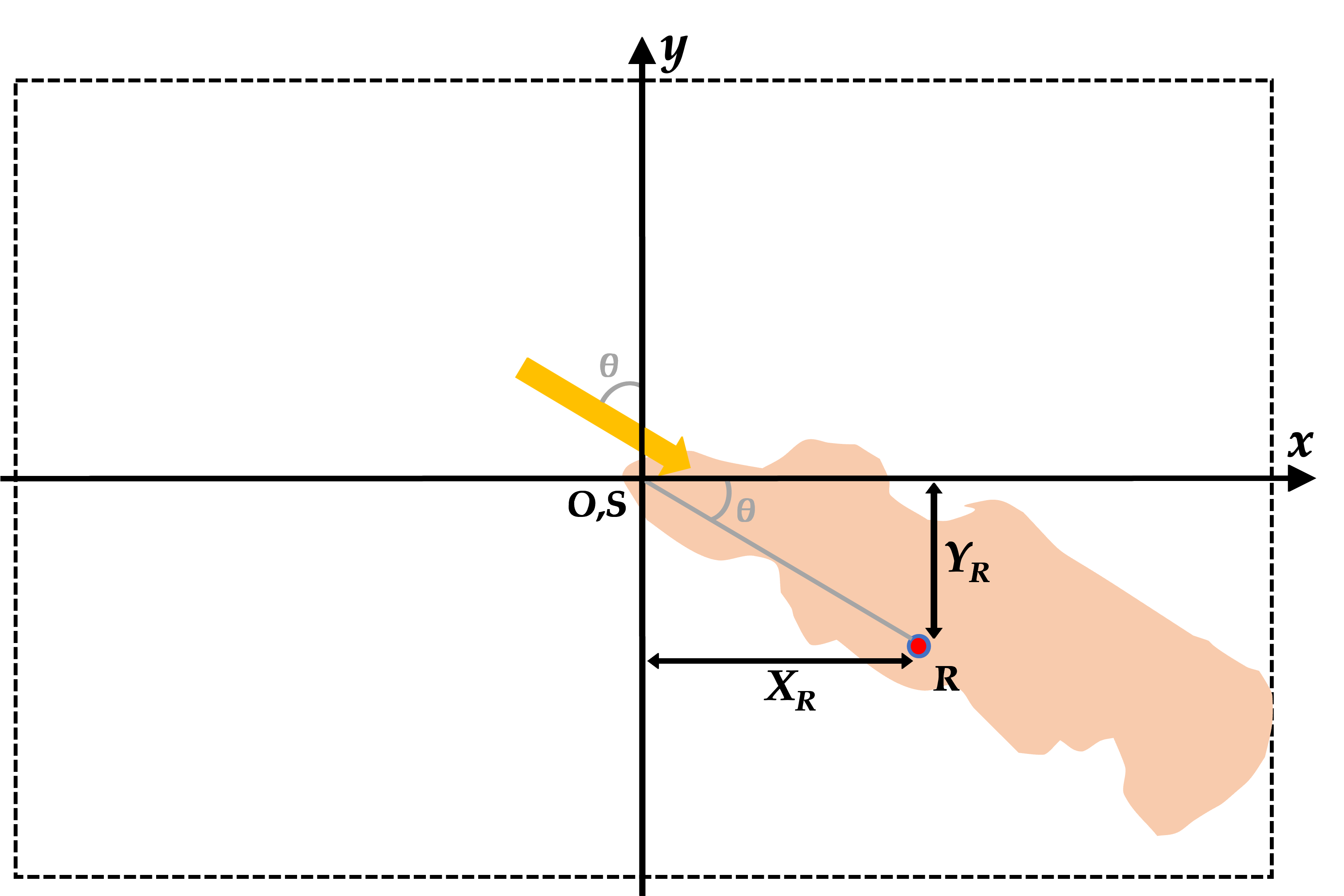}
  \caption{Top view scheme.}
  \label{subfig.Top view scheme.}
\end{subfigure}
\caption{Smokestack location schemes. Smokestack location, \emph{S}; image center, \emph{O}; desired point, \emph{R}; PC centerline, \emph{CL}; point horizontal distance from the image center, $x_{R}$; the point vertical distance from the image center, $z_{R}$; the point distance from the image center, $z_{st}$; depth of the point in the real world, $Y_{R}$; wind direction angle relative to the image plane, $\theta$; and the yellow arrow shows the wind direction.}
\label{fig.Photogrammetry parameters and smokestack location schemes.}
\end{figure}

As observed in Figures \ref{fig.Schematic top view of the region.} and \ref{fig.Photogrammetry parameters and smokestack location schemes.}, the PC representative point $R$ is affected by wind direction and is out of the image plane. Wind direction is always reported as degrees from the north, represented by $\varphi$. For instance, $\varphi=90^\circ$ is wind from the east, and $\varphi=180^\circ$ is wind from the south. For the configuration used in this study, wind direction relative to the image plane can be obtained as $\theta=|\varphi-252|$ (Figure \ref{fig.Schematic top view of the region.}). Accordingly, $X_{R}$ can be calculated by the following equation,

\begin{linenomath}\label{Eq.PS and GSD263223}
\begin{equation}
    X_{R} =
\begin{dcases}
    \frac{D}{\tan{\theta}+\frac{1}{\tan{\gamma}}} & \text{if } \theta\geq 0\\
    \frac{D}{\frac{1}{\tan{\gamma}}-\tan{\theta}} & \text{otherwise}
\end{dcases}
\end{equation}
\end{linenomath}

\noindent where $\gamma$ and $\theta$ are additional parameters which help us define equations as concisely as we can get. $D$ indicates the distance between the camera and the smokestack.

We define $G_{R}$ as the value of the ground sample distance at the location of \emph{R}. Consequently, based on \cite{luhmann2006close},

\begin{linenomath}\label{Eq.PS and GSD267865}
    \begin{equation}
    \begin{aligned}
    & G_{R} = \frac{X_{R}}{x_{R}}\\
    & Z_{R} = G_{R} \times z_{R}\\
    & Z_{st} = G \times z_{st}
    \end{aligned}
    \end{equation}
\end{linenomath}

\noindent where $Z_{st}$ and $z_{st}$ are the distance between the smokestack exit and the image center, respectively, in real-life and on image.

\unskip

Thus, the PR and the PR distance for each PC can be calculated as,

\begin{linenomath}\label{Eq.PS and zzD263223}
\begin{equation}
    \Delta{z} =
\begin{dcases}
    |Z_{R}|-|Z_{st}|,& \text{if } |Z_{R}|\geq|Z_st|\\
    |Z_{st}|-|Z_{R}|,& \text{otherwise}
\end{dcases}
\end{equation}
\end{linenomath}

\begin{linenomath}\label{Eq.PS and GSD626467865}
\begin{equation}
    X_{max} = \sqrt{X_{R}^2 + Y_{R}^2}
    \end{equation}
\end{linenomath}

\section{Experimental results and discussion}\label{section.Experimental results and discussion}
In this section, we describe our image datasets and the industrial area in which these image datasets have been collected and shared. Also, we will explain the validation metrics used to compare our proposed method with the other competitive methods in smoke border detection and recognition. Then, our discussion falls into two last sections, named comparison with existing smoke recognition methods and plume rise measurement, in which the performance of the proposed method is evaluated, and the PR is calculated based on our "DPRNet," respectively. To validate the performance of our proposed method, we used a computer equipped with Core i9, 3.70 GHz/4.90 GHz, 20 MB cache CPU, 64GB RAM and NVIDIA GeForce RTX 3080,10 GB graphic card. The total training time of the network was about one hour using Python 3.8 with PyTorch Deep Learning framework. Finally, for the geometric transformation and image processing analysis, we used MATLAB R2022b software.

\subsection{Site description}\label{subsection.Site description}
\begin{figure}[ht]
  \begin{subfigure}{0.5\textwidth}
    \centering
    \includegraphics[width=5 cm]{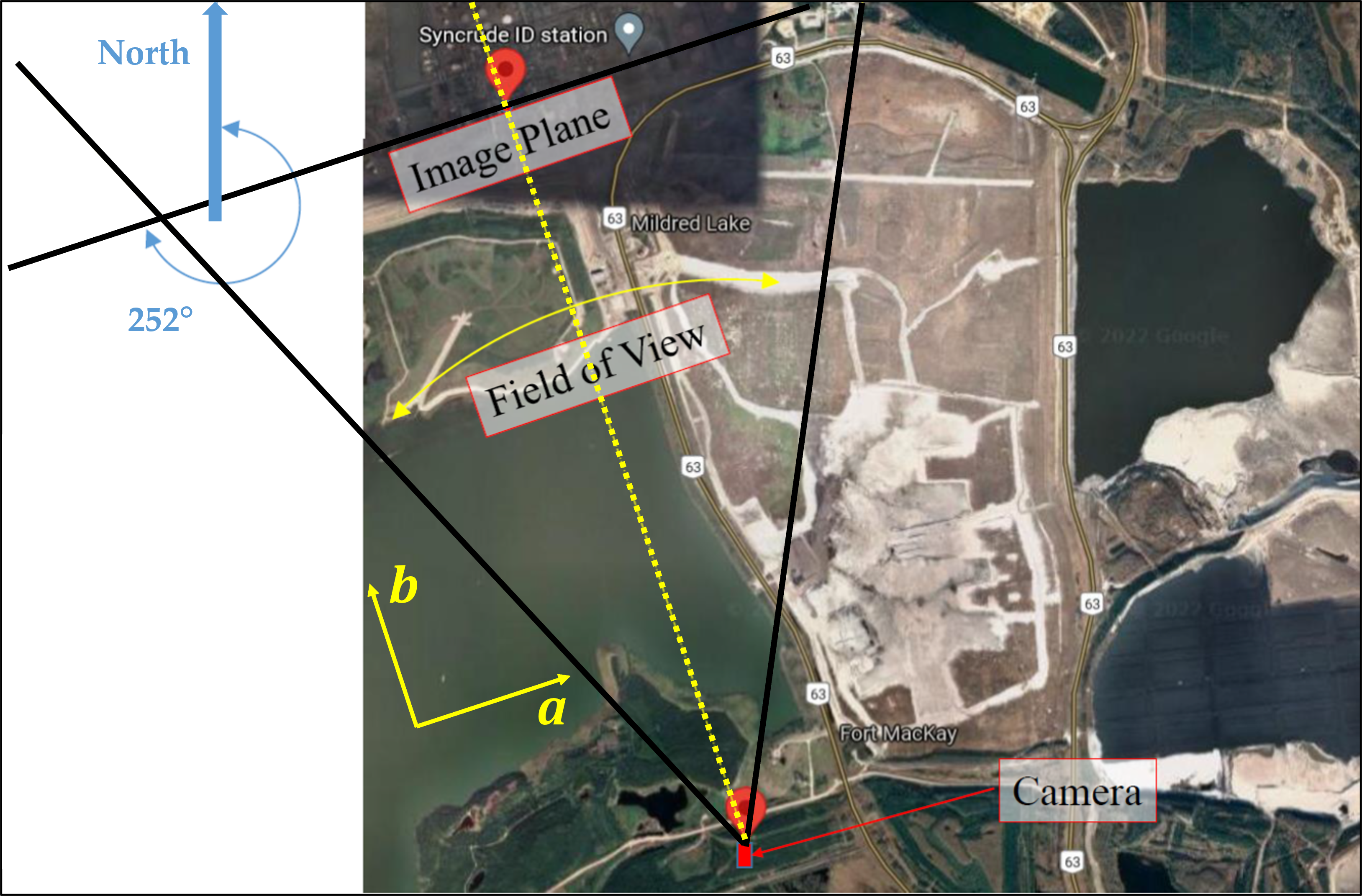}
    \caption{Top view}
    \label{subfig.Top view}
  \end{subfigure}
  \begin{subfigure}{0.5\textwidth}
    \centering
    \includegraphics[width=6.5 cm]{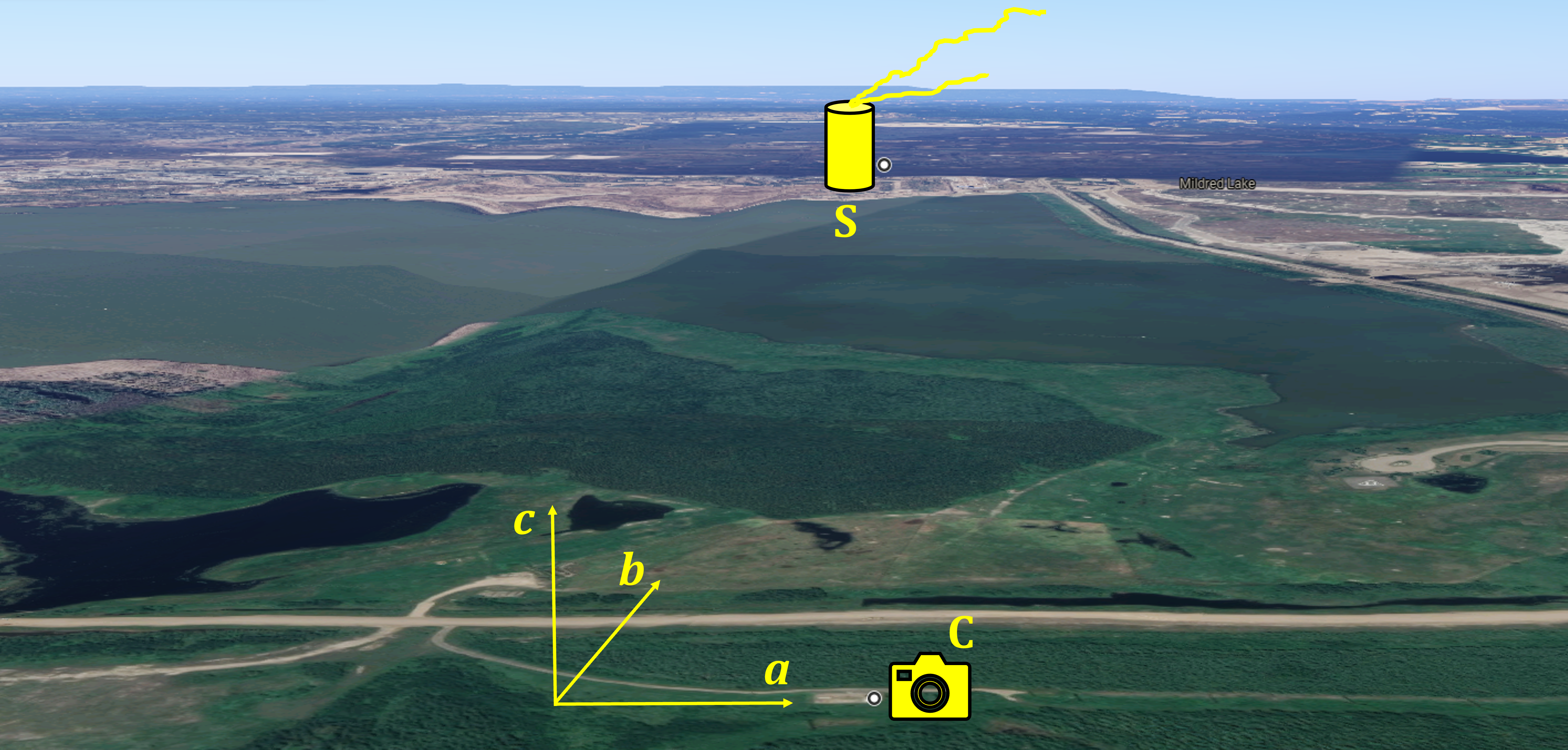}
    \caption{3D view}
    \label{subfig.3D view}
  \end{subfigure}
  \caption{Imaging situation. Camera station, \emph{C}; and smokestack position, \emph{S}. The \emph{abc} coordinate system is only for differentiating the side and camera views and is not used as a coordinate reference system.}
  \label{fig.Map views of the region}
\end{figure}

The imaging system was deployed on a meteorological tower with a clear sightline to the desired smokestack operated by the Wood Buffalo Environment Association (WBEA). It is located outside the Syncrude oil sands processing facility north of Fort McMurray, Alberta, Canada. Figure \ref{fig.Map views of the region} represents the satellite images, the location of the camera, and the desired smokestack.

\unskip

WBEA operates a 10-meter-tall meteorological tower with a clear sightline to the smokestack at Syncrude (\url{https://wbea.org/stations/buffalo-viewpoint}). The camera system is mounted on this tower above the tree canopy because they are on a hill sloped downward from the tower location, and the biggest smokestack and its PC are always visible. The system consists of a digital camera with shutter control and a camera housing for weather protection with interior heating for window defrost and de-icing. The station powers the camera activation, and the images are recorded on a laptop.

The Syncrude processing facility has six main smokestacks. The tallest one is about 183 m, and the heights of the other five are between 31 m to 76 m. To isolate a single smoke plume rise, we have concentrated on the area's tallest one, which can help find the PR for one plume source. All six smokestacks are listed in Table \ref{table.Syncrude stacks information, including location}. Wind directions during the capturing period were determined from the Mildred lake Air Monitoring Station (\url{https://wbea.org/stations/mildred-lake}), which is located at the Mildred Lake airstrip (AMS02: Latitude: $57.05^{\circ}$, Longitude: $-111.56^{\circ}$), approximately 5 km from the Syncrude facility.

\begin{table}[H]
\caption{Syncrude smokestacks information, including location, smokestack height ($h_{s}$), smokestack diameter ($d_{s}$), effluent velocity at the smokestack exit ($\omega_{s}$), and effluent temperature at the smokestack exit ($T_{s}$). The velocities and temperatures are averages for the entire capturing period.\label{table.Syncrude stacks information, including location}}
\newcolumntype{C}{>{\centering\arraybackslash}X}
\begin{tabularx}{\textwidth}{CCCCCCC}
\toprule
\textbf{Reported ID} & \textbf{Latitude} & \textbf{Longitude} & $\boldsymbol{h_{s}}$\textbf{(m)} & \textbf{$\boldsymbol{d_{s}}$\textbf{(m)}} & \textbf{$\boldsymbol{\omega_{s}}$($\mathbf{ms^{-1}}$)} & \textbf{$\boldsymbol{T_{s}}$\textbf{(K)}}\\
\midrule
Syn. 12908 & 57.041 & -111.616 & 183.0 & 7.9 & 12.0 & 427.9 \\
Syn. 12909 & 57.048 & -111.613 & 76.2 & 6.6 & 10.1 & 350.7 \\
Syn. 13219 & 57.296 & -111.506 & 30.5 & 5.2 & 8.8 & 355.0 \\
Syn. 16914 & 57.046 & -111.602 & 45.7 & 1.9 & 12.0 & 643.4 \\
Syn. 16915 & 57.046 & -111.604 & 31.0 & 5.0 & 9.0 & 454.5 \\
Syn. 16916 & 57.297 & -111.505 & 31.0 & 5.2 & 9.2 & 355.0 \\
\bottomrule
\end{tabularx}
\end{table}

\subsection{DPRD}\label{subsection.Dataset}
The greatest challenge in using deep learning for PC recognition is inadequate annotated images for training. Hence, creating image datasets for PC recognition for research and industry purposes is invaluable. For this study, 96 images were captured every day, and for the first part of the project, \textbf{35K} images were collected from \textbf{January 2019} to \textbf{December 2019}. The collected images demonstrated various types of plume shapes in different atmospheric conditions. Dataset has been classified into day, night, and cloudy/foggy conditions. The collected dataset revealed that among 96 images captured daily, we have 48 day and 48 night images. There were some outlier images for different reasons, such as camera handle shaking, auto-focus problems, disturbing smoke and severe snow and hail. Furthermore, some PCs could not be recognized from their background, even by visual image inspection. As a consequence, among \textbf{35K} collected images, \textbf{10684} images were valid. Note that, among \textbf{10684} collected valid images, the facility is not working in \textbf{2374} images.

For this paper, a new benchmark, DPRD including a 2500 annotated dataset is introduced. 60\% of DPRD is considered as training data, and 40\% is used for validation and testing purposes. Rows (a) and (b) in Figure \ref{fig.RRResults} shows sample images from the region and their corresponding ground truth, which are generated by the "Labelme" graphical image annotation tool at \url{https://github.com/wkentaro/labelme}. We tried to select images of different atmospheric conditions, such as clear daytime, nighttime, cloudy, and foggy, to represent the results of different situations.

\begin{figure}[ht]
\centering
\begin{subfigure}[t]{0.15\textwidth}
    \includegraphics[width=2.8 cm]{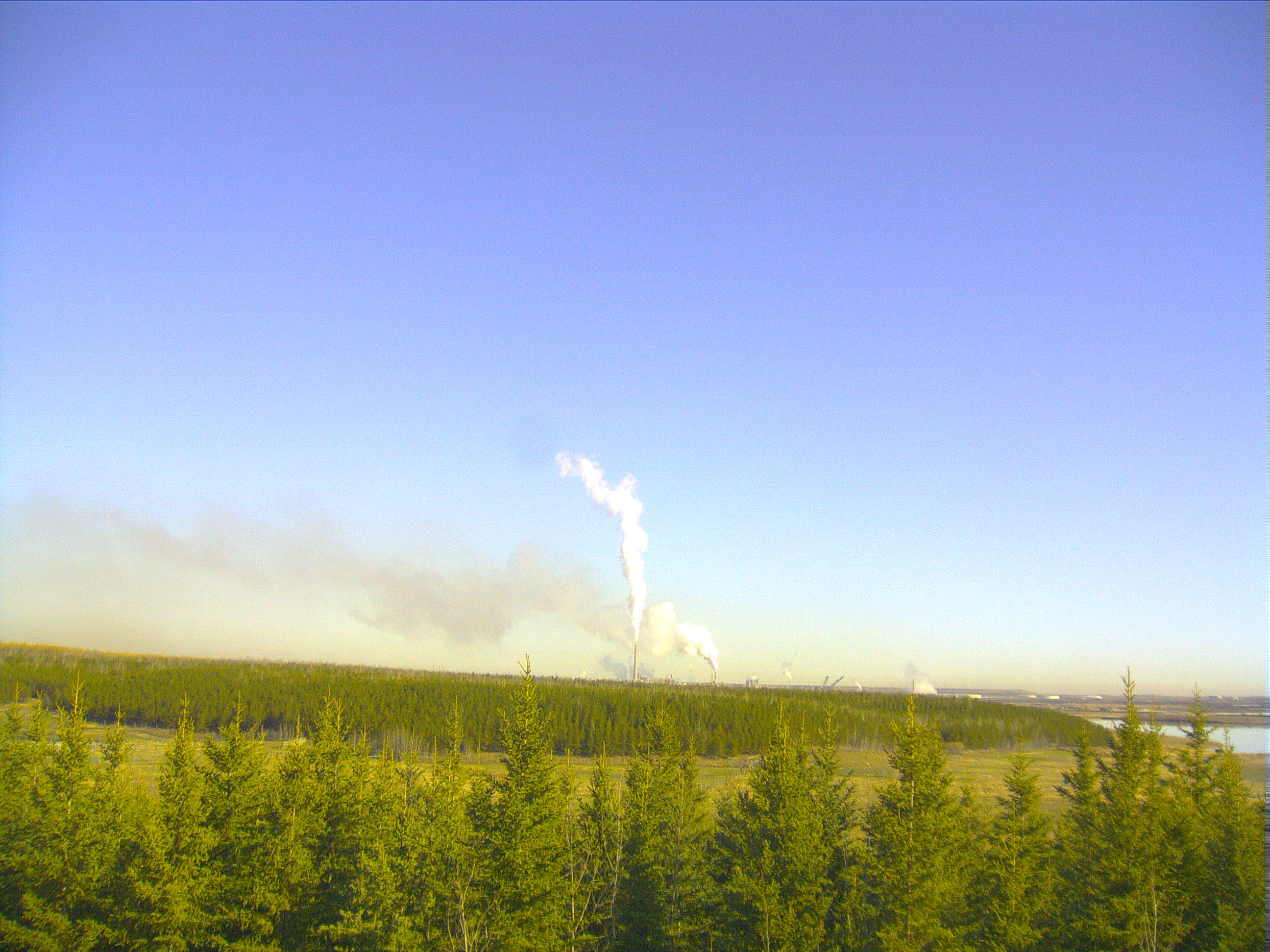}
    \includegraphics[width=2.8 cm]{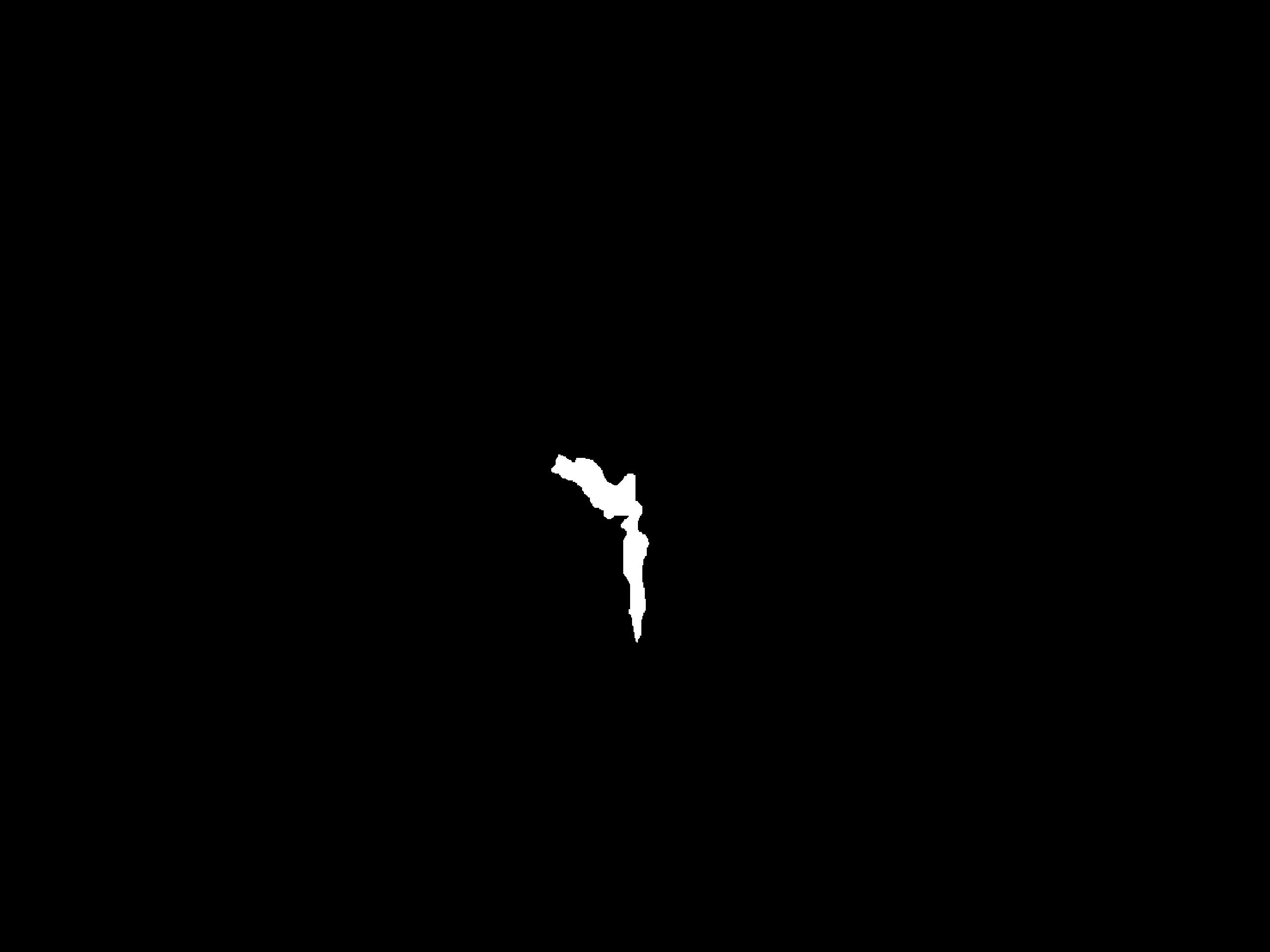}
    \caption{}
\end{subfigure}\hfill
\begin{subfigure}[t]{0.15\textwidth}
    \includegraphics[width=2.8 cm]{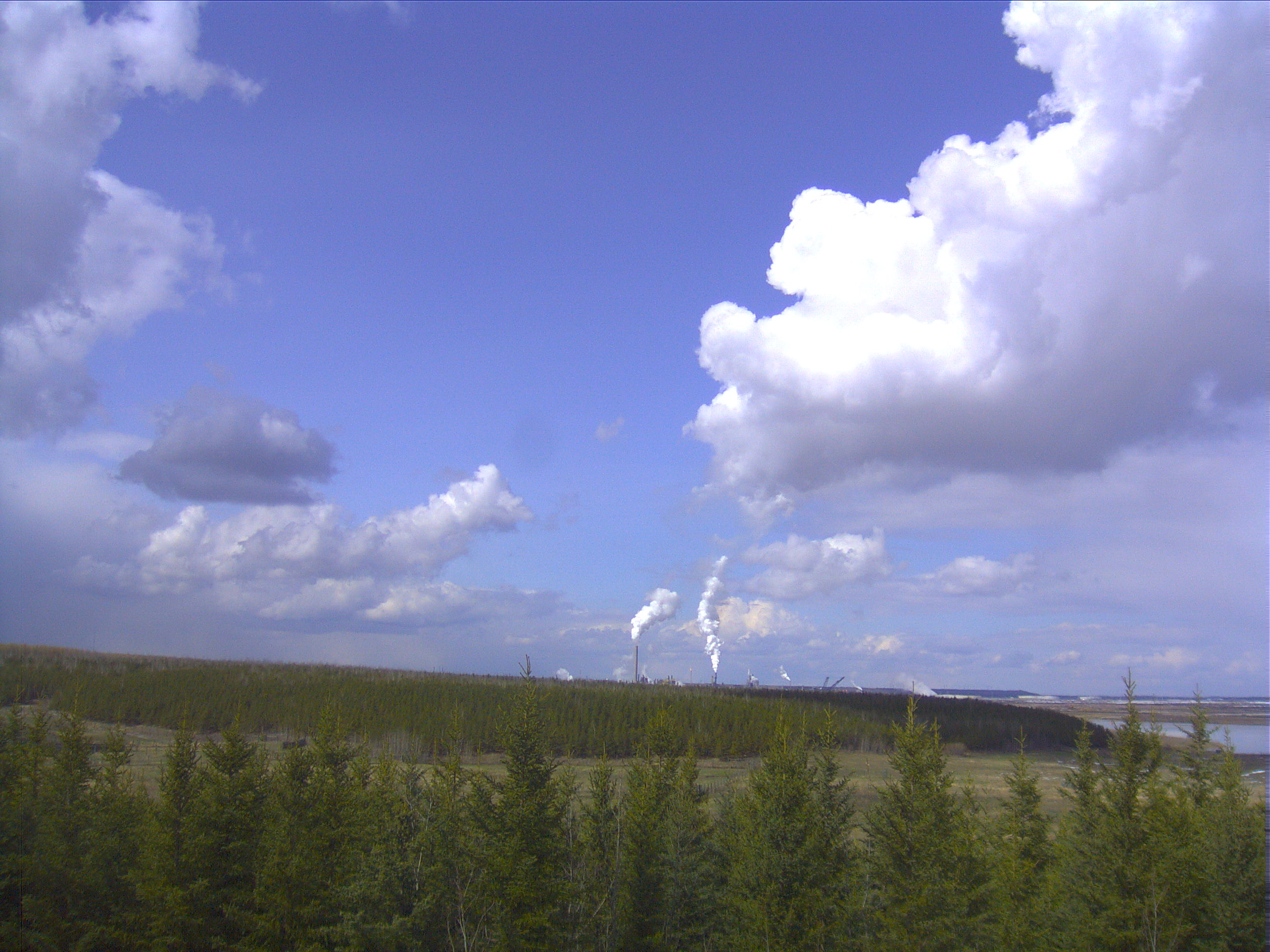}
    \includegraphics[width=2.8 cm]{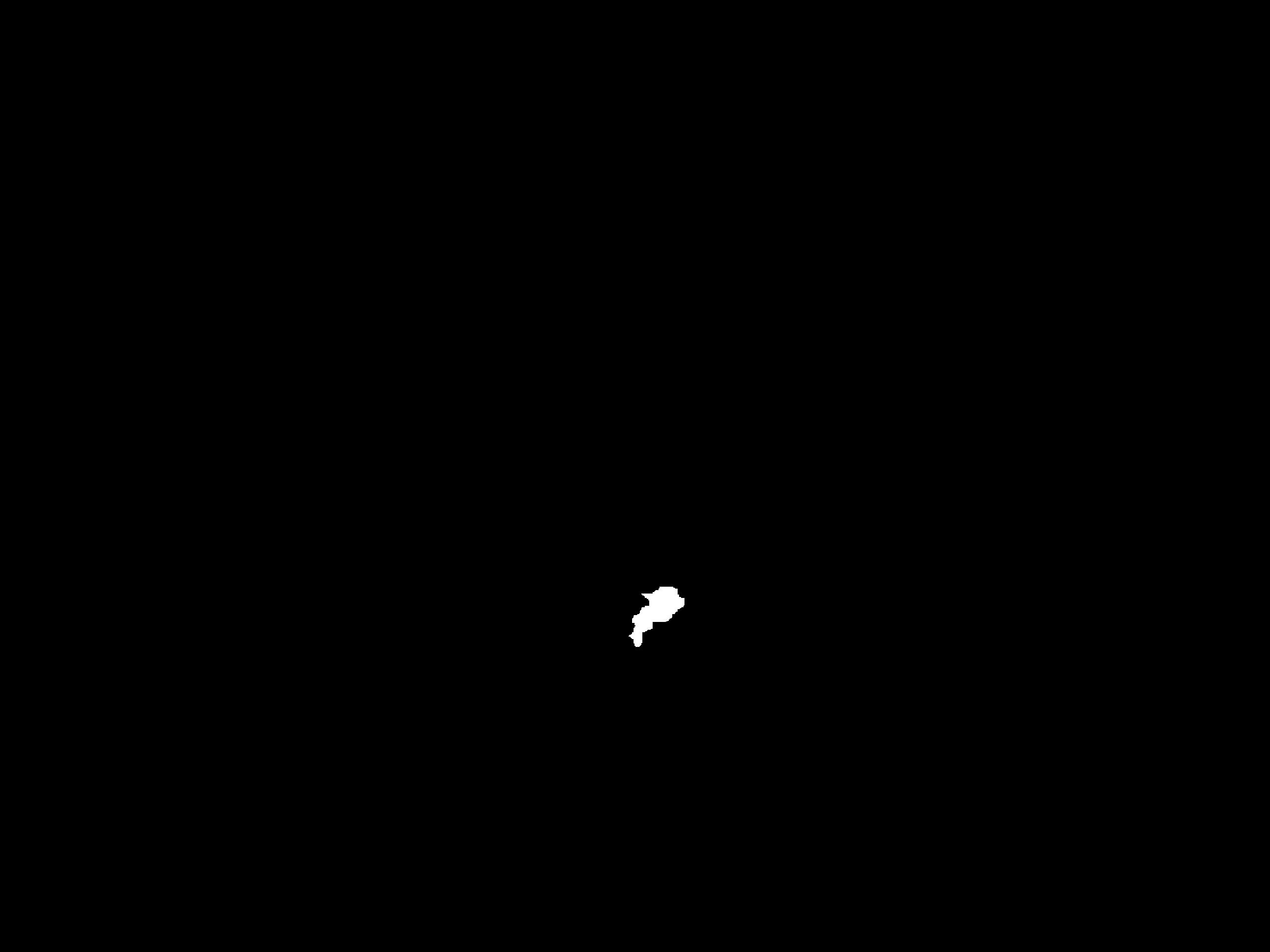}
    \caption{}
\end{subfigure}\hfill
\begin{subfigure}[t]{0.15\textwidth}
    \includegraphics[width=2.8 cm]{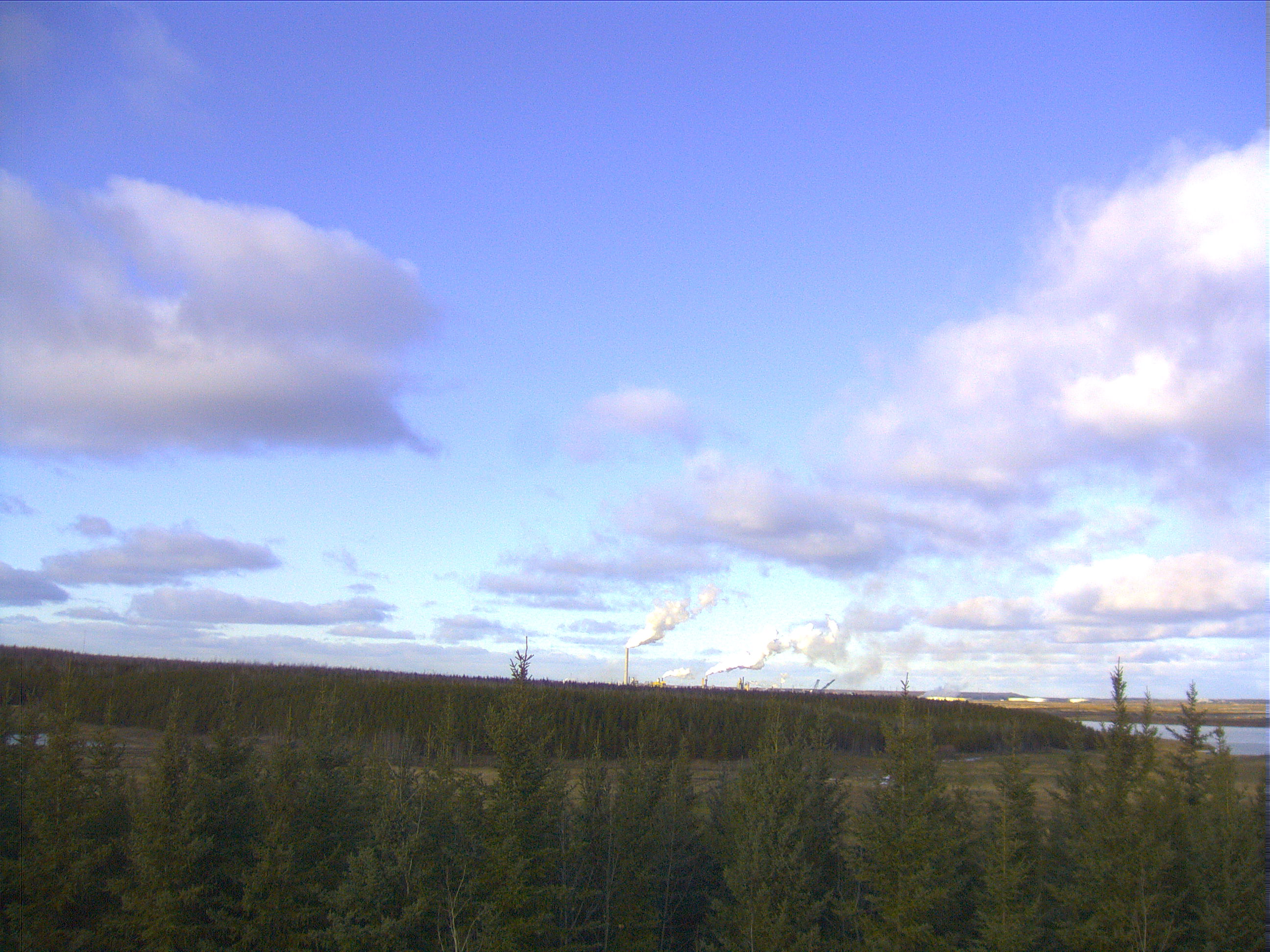}
    \includegraphics[width=2.8 cm]{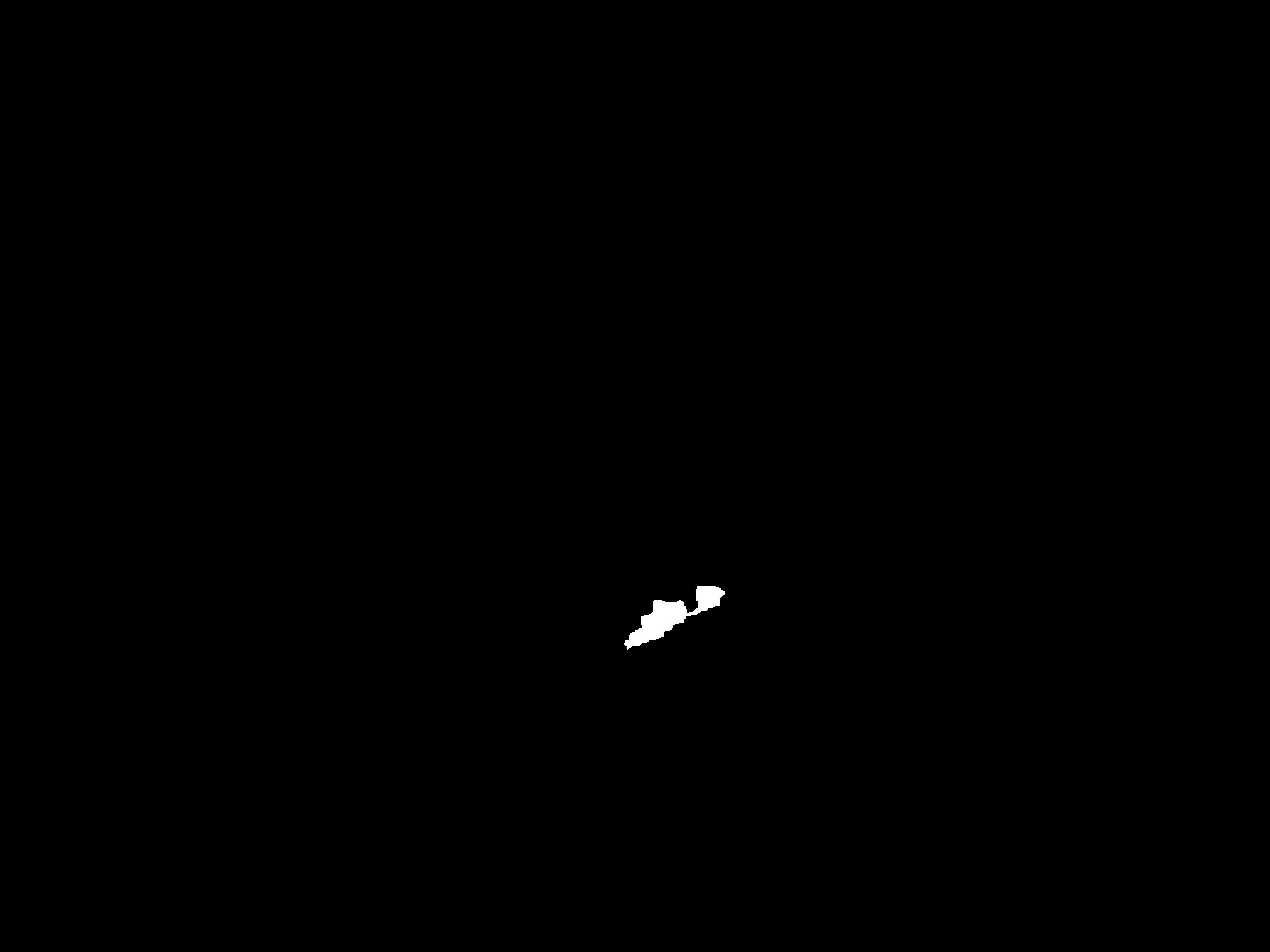}
    \caption{}
\end{subfigure}\hfill
\begin{subfigure}[t]{0.15\textwidth}
    \includegraphics[width=2.8 cm]{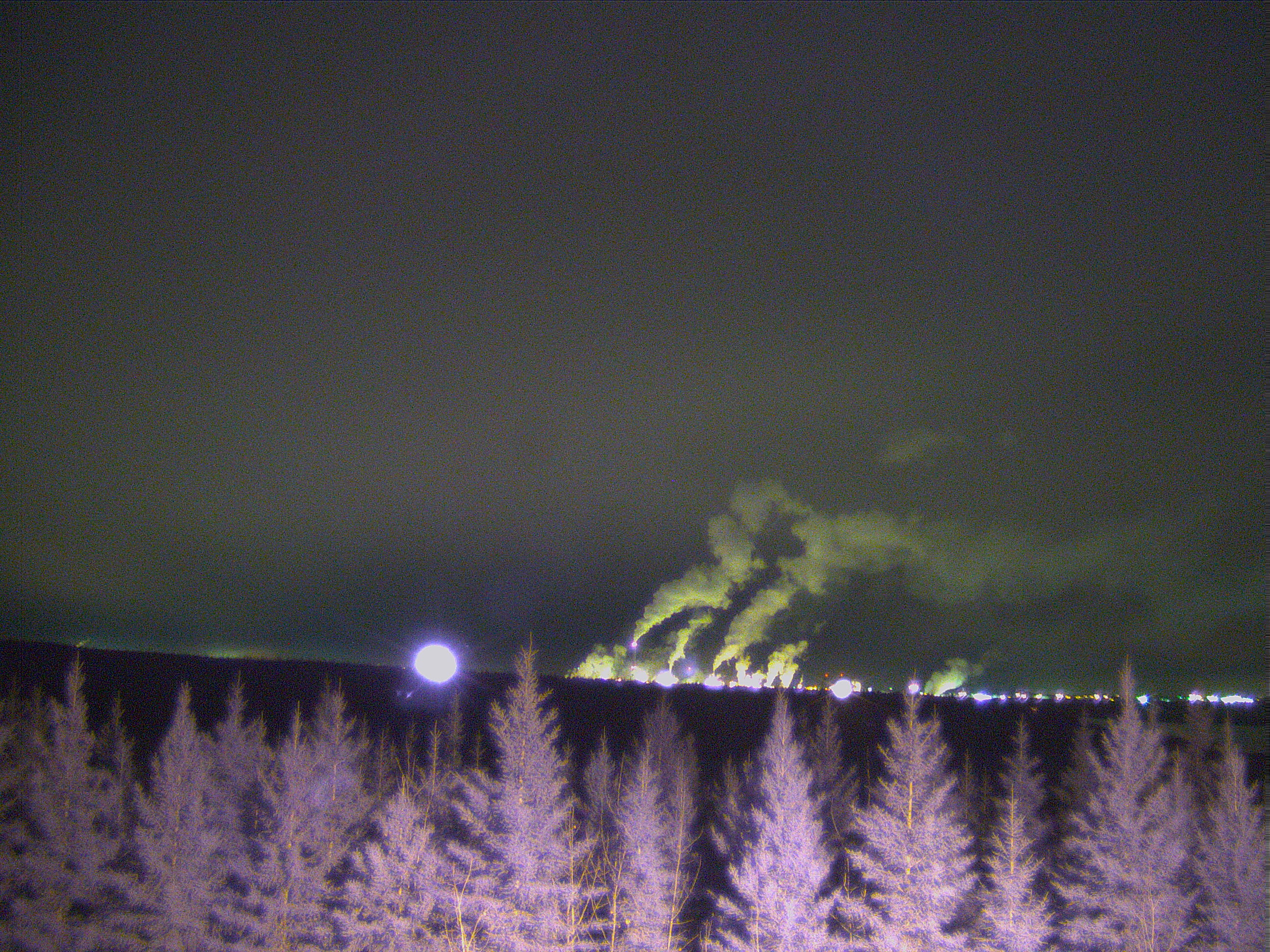}
    \includegraphics[width=2.8 cm]{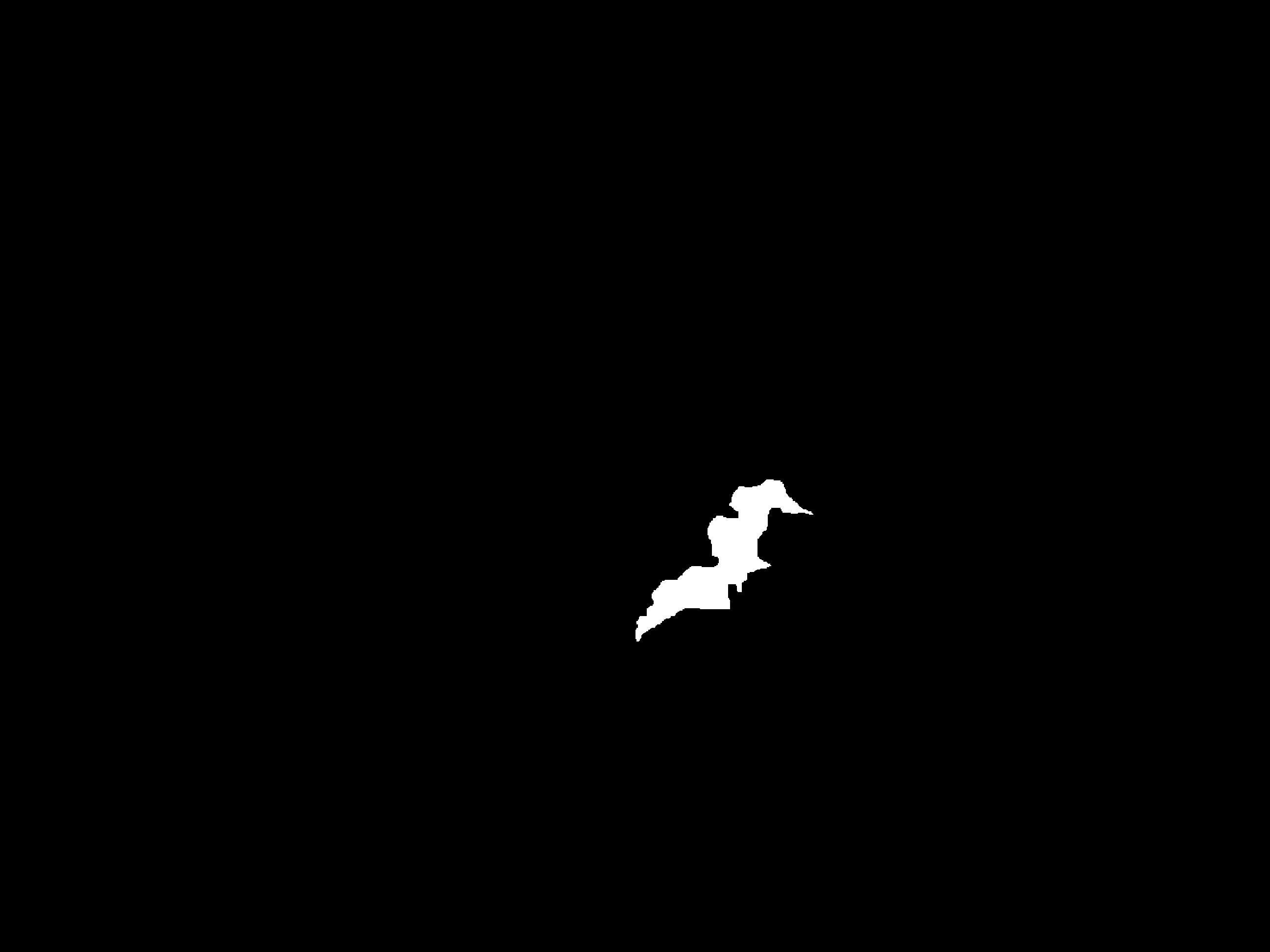}
    \caption{}
\end{subfigure}\hfill
\begin{subfigure}[t]{0.15\textwidth}
    \includegraphics[width=2.8 cm]{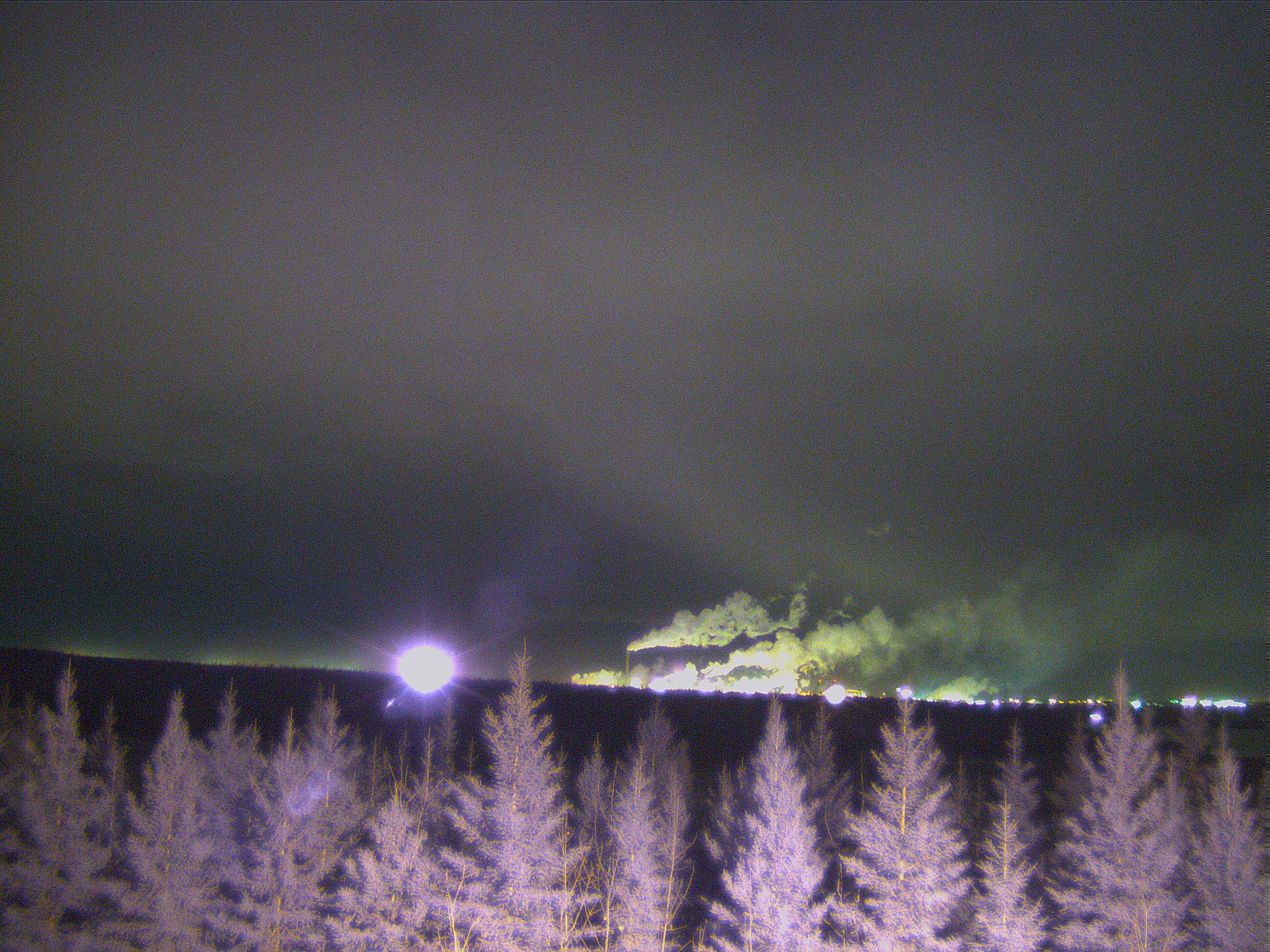}
    \includegraphics[width=2.8 cm]{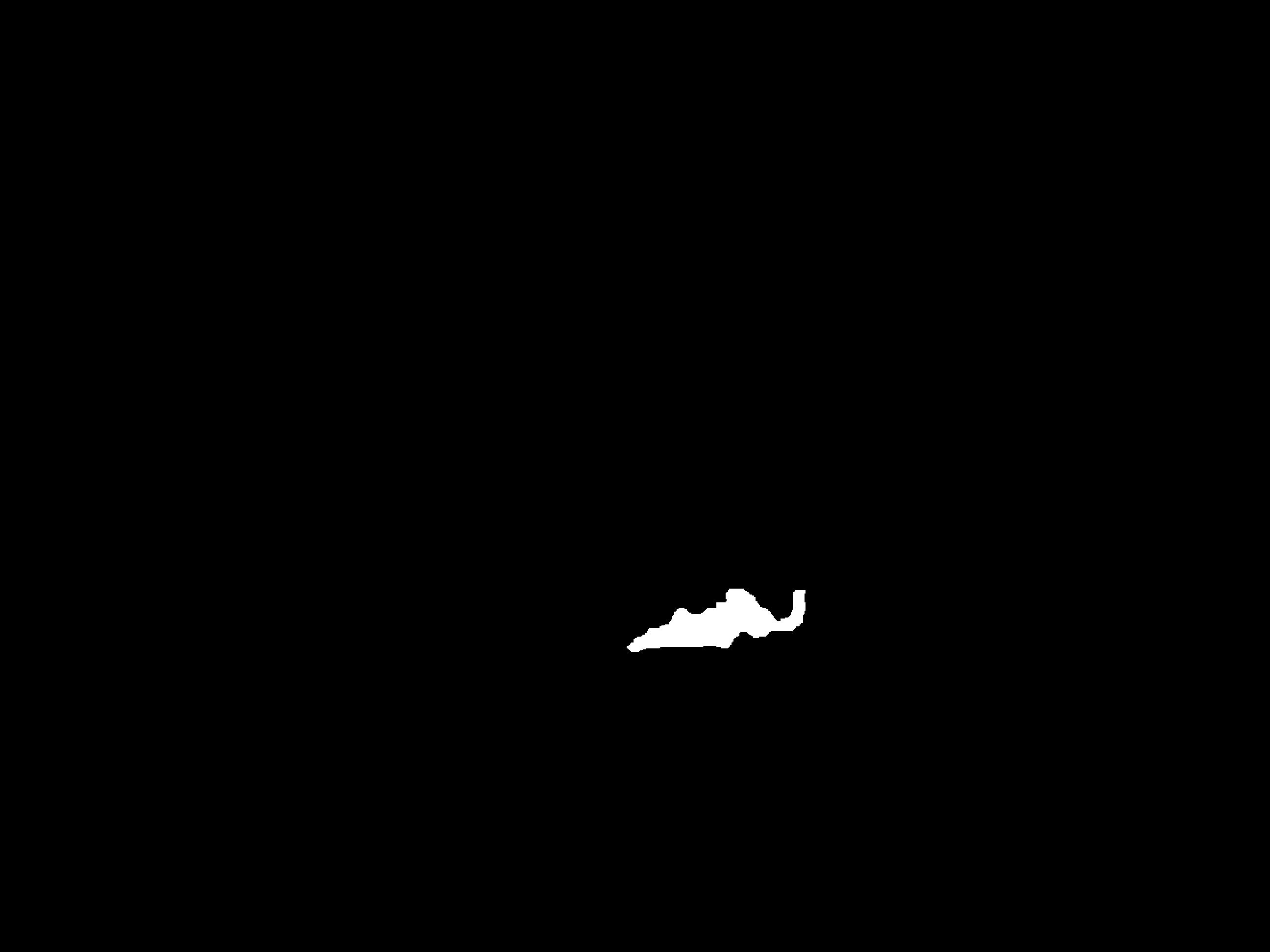}
    \caption{}
\end{subfigure}\hfill
\caption{Sample images (up) and their corresponding ground truth (down) from our DPR dataset listed as (\textbf{a}) Clear daytime, (\textbf{b})\&(\textbf{c}) cloudy day, and (\textbf{d})\&(\textbf{e}) clear nighttime.}
\label{fig.SSSamplesss}
\end{figure}
\unskip

\subsection{Model validation metrics}\label{subsection.Model validation metrics}
The performance of the methods in question is evaluated using the metrics of accuracy, recall, precision and F1 score. These metrics are defined using four values of True Positive (TP), True Negative (TN), False Positive (FP), and False Negative (FN) obtained from the confusion matrix of each introduced method \cite{dougherty2012pattern}. The accuracy validation metric is the ratio of observations predicted correctly to the total observations. In our application, the model's accuracy represents how accurately our model can recognize the PC pixels. This criterion is valid as long as the values of FP and FN are almost the same \cite{dougherty2012pattern}. Otherwise, other validation metrics should be considered. The foreground pixel coverage of the sample images are shown in Figure \ref{fig.SSSamplesss}, which confirms the fact that accuracy is not suitable for this study. Recall or sensitivity is the ratio of positive observations predicted correctly to all actual observations. Recall shows how many PC pixels are labelled among all the actual PC pixels. The recall is obtained as follows,

\begin{linenomath}\label{Eq.Recall}
    \begin{equation}
    Recall = \frac{TP}{TP+FN}
    \end{equation}
\end{linenomath}

Precision is the ratio of positive observations which are predicted correctly to all observations which are predicted as positive. This metric represents how many PC pixels exist among all the pixels labelled as PC. Therefore, a low rate of FP can achieve high precision. This validation metric is obtained as follows,

\begin{linenomath}\label{Eq.Precision}
    \begin{equation}
    Precision = \frac{TP}{TP+FP}
    \end{equation}
\end{linenomath}

As it is implied from the Equations \ref{Eq.Recall} and \ref{Eq.Precision}, precision and recall take either FP or FN into account. The last validation measure in this paper, the F1 score, considers both FP and FN as a weighted average of recall and precision metrics. Unlike accuracy, this metric is more useful when FP and FN are not the same as in our study. Our FP is less than FN, or the amount of non-actual PC pixels predicted as PC pixels is less than that of actual PC pixels predicted as non-PC pixels. Therefore, the F1 score helps us look at both recall and precision validation metrics as follows,

\begin{linenomath}\label{Eq.F1Score}
    \begin{equation}
    F1 score = \frac{{2}\times{Recall}\times{Precision}}{Recall+Precision}
    \end{equation}
\end{linenomath}

\subsection{Comparison with existing smoke recognition methods}\label{subsection.Competing methods and discussion}
In this section, we evaluate the performance of DPRNet and compare it with several competitors. To choose suitable smoke recognition methods for comparison, we considered both identification accuracy and computational complexity of the reviewed methods, which led to the selection of DeepLabv3+ \cite{chen2018encoder}, FCN \cite{long2015fully}, and regular Mask R-CNN. Our proposed DPRNet is evaluated using three metrics introduced in Section \ref{subsection.Model validation metrics}. The confusion matrix of DPRNet recognition is given in Table \ref{table.Confusion matrix of our DPRNet recognition model.} for three specific conditions of daytime, nighttime, and cloudy \& foggy.

\begin{table}[H]
\caption{Average confusion matrix of our DPRNet recognition model.\label{table.Confusion matrix of our DPRNet recognition model.}}
\newcolumntype{C}{>{\centering\arraybackslash}X}
\begin{tabularx}{\textwidth}{CCCCCCC}
\toprule
{}	& \textbf{P-D (\%)} & \textbf{NP-D (\%)} & \textbf{P-N (\%)} & \textbf{NP-N (\%)} & \textbf{P-CF (\%)} & \textbf{NP-CF (\%)}\\
\midrule
\textbf{P\textsuperscript{1}-D\textsuperscript{2}}		& 99.2			& 0.04		& -		& -		& -		& -\\
\textbf{NP\textsuperscript{3}-D}		& 0.12		    & 0.68		& -		& -		& -	    & -\\
\textbf{P-N\textsuperscript{4}}		& -			& -		& 99.06		& 0.09		& -		& -\\
\textbf{NP-N}		& -			& -		& 0.09		& 0.76		& -		& -\\
\textbf{P-CF\textsuperscript{5}}		& -			& -		& -		& -		& 99.09		& 0.05\\
\textbf{NP-CF}		& -			& -		& -		& -		& 0.13		& 0.72\\
\bottomrule
\end{tabularx}
\noindent{\footnotesize{\textsuperscript{1} Plume}}
\noindent{\footnotesize{\textsuperscript{2} Day}}
\noindent{\footnotesize{\textsuperscript{3} Non-plume}}
\noindent{\footnotesize{\textsuperscript{4} Night}}
\noindent{\footnotesize{\textsuperscript{5} Cloudy and foggy}}
\end{table}

As is clear from Table \ref{table.fr111}, DPRNet has much better performance than competitive methods in terms of all validation metrics. In detail, the recall and precision metrics express the reasonable difference between the models, which shows the effectiveness of the proposed model in recognizing the actual PC pixels. Compared to the rivals, the more considerable value of the F1 score guarantees that DPRNet outperforms the other three methods and shows the efficacy of this method. Among our competitive methods, DeepLabv3 performed better regarding all validation metrics, and Mask R-CNN had the worst performance.

\begin{table}[H]
\caption{Comparison of different methods for plume cloud recognition using average validation metrics values.\label{table.fr111}}
\newcolumntype{C}{>{\centering\arraybackslash}X}
\begin{tabularx}{\textwidth}{CCCC}
\toprule
\textbf{Model} & \textbf{Recall} & \textbf{Precision} & \textbf{F1 score}\\
\midrule
Mask R-CNN & 0.556 & 0.727 & 0.607 \\
FCN & 0.591 & 0.859 & 0.599 \\
DeepLabv3 & 0.654 & 0.892 & 0.721 \\
DPRNet & \textbf{0.846} & \textbf{0.925} & \textbf{0.881} \\
\bottomrule
\end{tabularx}
\end{table}

Besides these average values, the detailed statistics for each model are given in Figure \ref{fig.Performance of different methods in terms of recall, precision and F1} in terms of each used validation metrics for 90 test images selected from various day, night, foggy, and cloudy conditions. At a glance, it is observed that our proposed method shows the most robustness in all circumstances. Of competitors, Mask R-CNN and FCN have the worst performance, whereas, DeepLabv3 has the best efficiency slightly.

\begin{figure}[h!]
  \begin{subfigure}{\textwidth}
    \centering
    \includegraphics[width=13 cm]{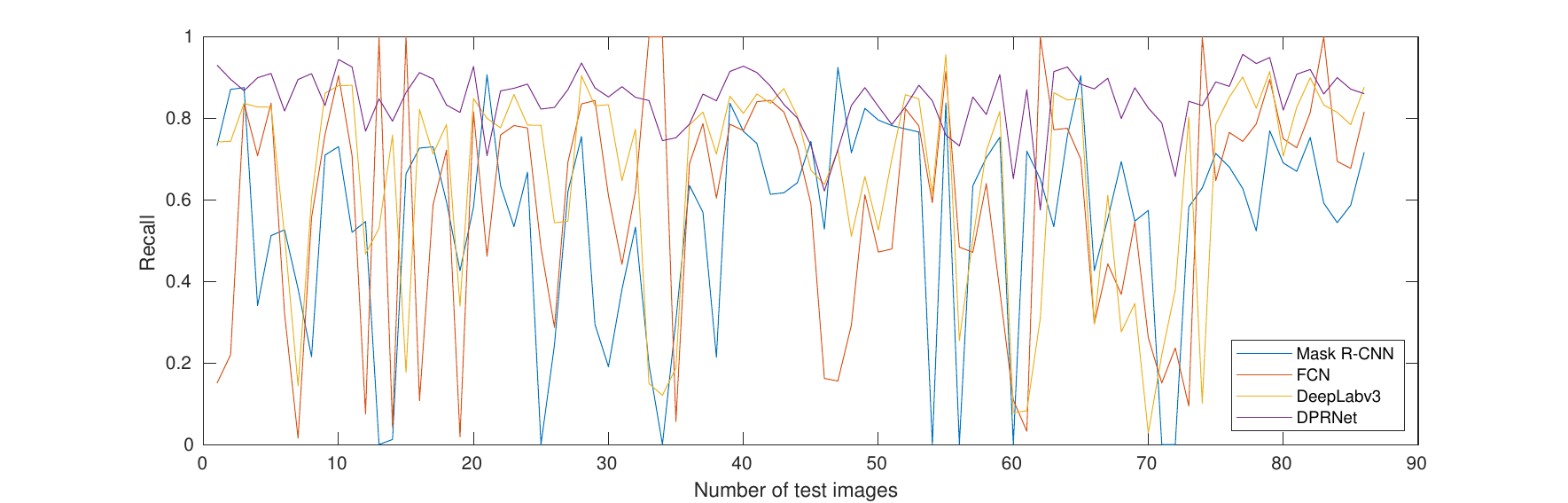}
    \caption{}
    \label{subfig.Top view}
  \end{subfigure} \\
  \begin{subfigure}{\textwidth}
    \centering
    \includegraphics[width=13 cm]{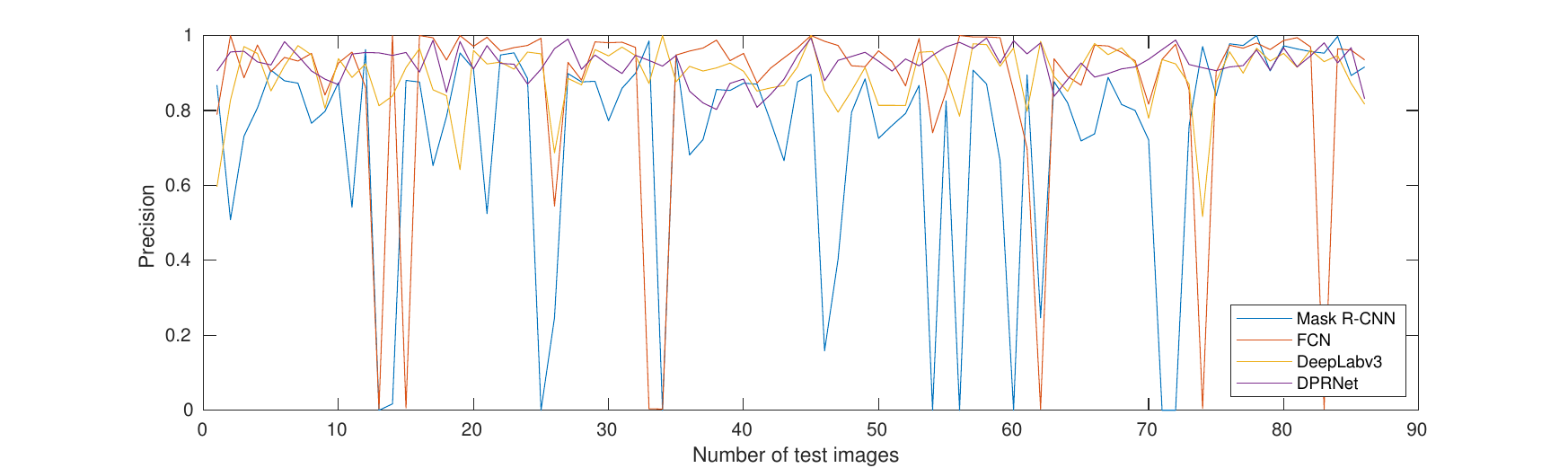}
    \caption{}
    \label{subfig.3D view}
  \end{subfigure} \\
    \begin{subfigure}{\textwidth}
    \centering
    \includegraphics[width=13 cm]{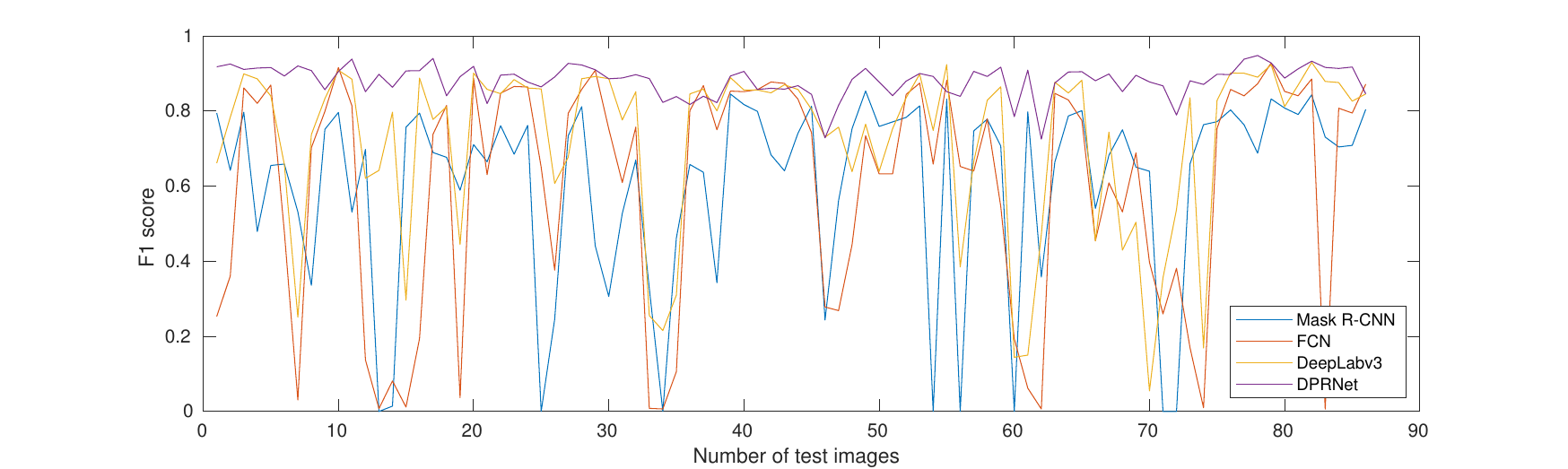}
    \caption{}
    \label{subfig.Top view}
  \end{subfigure}
  \caption{Performance of different methods regarding some test images (\textbf{a}) recall, (\textbf{b}) precision and (\textbf{c}) F1 score metrics.}
  \label{fig.Performance of different methods in terms of recall, precision and F1}
\end{figure}
\unskip

\begin{figure}[th]
\begin{subfigure}{0.3\textwidth}
  \centering
  \includegraphics[width=4 cm]{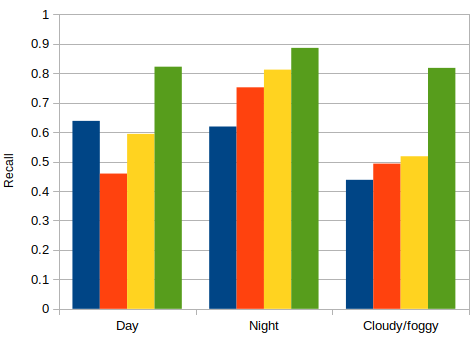}
  \caption{}
  \label{subfig.Image 1}
\end{subfigure}
\begin{subfigure}{0.3\textwidth}
  \centering
  \includegraphics[width=4 cm]{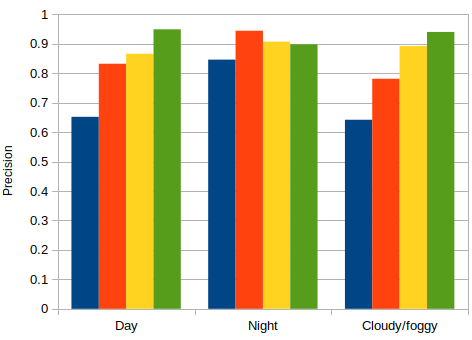}
  \caption{}
  \label{subfig.Image 2}
\end{subfigure}
\begin{subfigure}{0.3\textwidth}
  \centering
  \includegraphics[width=5 cm]{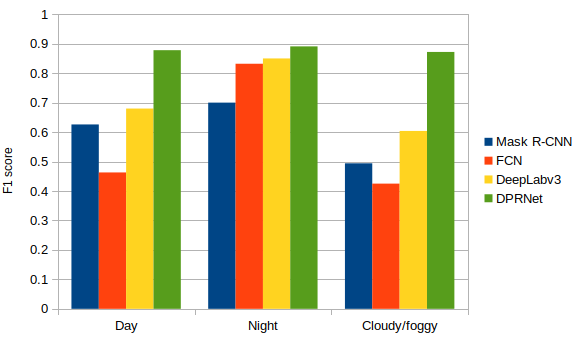}
  \caption{}
  \label{subfig.Image 3}
\end{subfigure}
\caption{Detailed comparison of methods over three datasets employing (\textbf{a}) recall, (\textbf{b}) precision and (\textbf{c}) F1 score metrics.}
\label{fig.Detailed comparison of methods over sex sample images}
\end{figure}

To further validate our DPRNet performance, we compared the models over the day, night and foggy \& cloudy datasets in terms of different validation metrics, which is given in Figure \ref{fig.Detailed comparison of methods over sex sample images}. It can be observed that all methods, except Mask R-CNN, have acceptable performance using day and night datasets. Even with night precision, FCN is better than our proposed method. However, as discussed in Section \ref{subsection.Model validation metrics}, this metric can not completely convey the merit of a model individually, and it needs to be analyzed with the F1 score. Our proposed DPRNet seems to outperform the other rival methods by recognizing roughly all of the PC pixels correctly. Most datasets are related to cloudy and foggy conditions and are frequently seen within image batches. The strength of our DPRNet is its powerful performance in this case, which is of paramount importance in our application. The DPRNet could improve the recall metric by \%66, \%58, and \%87 on average in cloudy and foggy conditions relative to FCN, DeepLabv3, and Mask R-CNN frameworks, respectively, which means that the proposed method is able to find the PC regions appropriately, using $L_{sse}$. This capability produces high-quality image recognition with a more complicated mixture of PCs and the sky behind. These high recall values help us meet our research application requirement, in which we should identify the entire PC stream for PR distance measurement.

\begin{figure}[th]
\centering
\begin{subfigure}[t]{\dimexpr0.15\textwidth+20pt\relax}
    \makebox[20pt]{\raisebox{30pt}{\rotatebox[origin=c]{0}{(a)}}}%
    \includegraphics[width=\dimexpr\linewidth-20pt\relax]
    {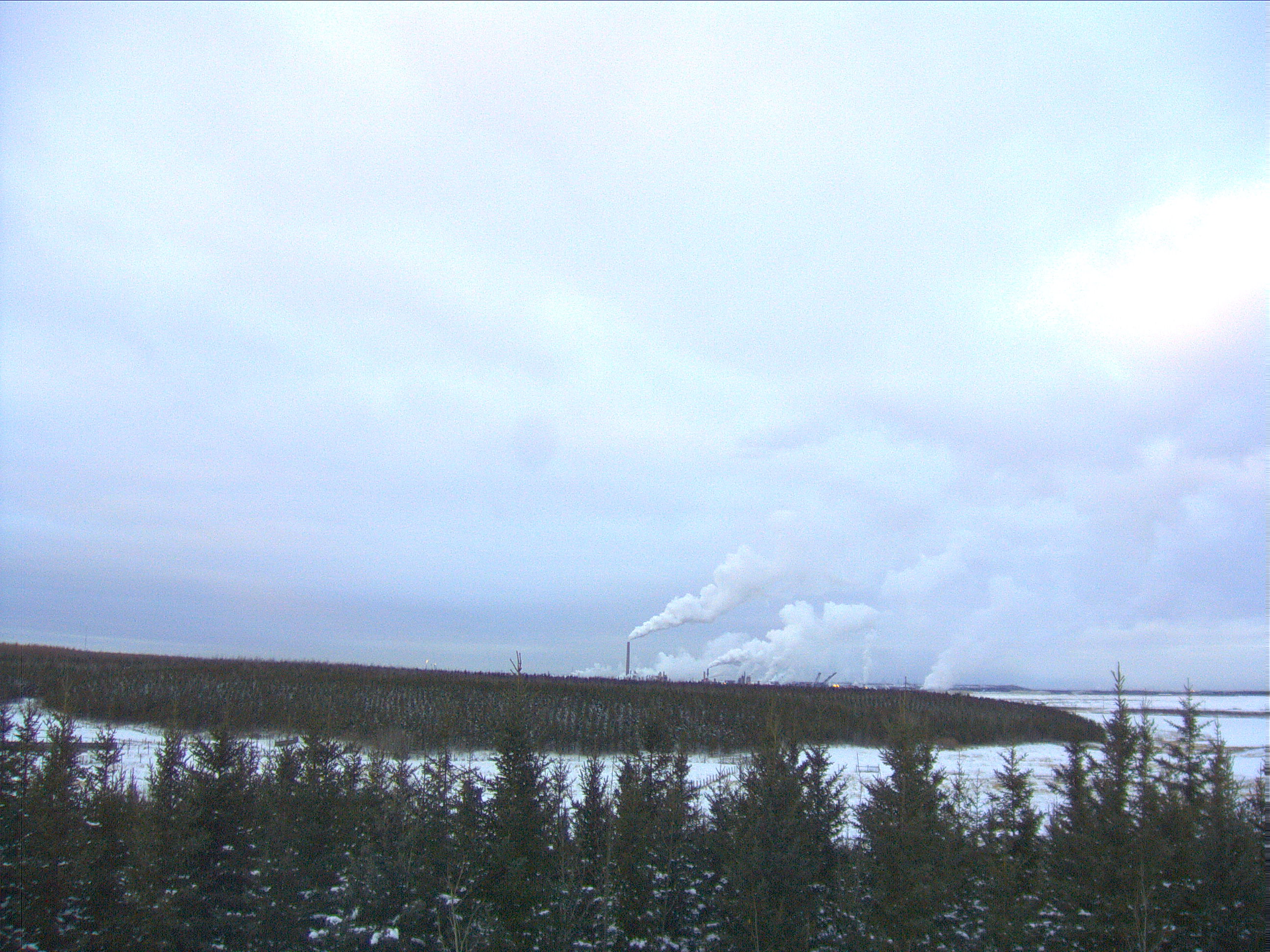}
    \makebox[20pt]{\raisebox{30pt}{\rotatebox[origin=c]{0}{(b)}}}%
    \includegraphics[width=\dimexpr\linewidth-20pt\relax]
    {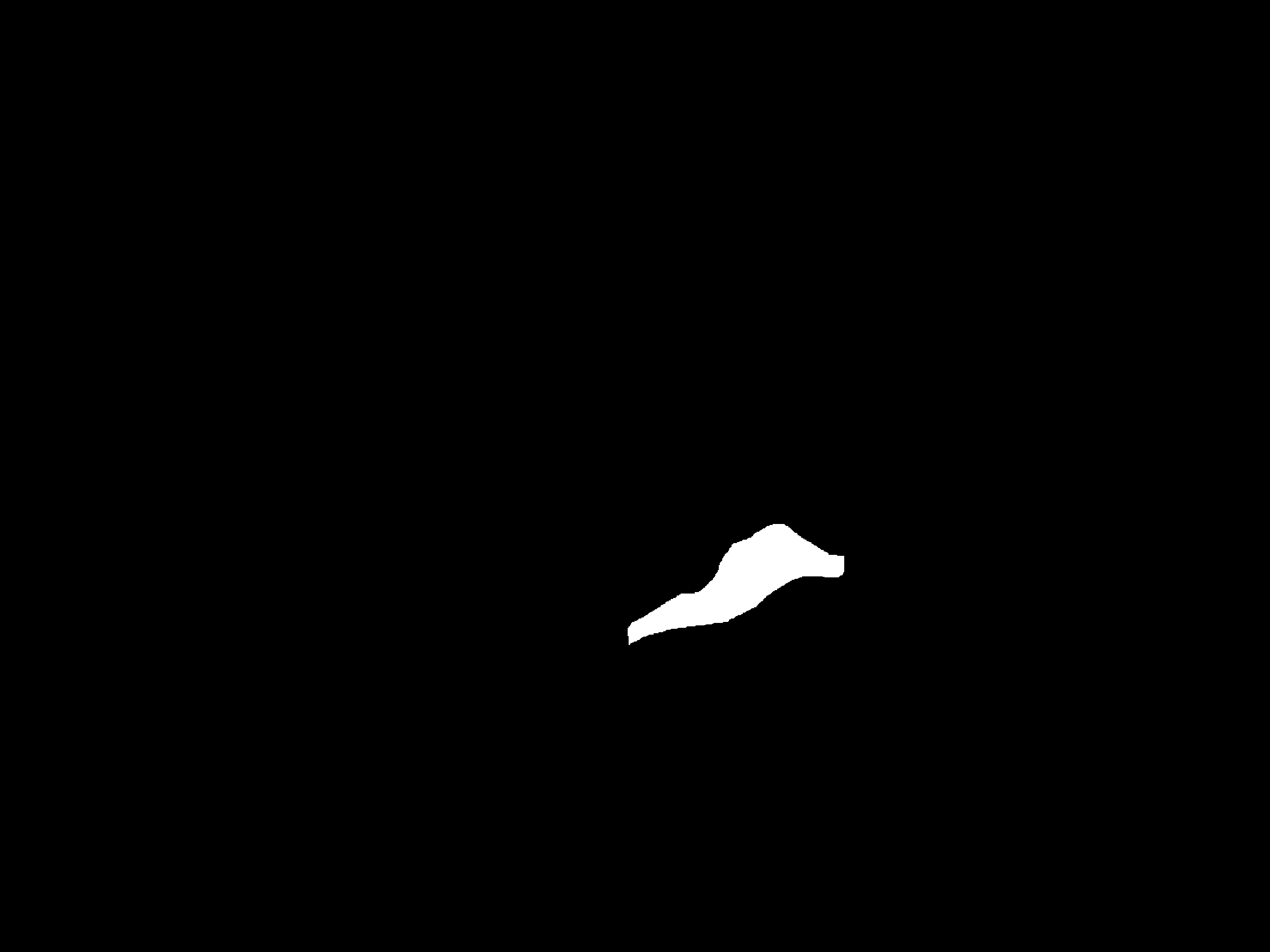}
    \makebox[20pt]{\raisebox{30pt}{\rotatebox[origin=c]{0}{(c)}}}%
    \includegraphics[width=\dimexpr\linewidth-20pt\relax]
    {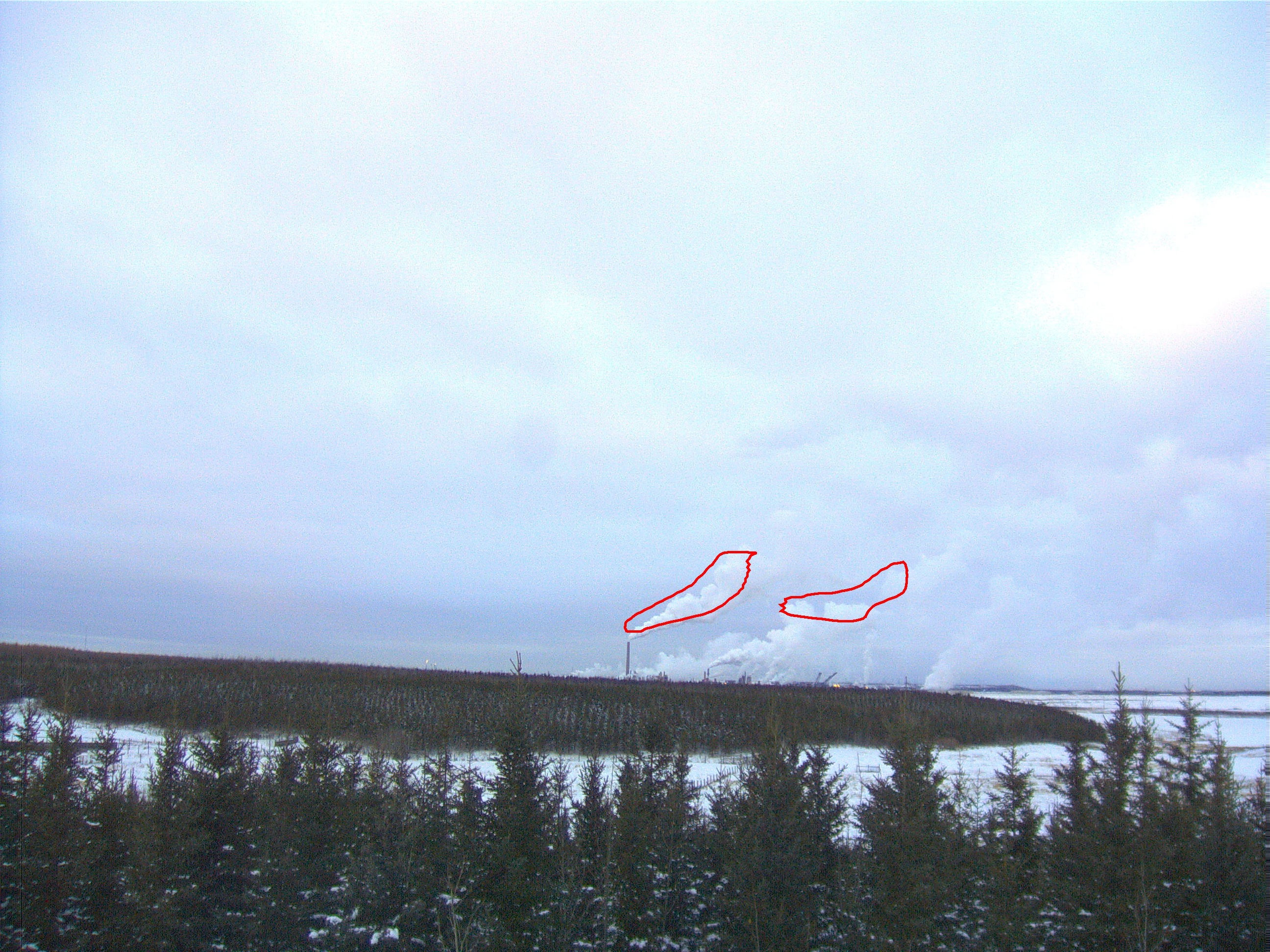}
    \makebox[20pt]{\raisebox{30pt}{\rotatebox[origin=c]{0}{(d)}}}%
    \includegraphics[width=\dimexpr\linewidth-20pt\relax]
    {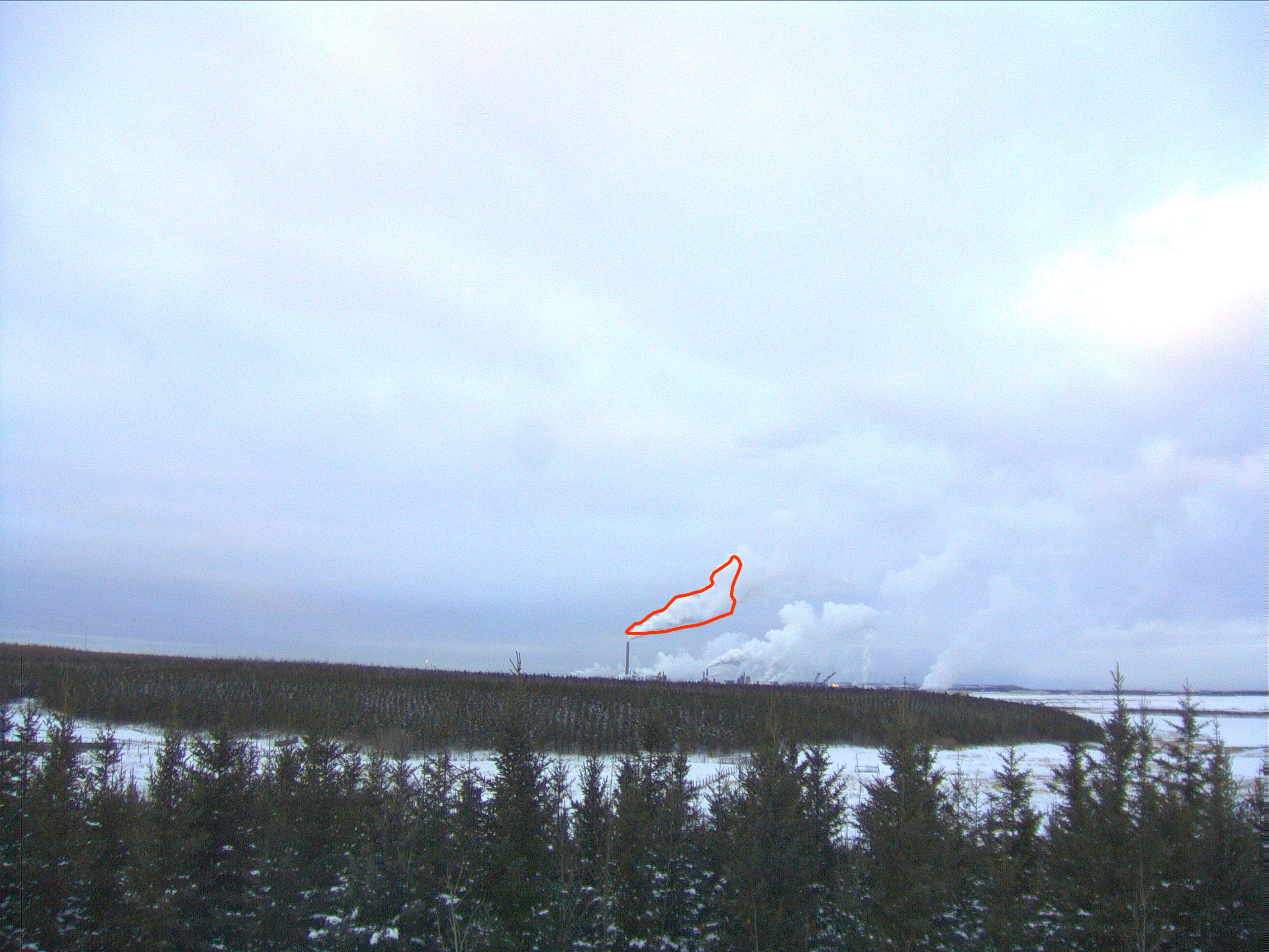}
    \makebox[20pt]{\raisebox{30pt}{\rotatebox[origin=c]{0}{(e)}}}%
    \includegraphics[width=\dimexpr\linewidth-20pt\relax]
    {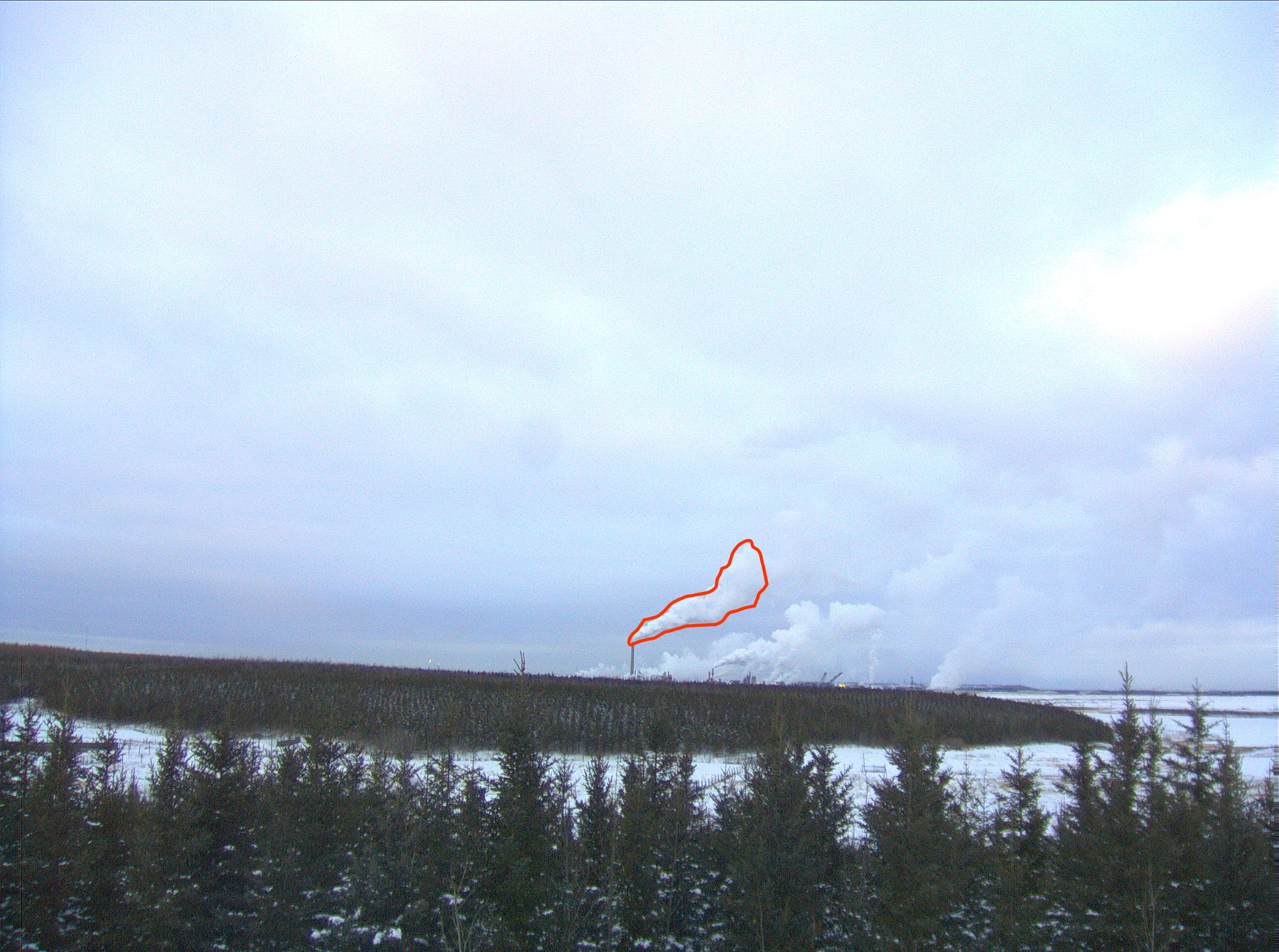}
    \makebox[20pt]{\raisebox{30pt}{\rotatebox[origin=c]{0}{(f)}}}%
    \includegraphics[width=\dimexpr\linewidth-20pt\relax]
    {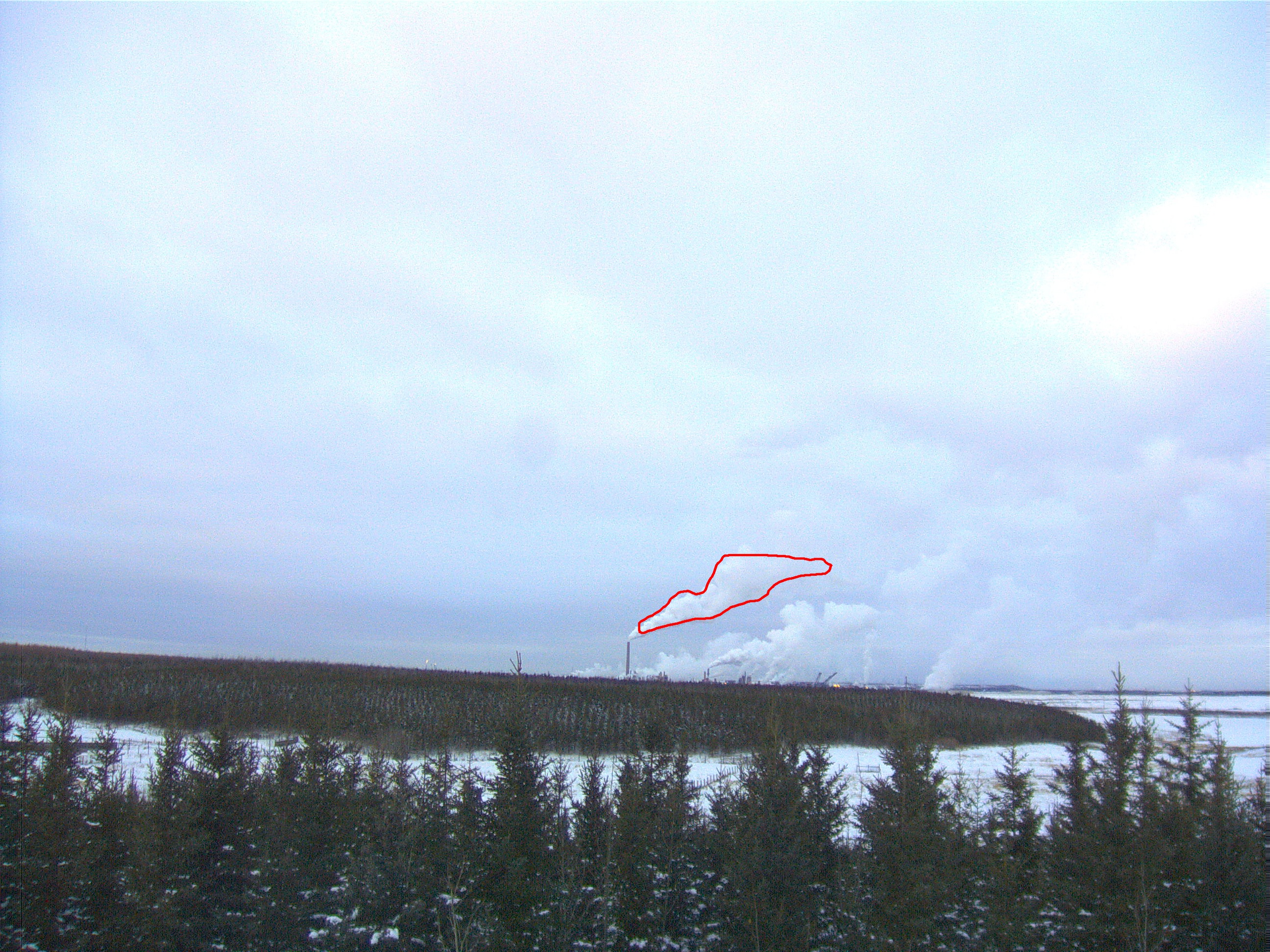}
\end{subfigure}\hfill
\begin{subfigure}[t]{0.15\textwidth}
    \includegraphics[width=\textwidth]
    {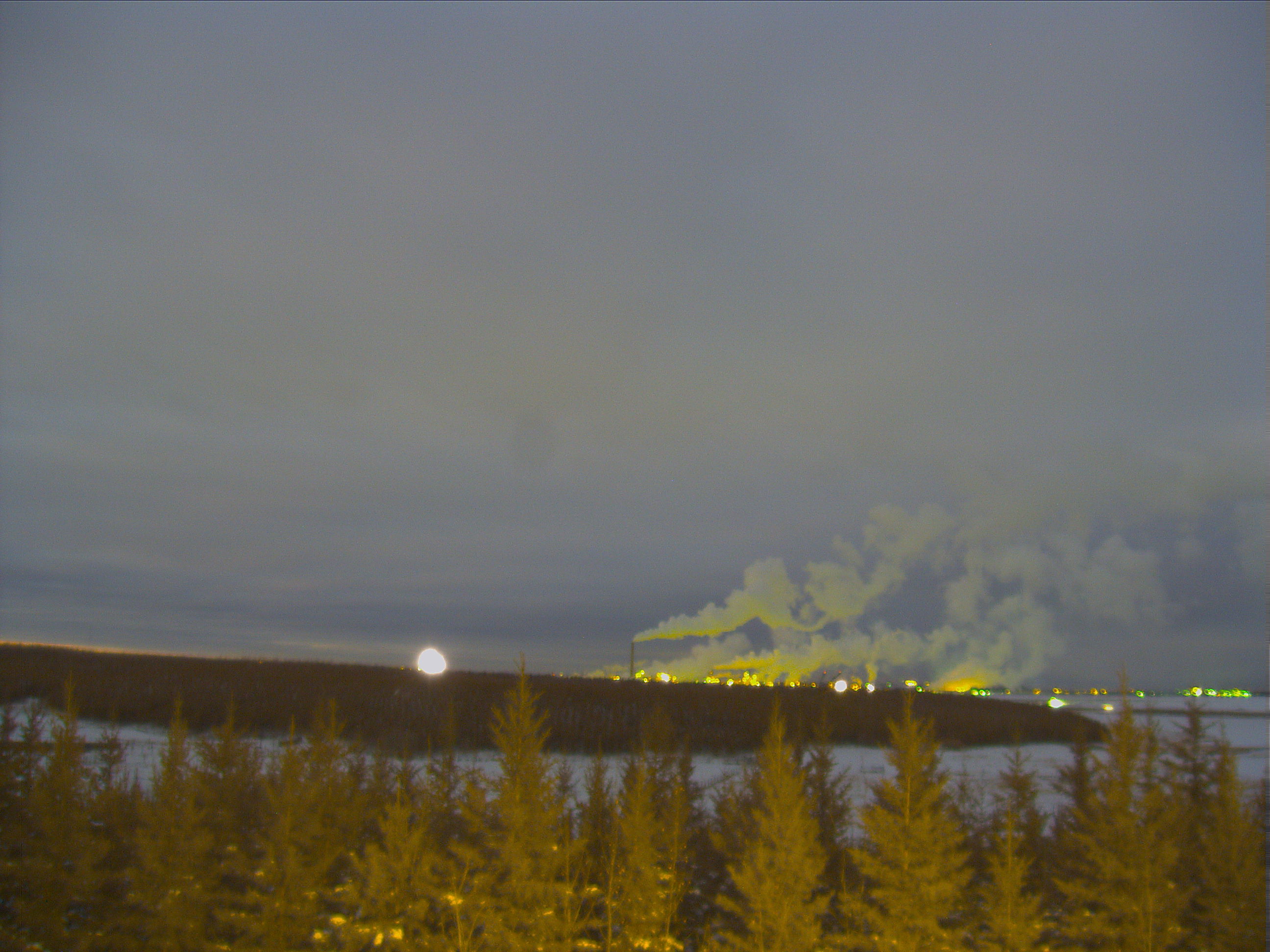}
    \includegraphics[width=\textwidth]
    {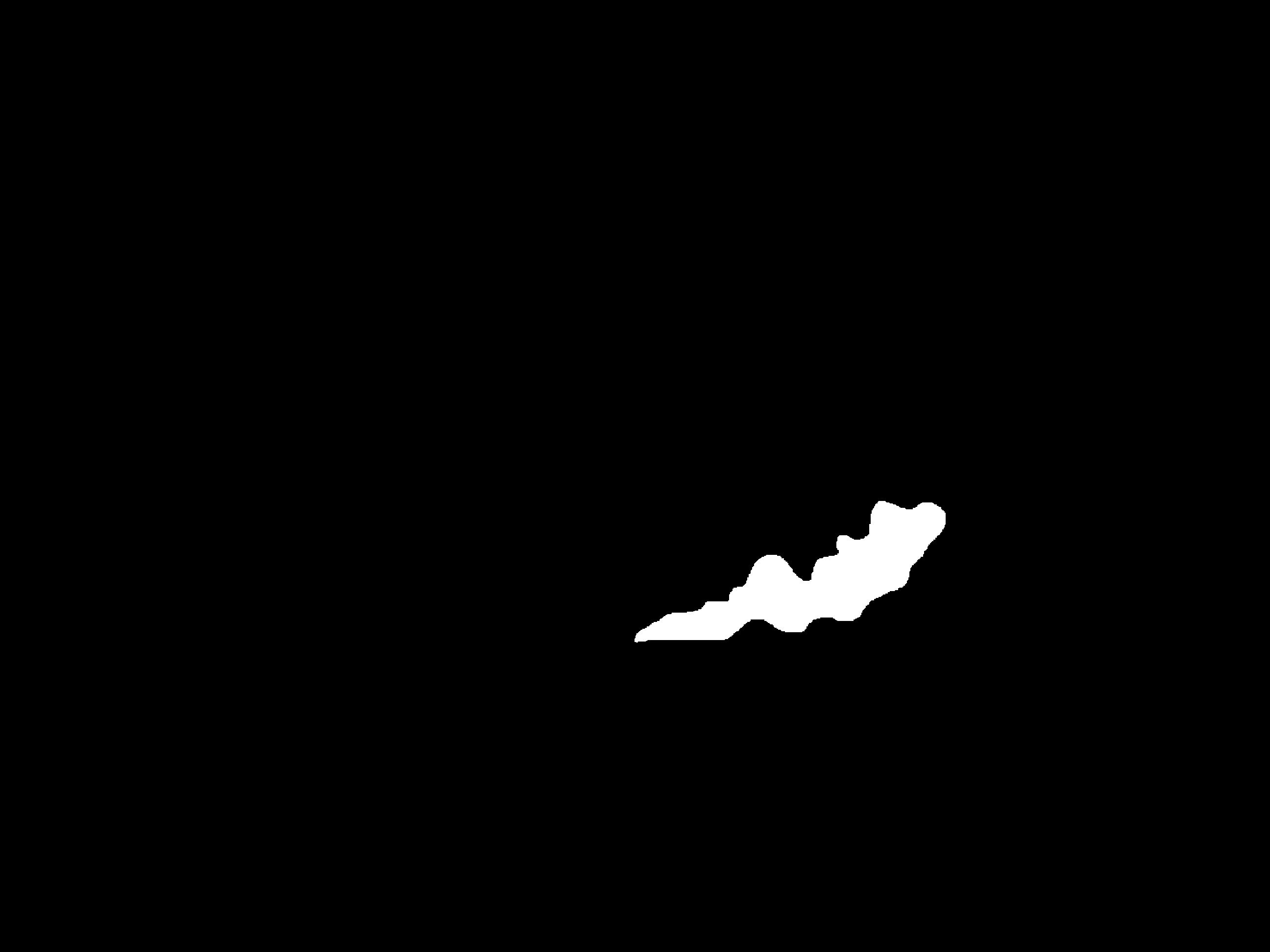}
    \includegraphics[width=\textwidth]
    {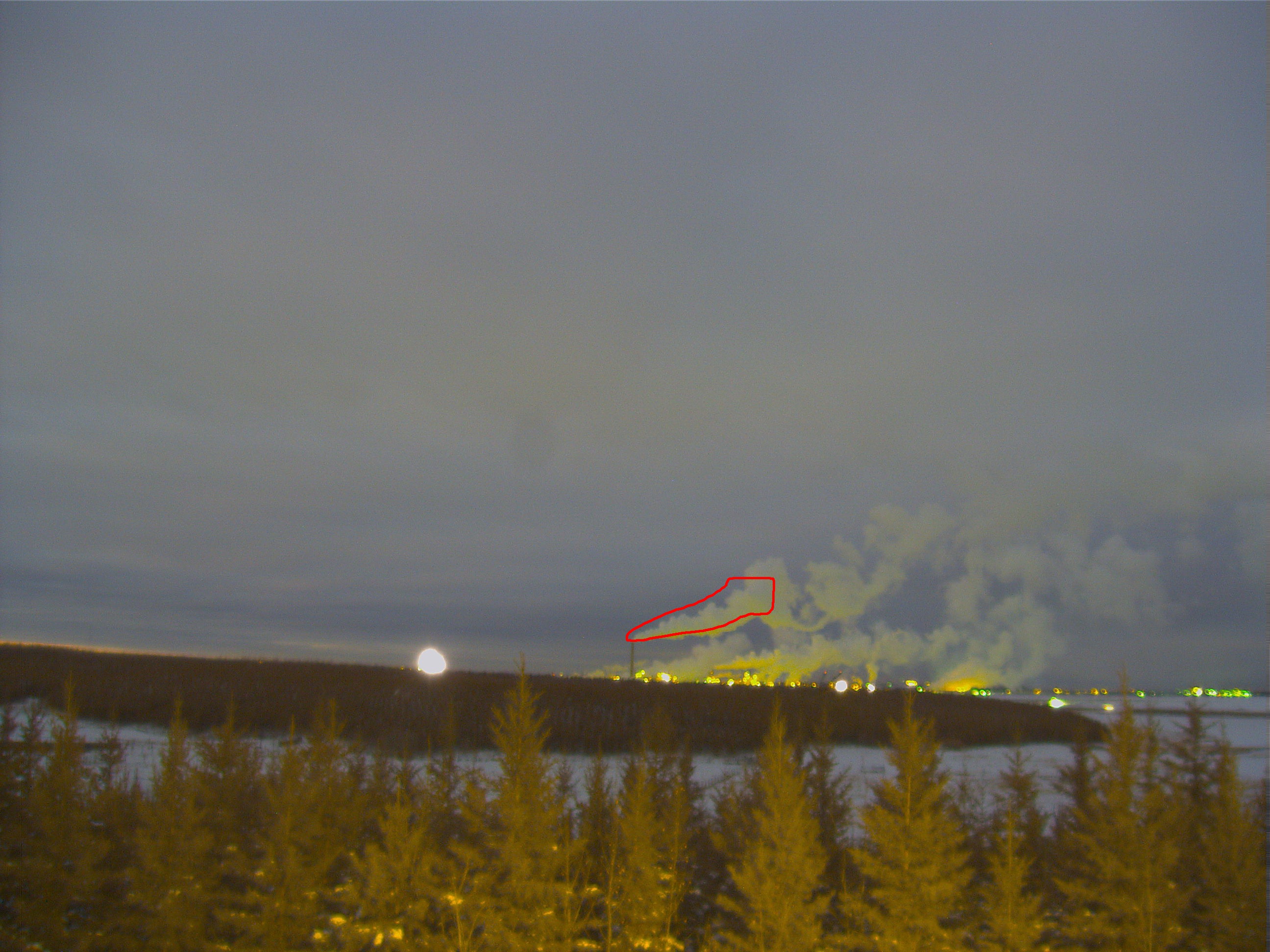}
    \includegraphics[width=\textwidth]
    {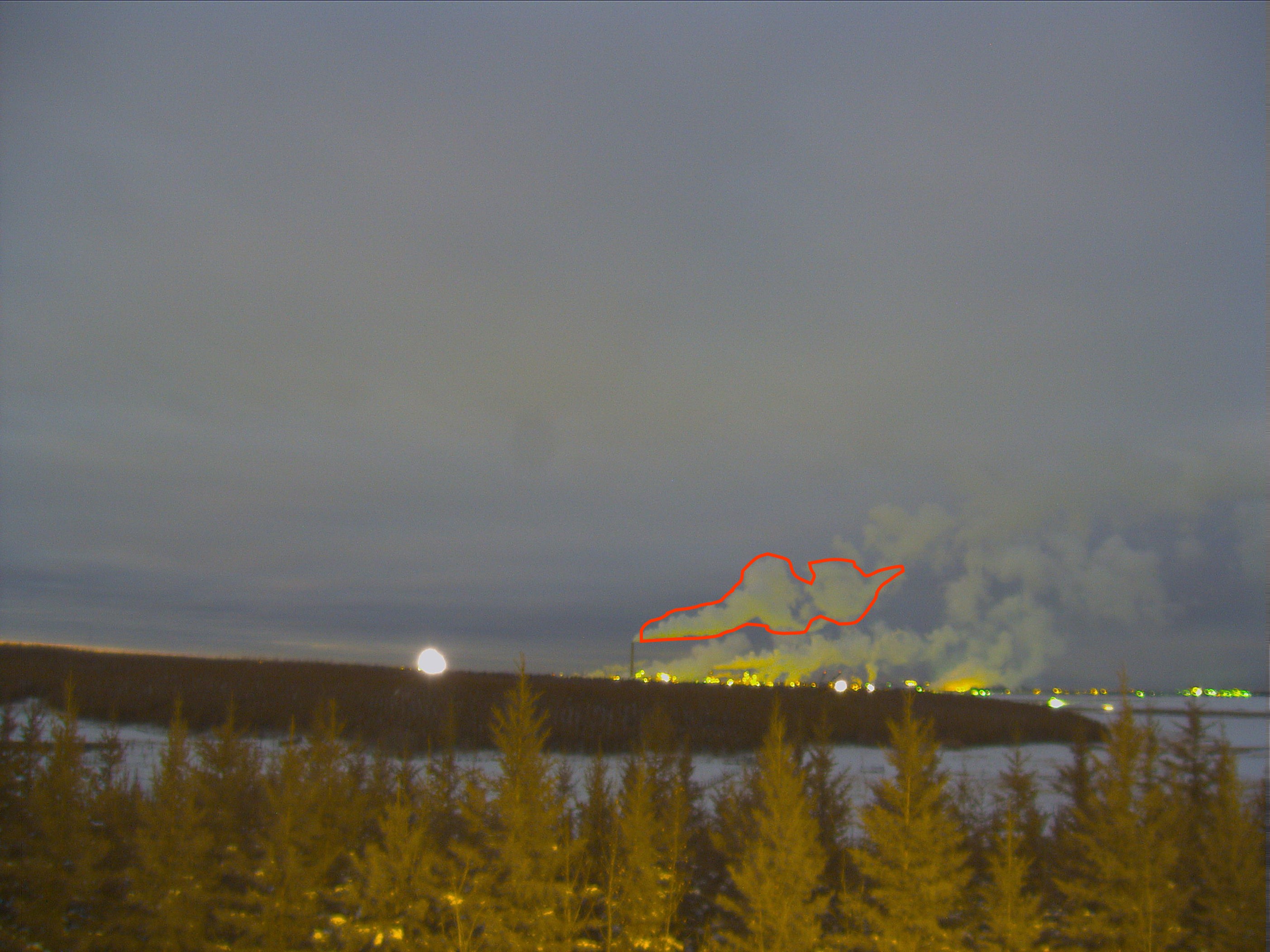}
    \includegraphics[width=\textwidth]
    {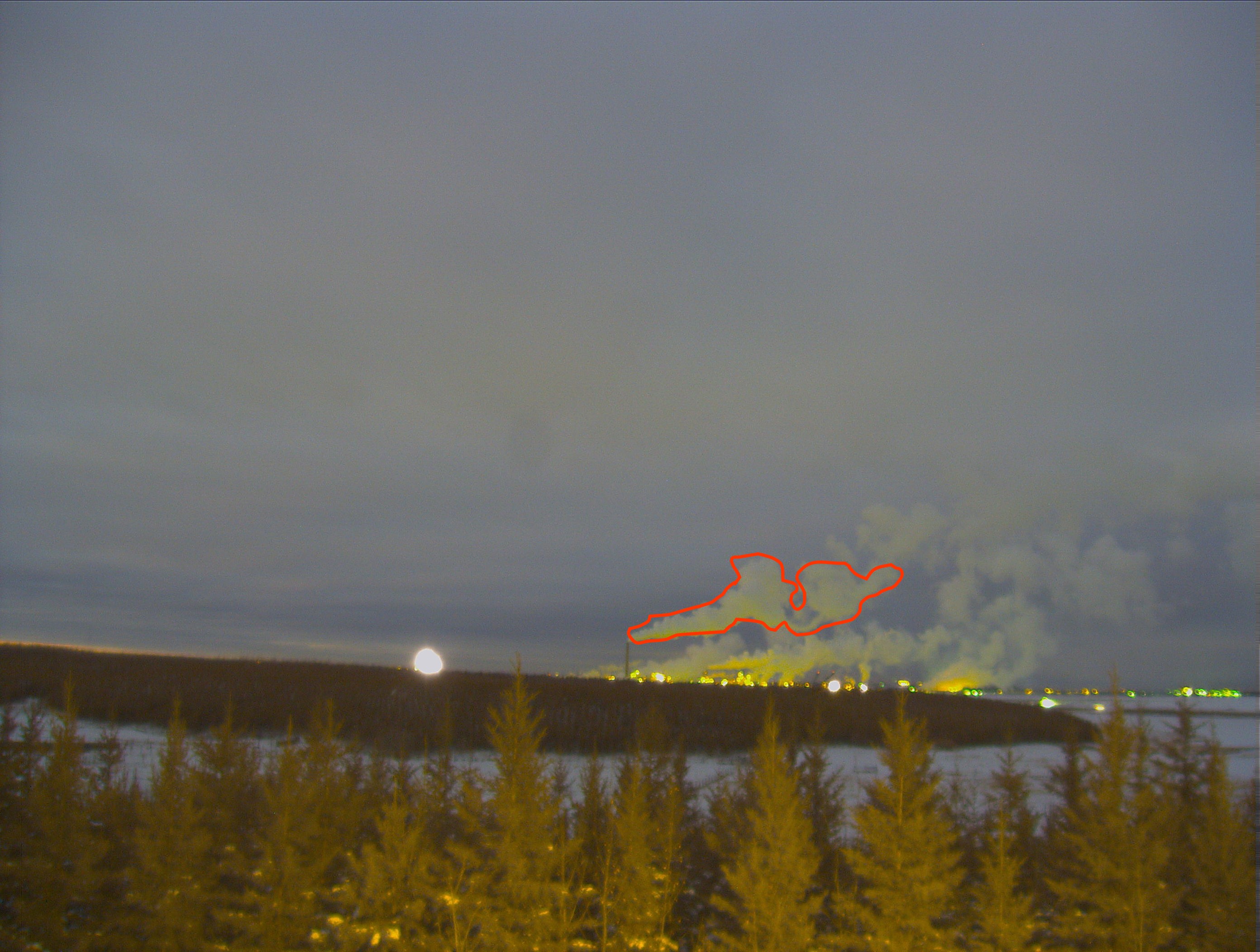}
    \includegraphics[width=\textwidth]
    {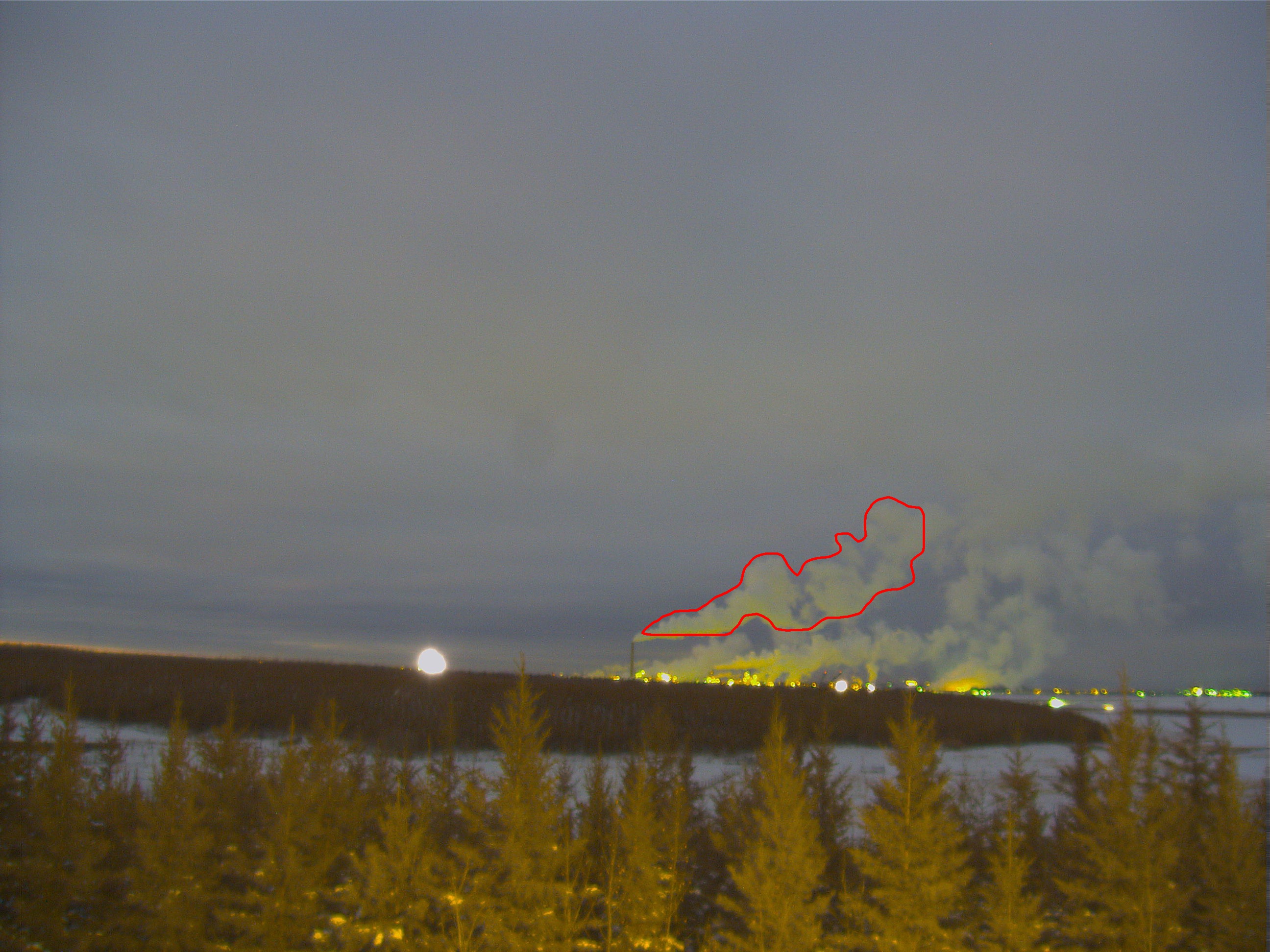}
\end{subfigure}\hfill
\begin{subfigure}[t]{0.15\textwidth}
    \includegraphics[width=\textwidth]
    {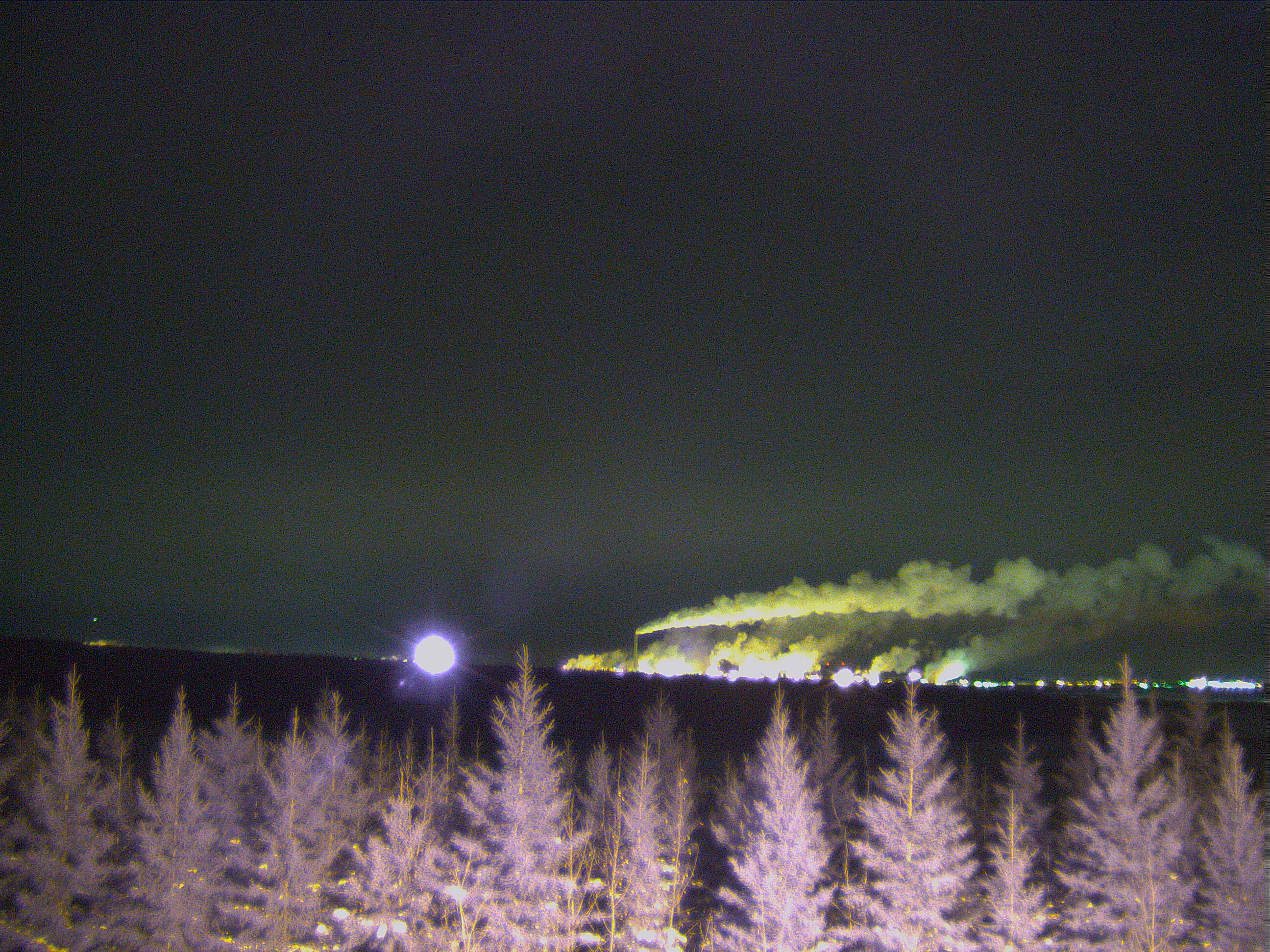}
    \includegraphics[width=\textwidth]
    {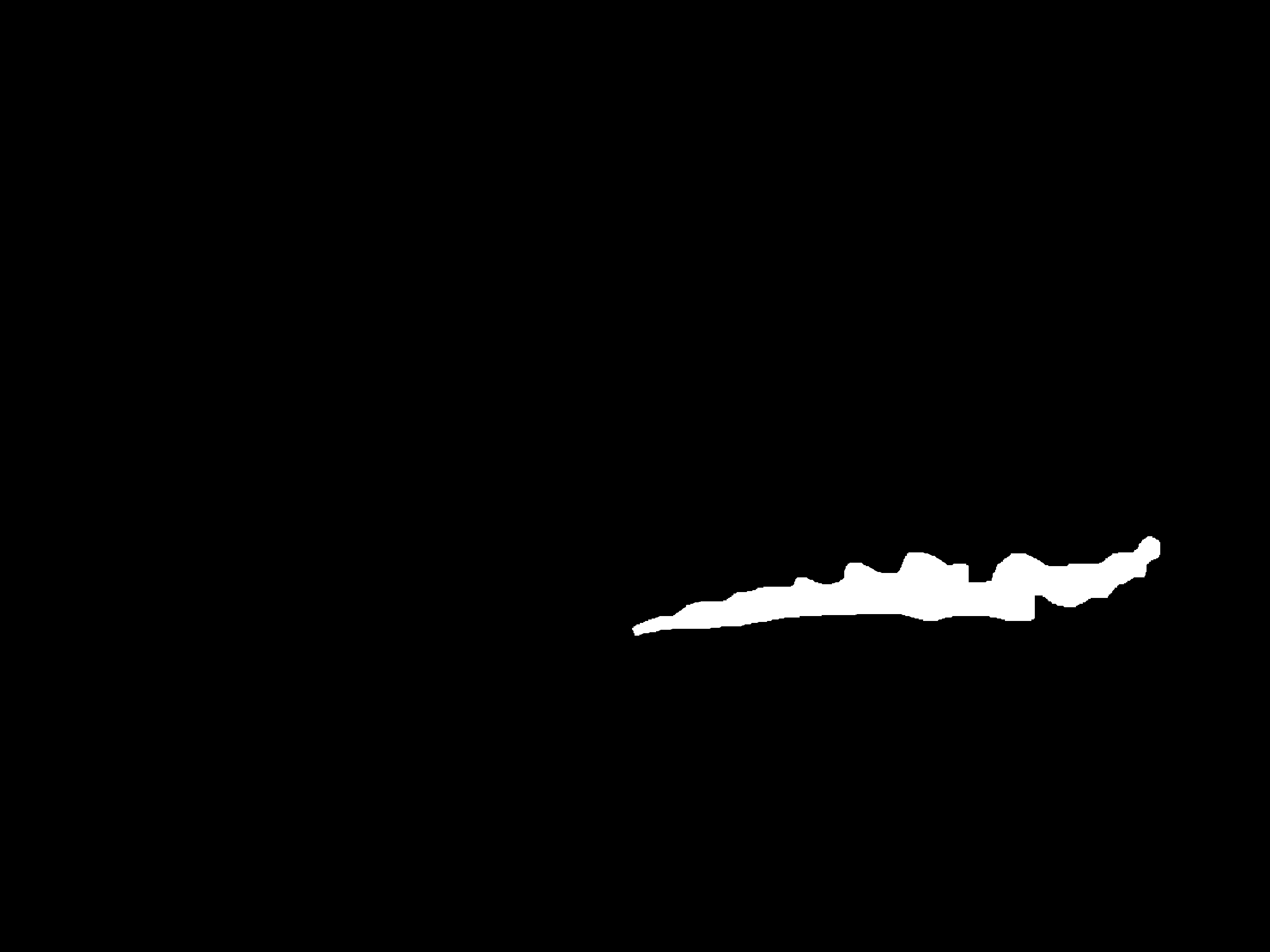}
    \includegraphics[width=\textwidth]
    {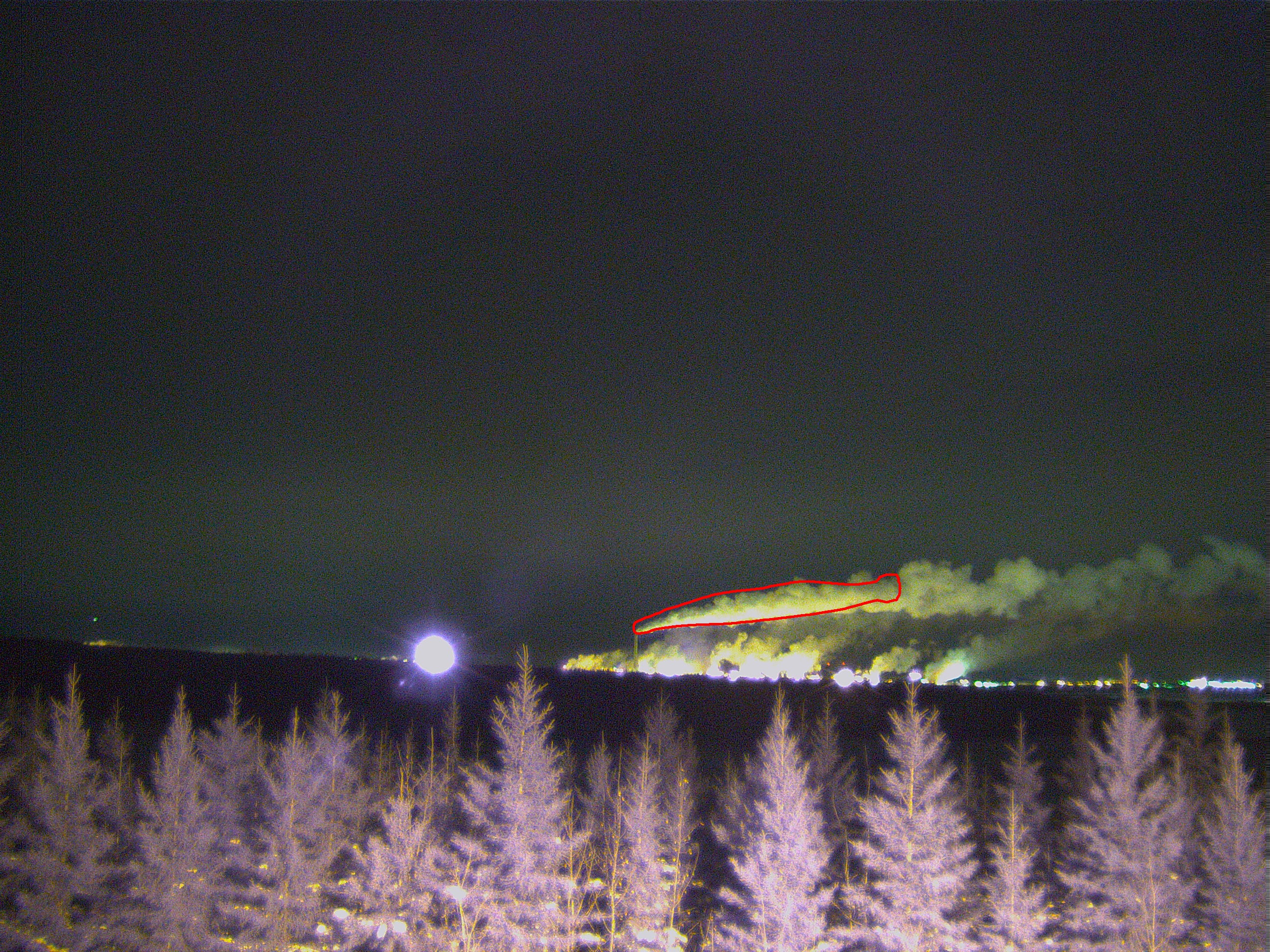}
    \includegraphics[width=\textwidth]
    {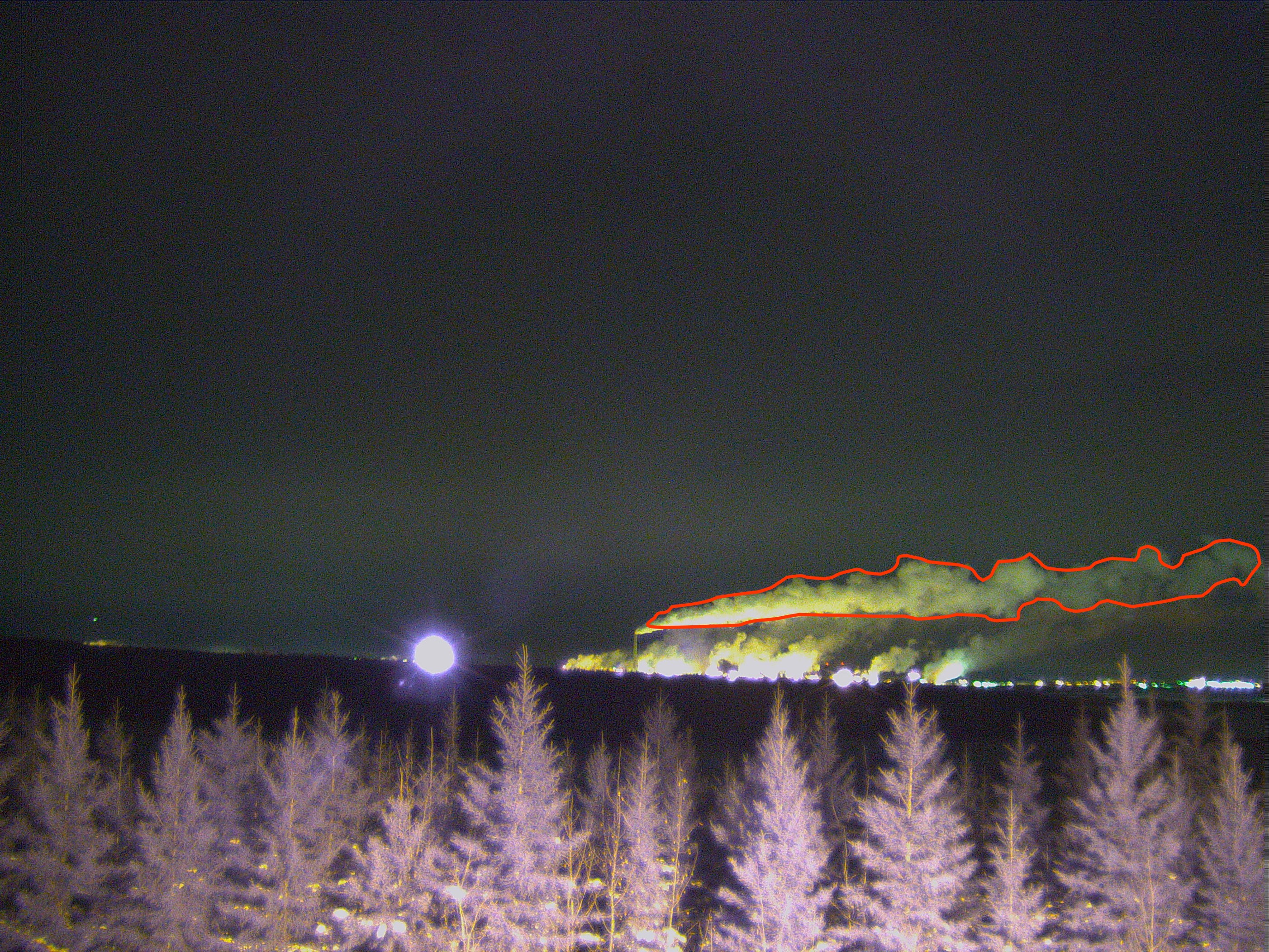}
    \includegraphics[width=\textwidth]
    {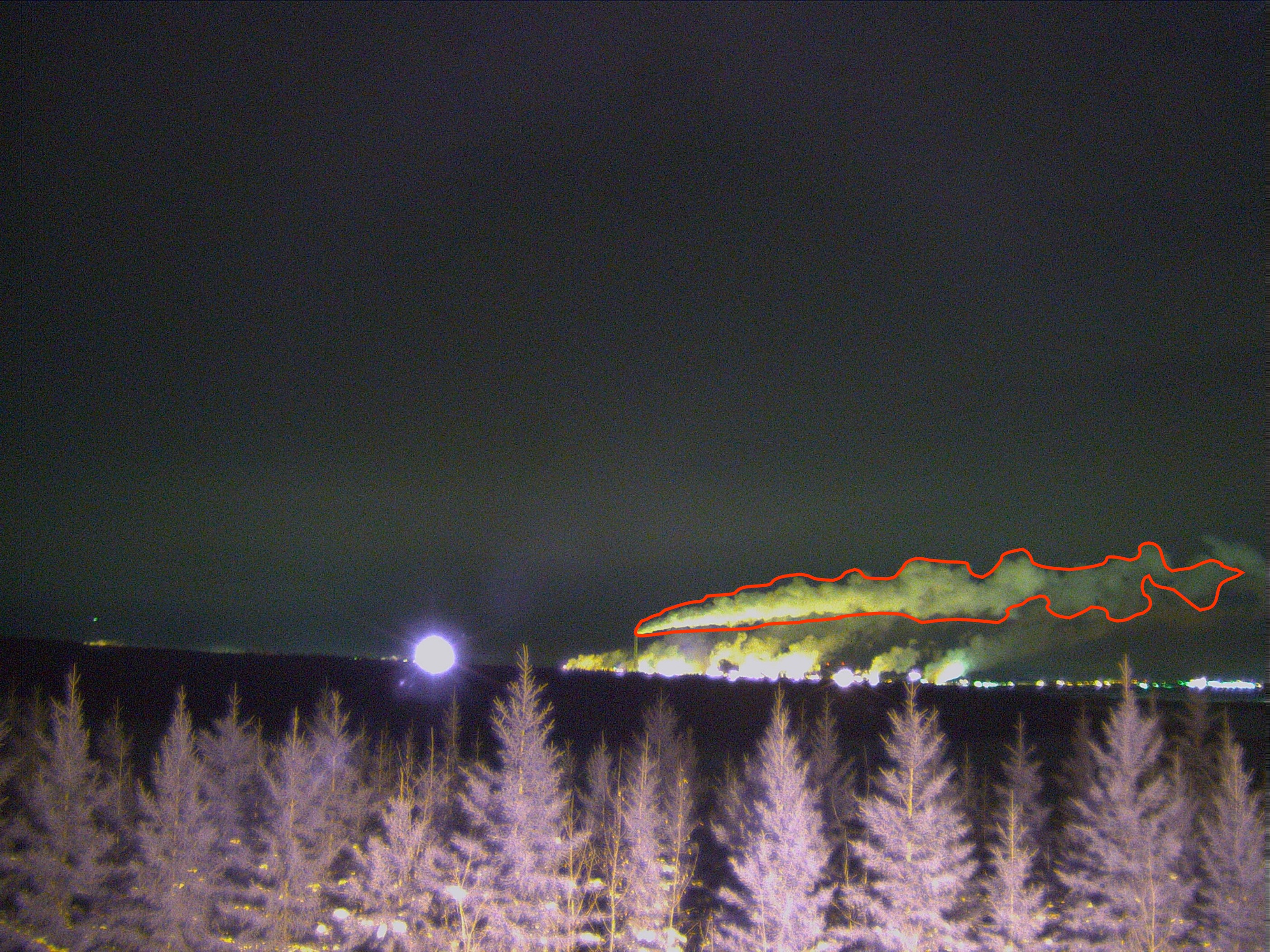}
    \includegraphics[width=\textwidth]
    {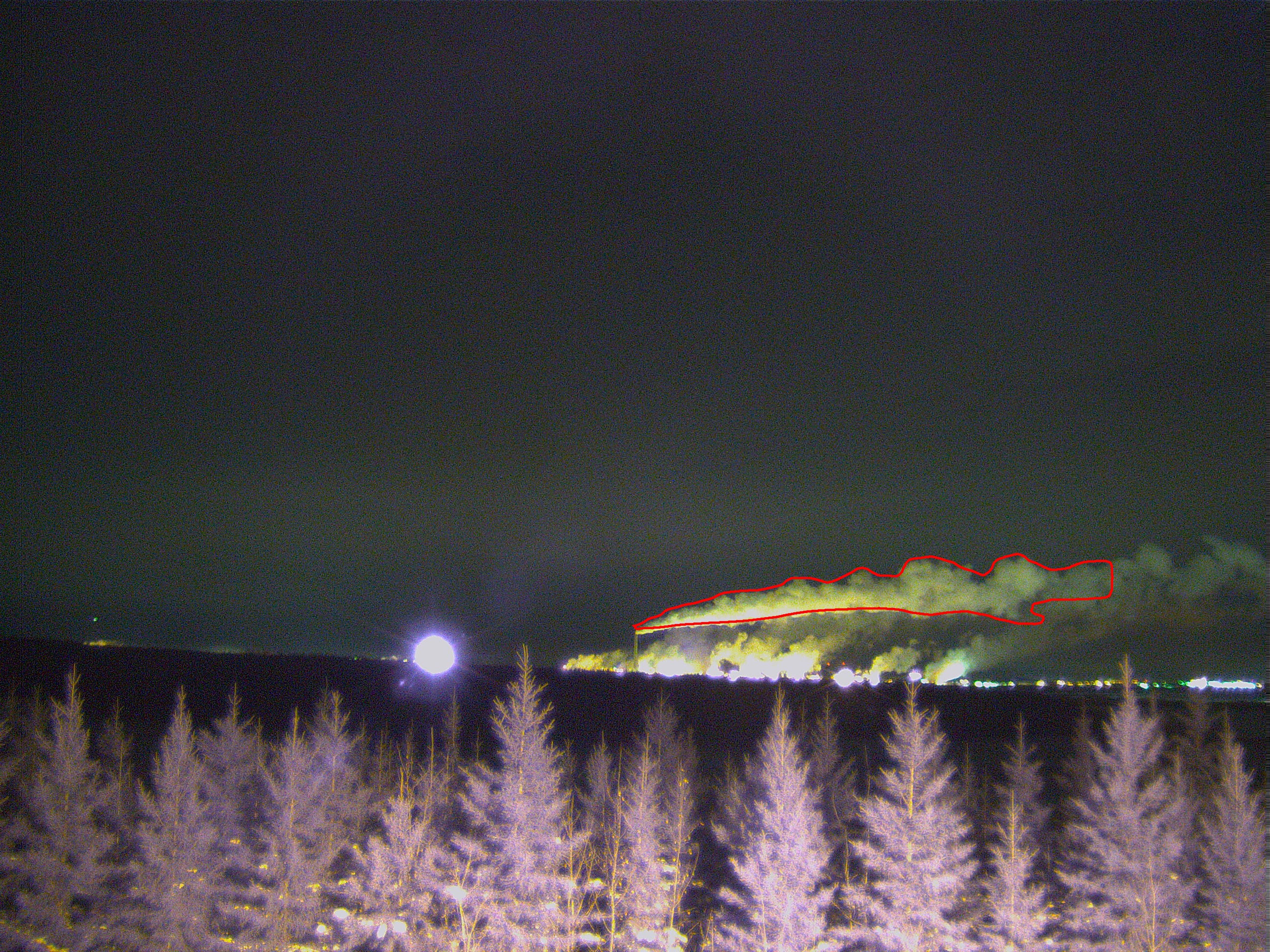}
\end{subfigure}\hfill
\begin{subfigure}[t]{0.15\textwidth}
    \includegraphics[width=\textwidth]
    {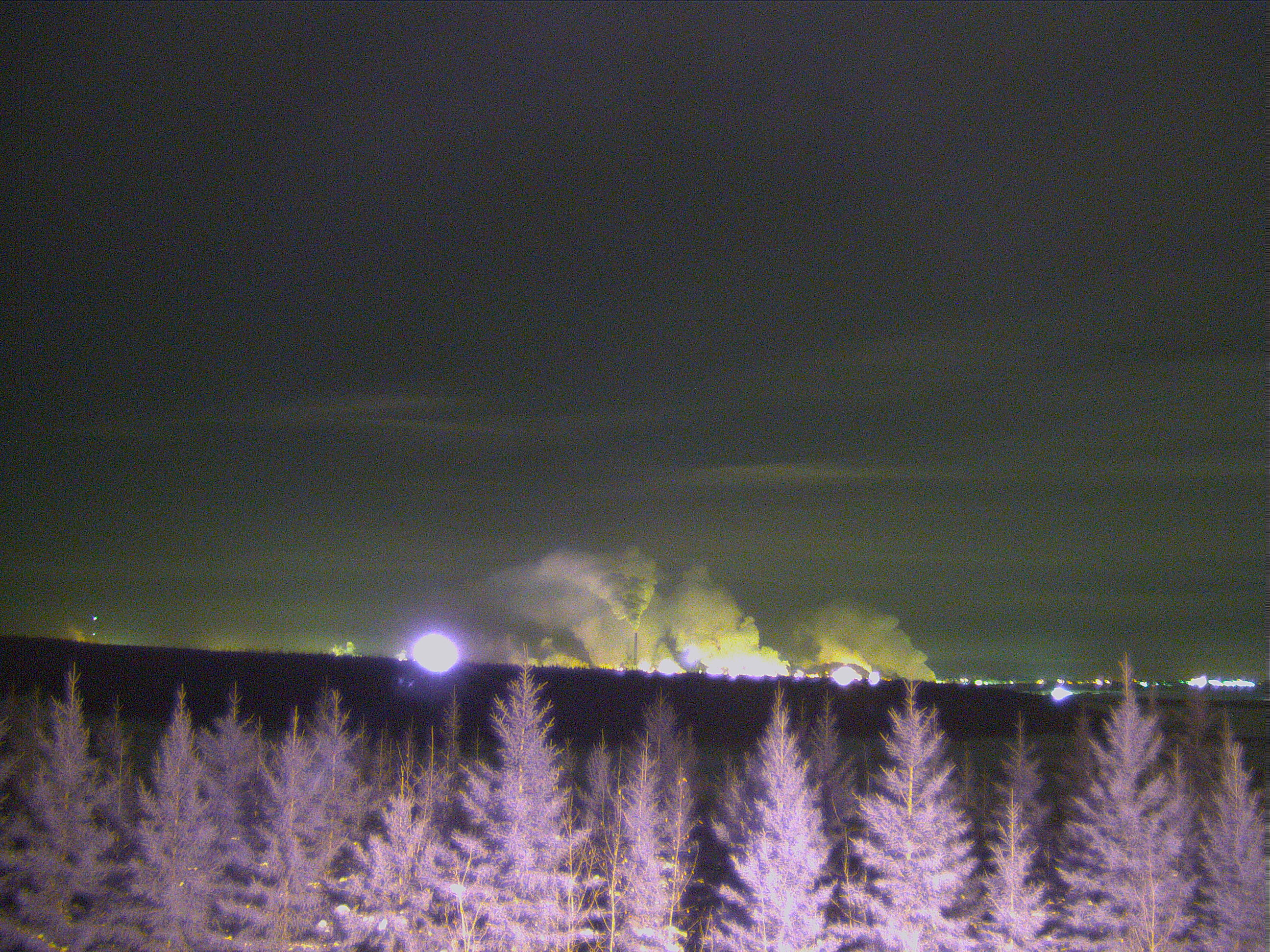}
    \includegraphics[width=\textwidth]
    {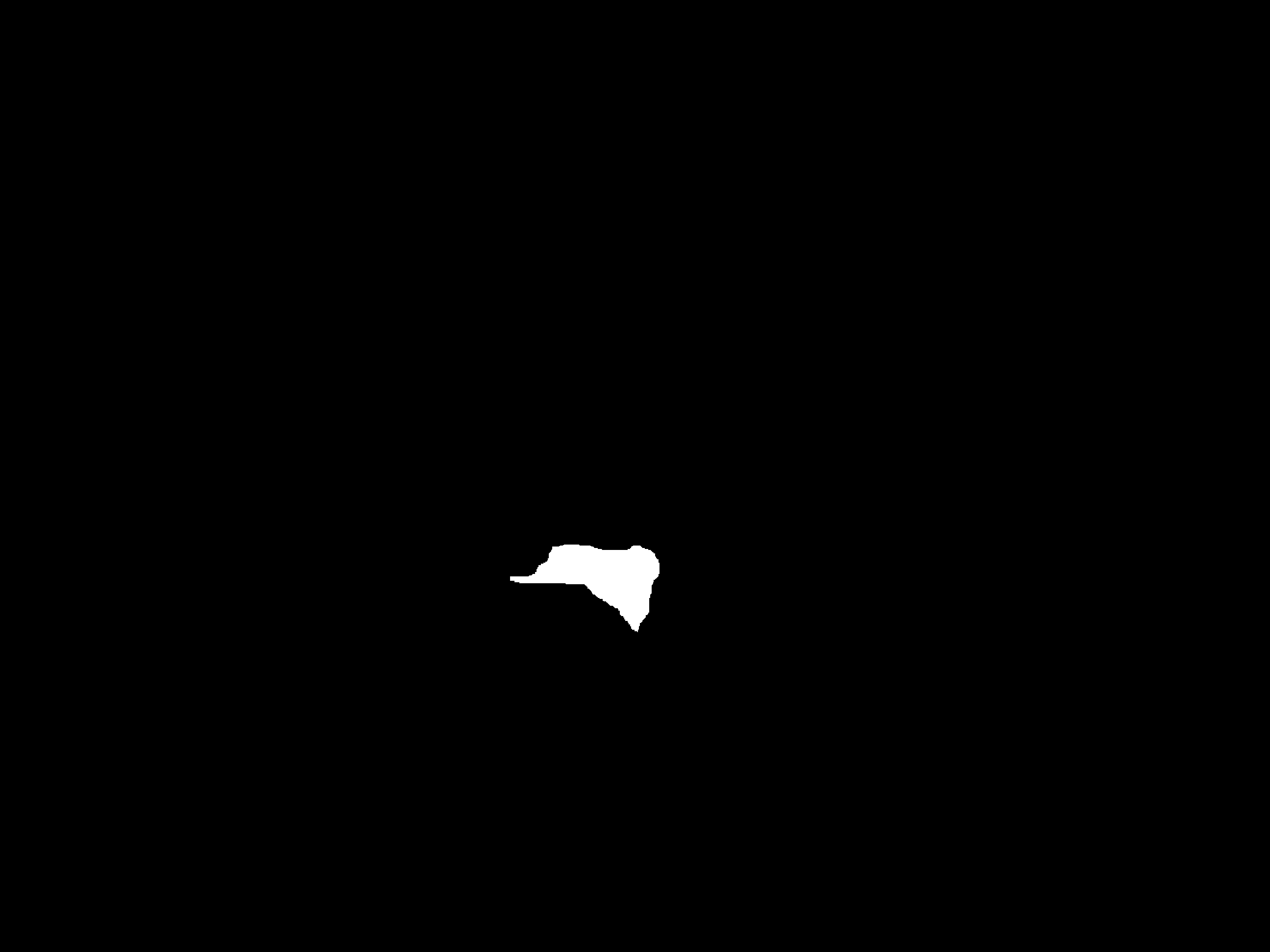}
    \includegraphics[width=\textwidth]
    {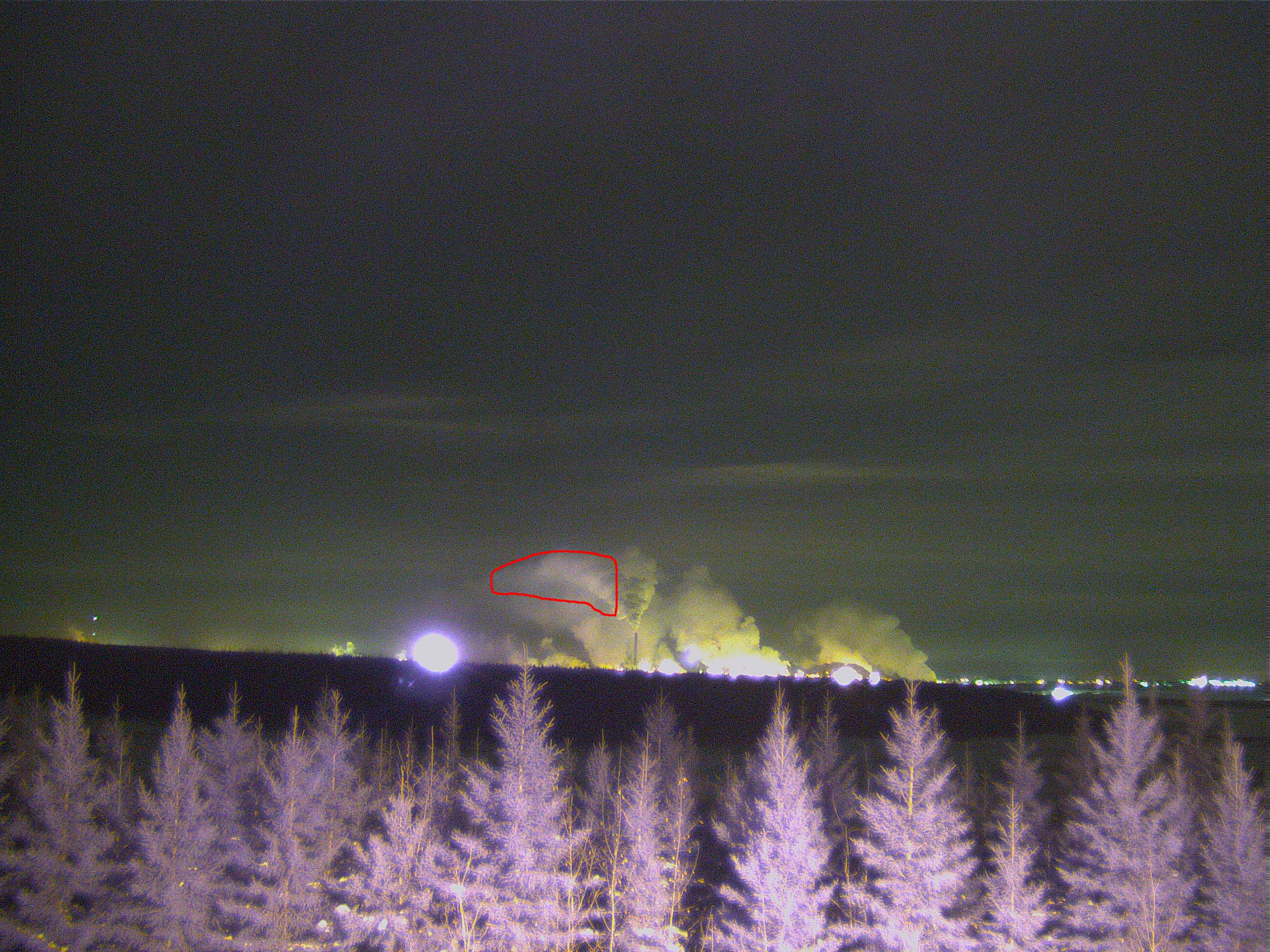}
    \includegraphics[width=\textwidth]
    {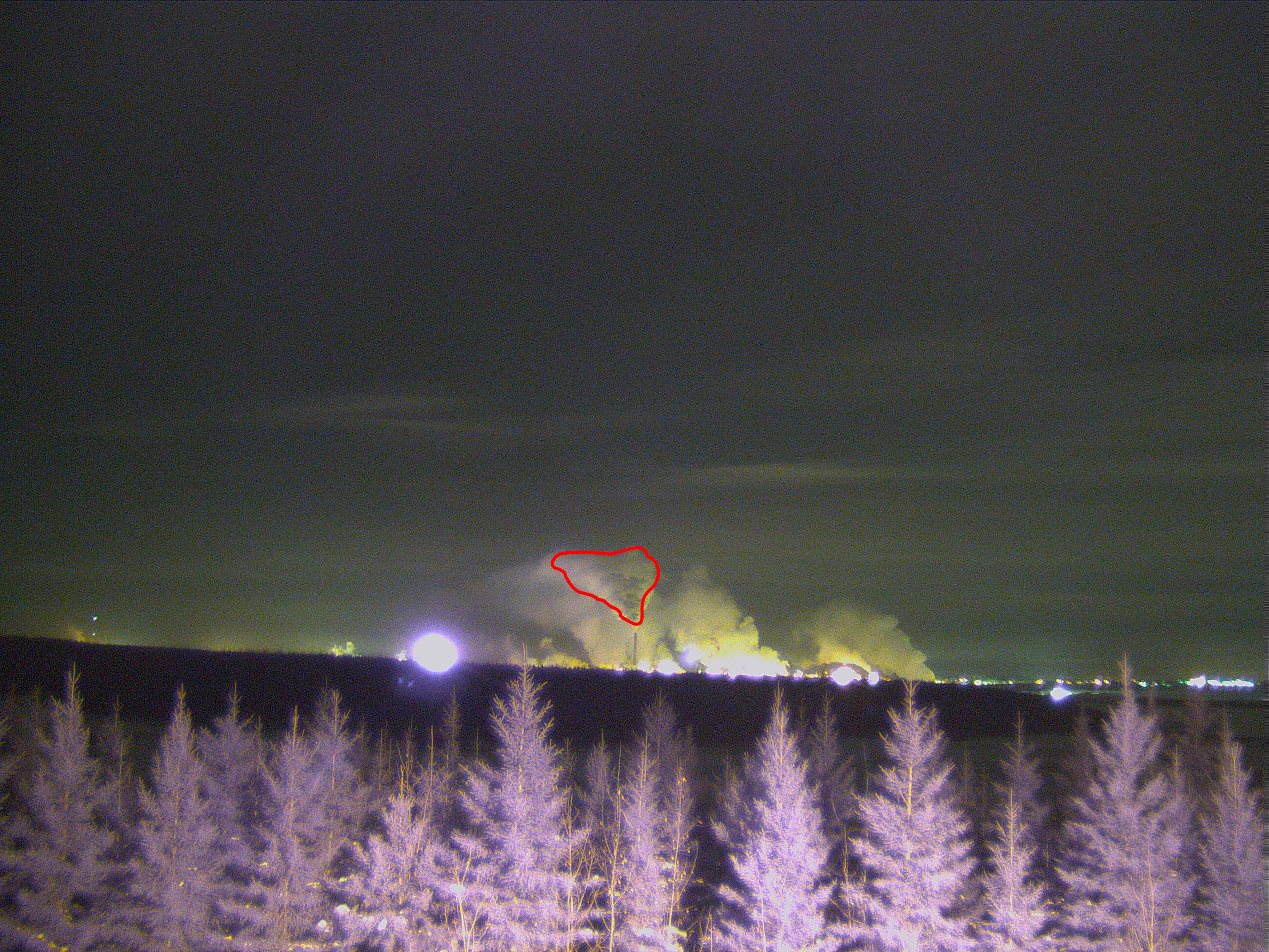}
    \includegraphics[width=\textwidth]
    {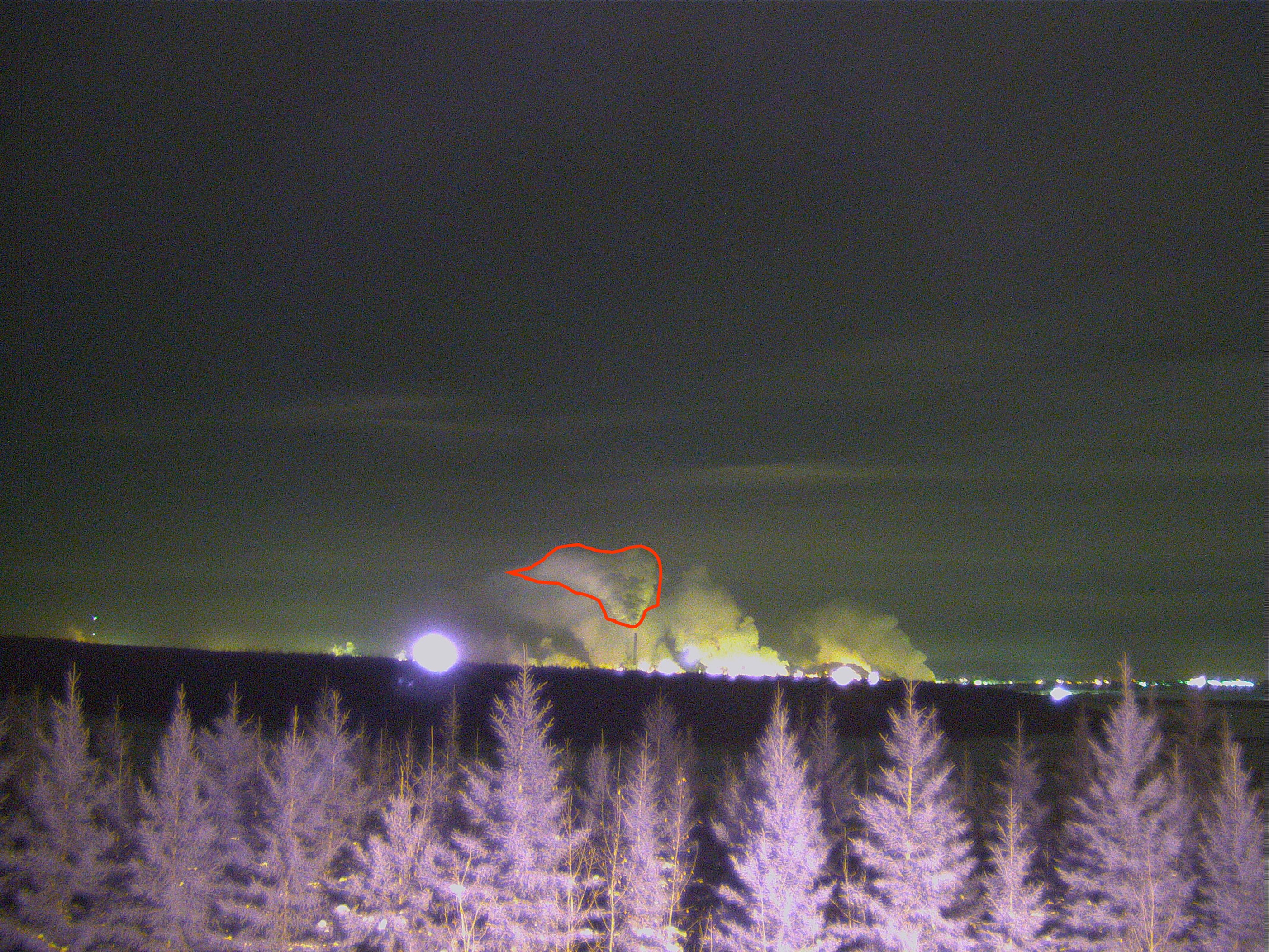}
    \includegraphics[width=\textwidth]
    {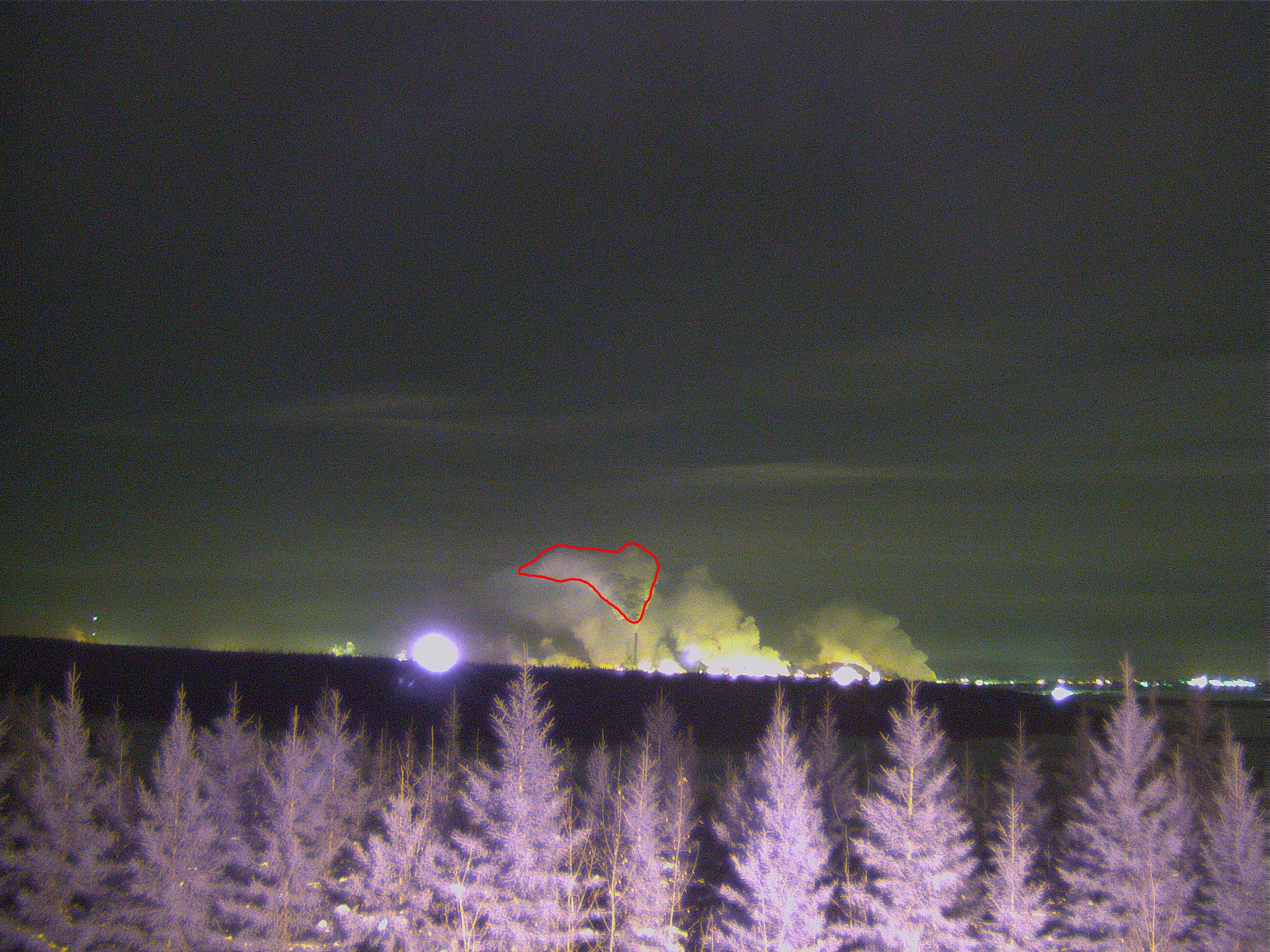}
\end{subfigure}\hfill
\begin{subfigure}[t]{0.15\textwidth}
    \includegraphics[width=\textwidth]
    {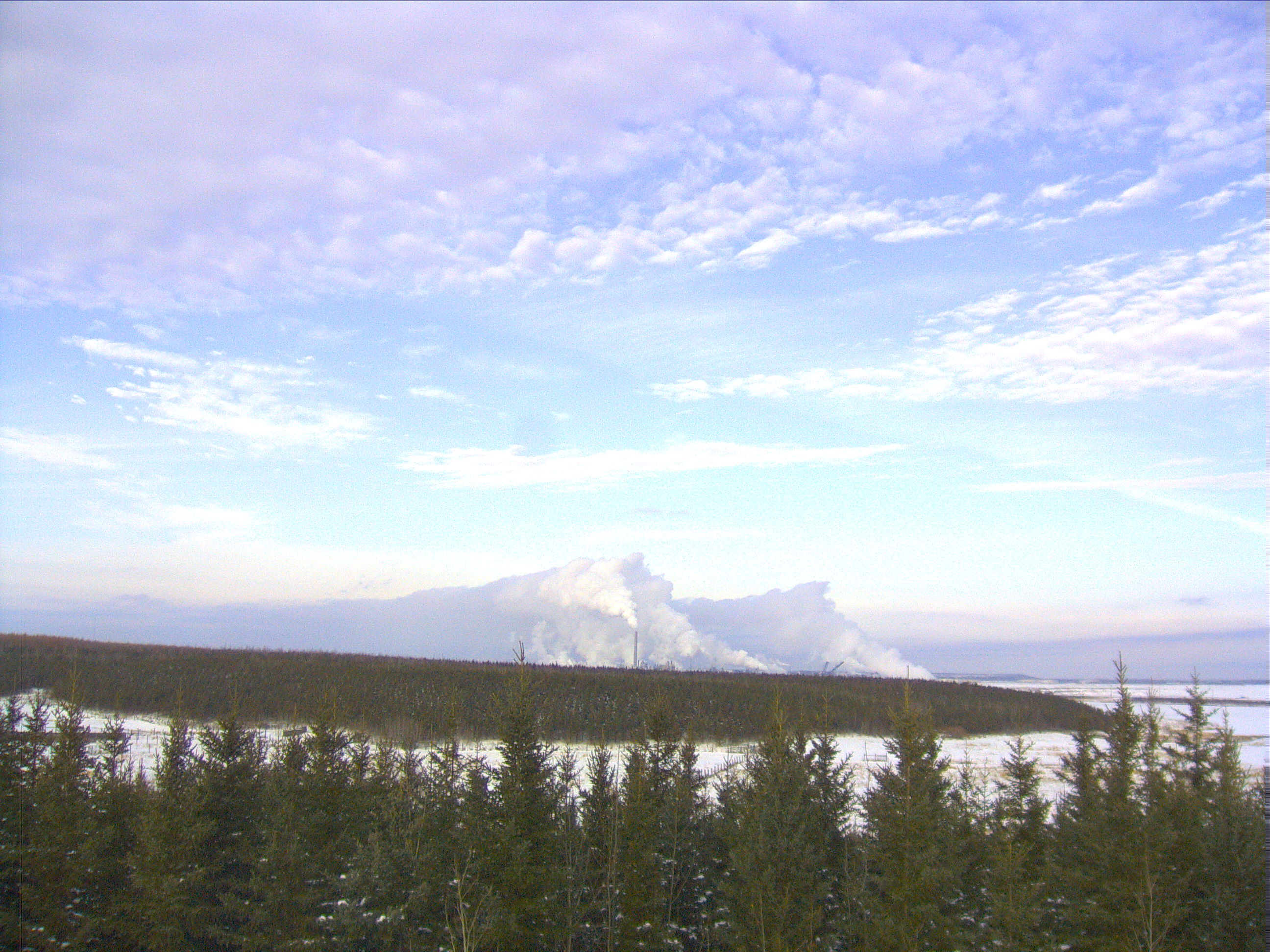}
    \includegraphics[width=\textwidth]
    {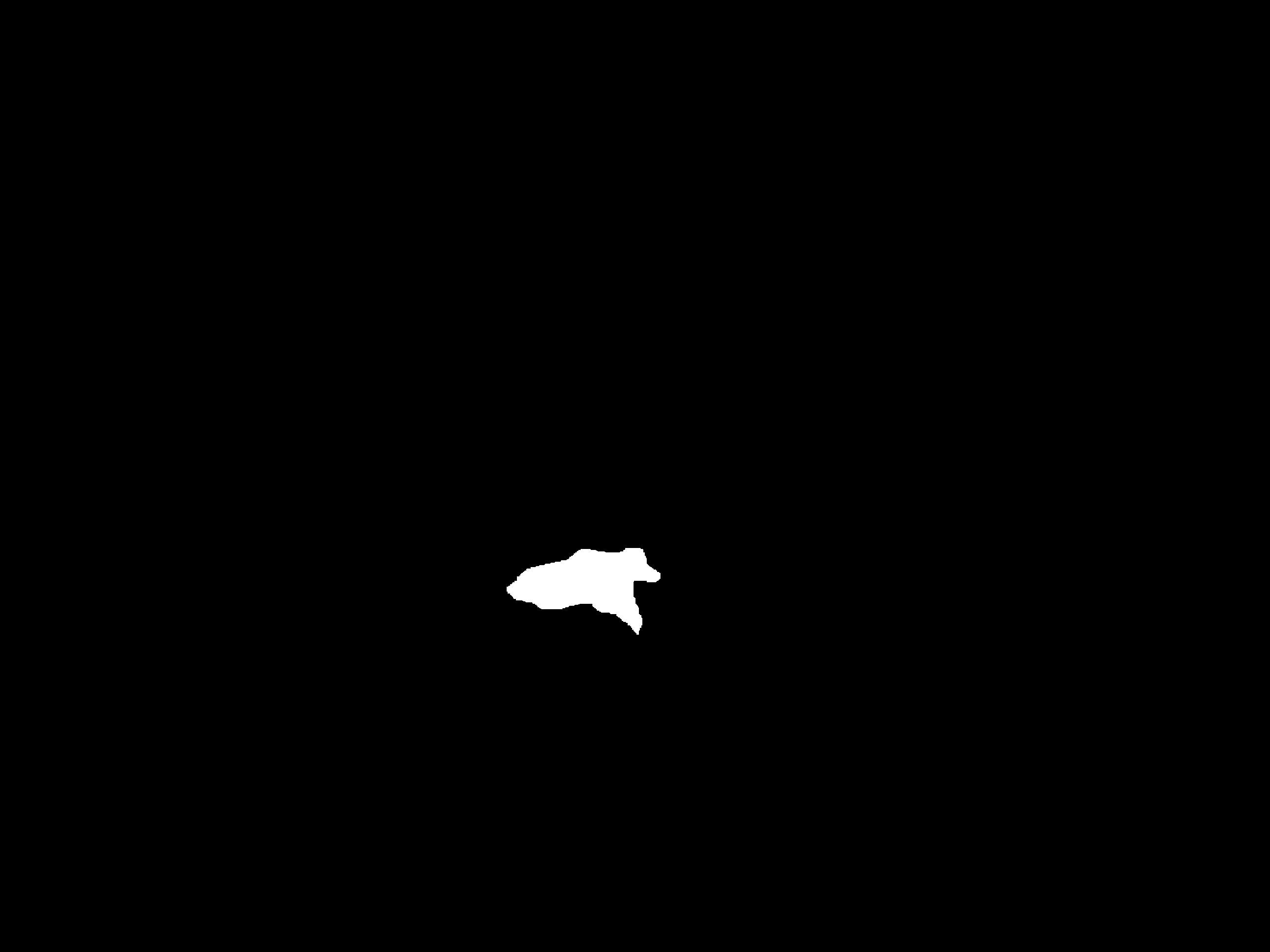}
    \includegraphics[width=\textwidth]
    {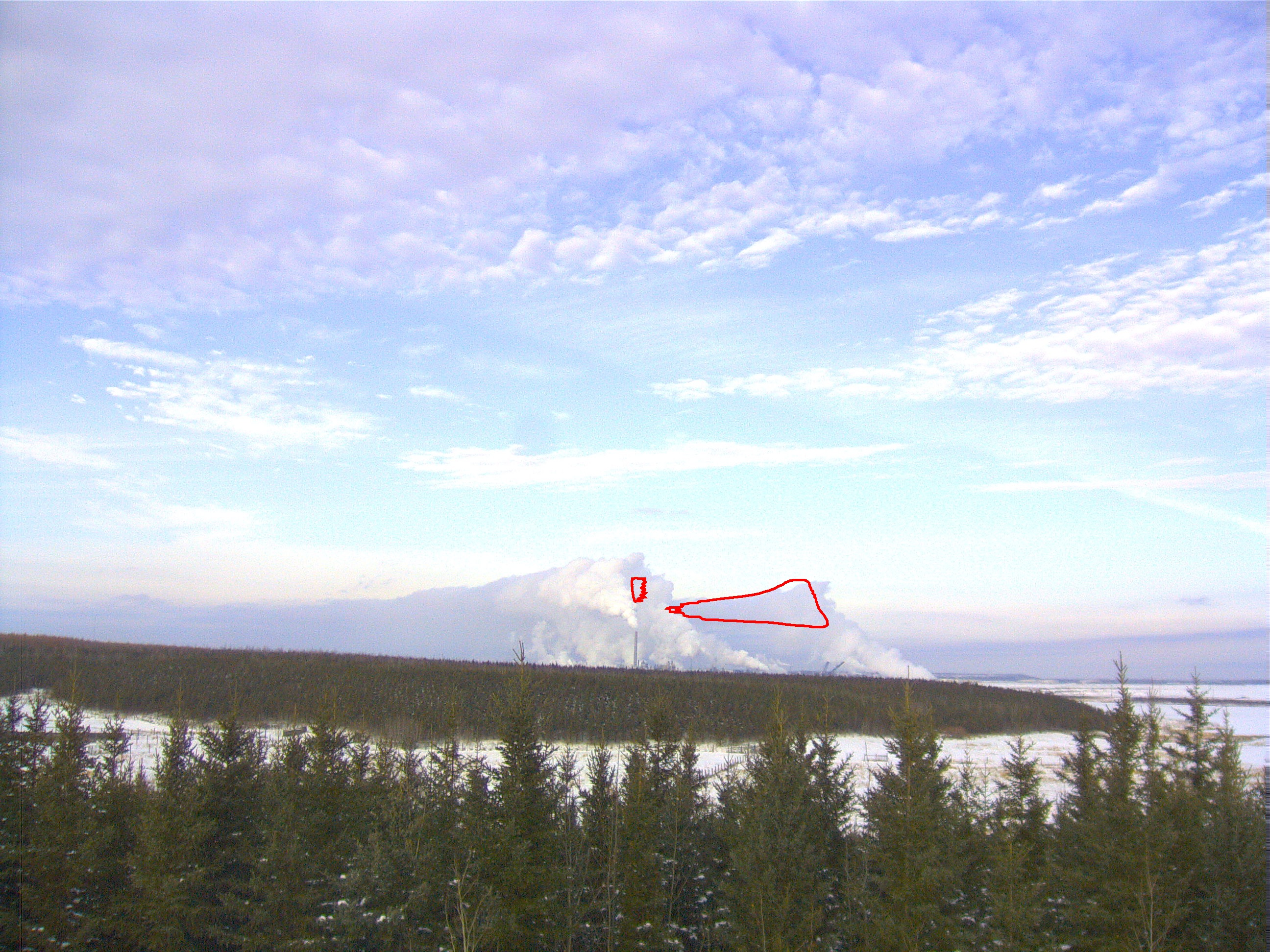}
    \includegraphics[width=\textwidth]
    {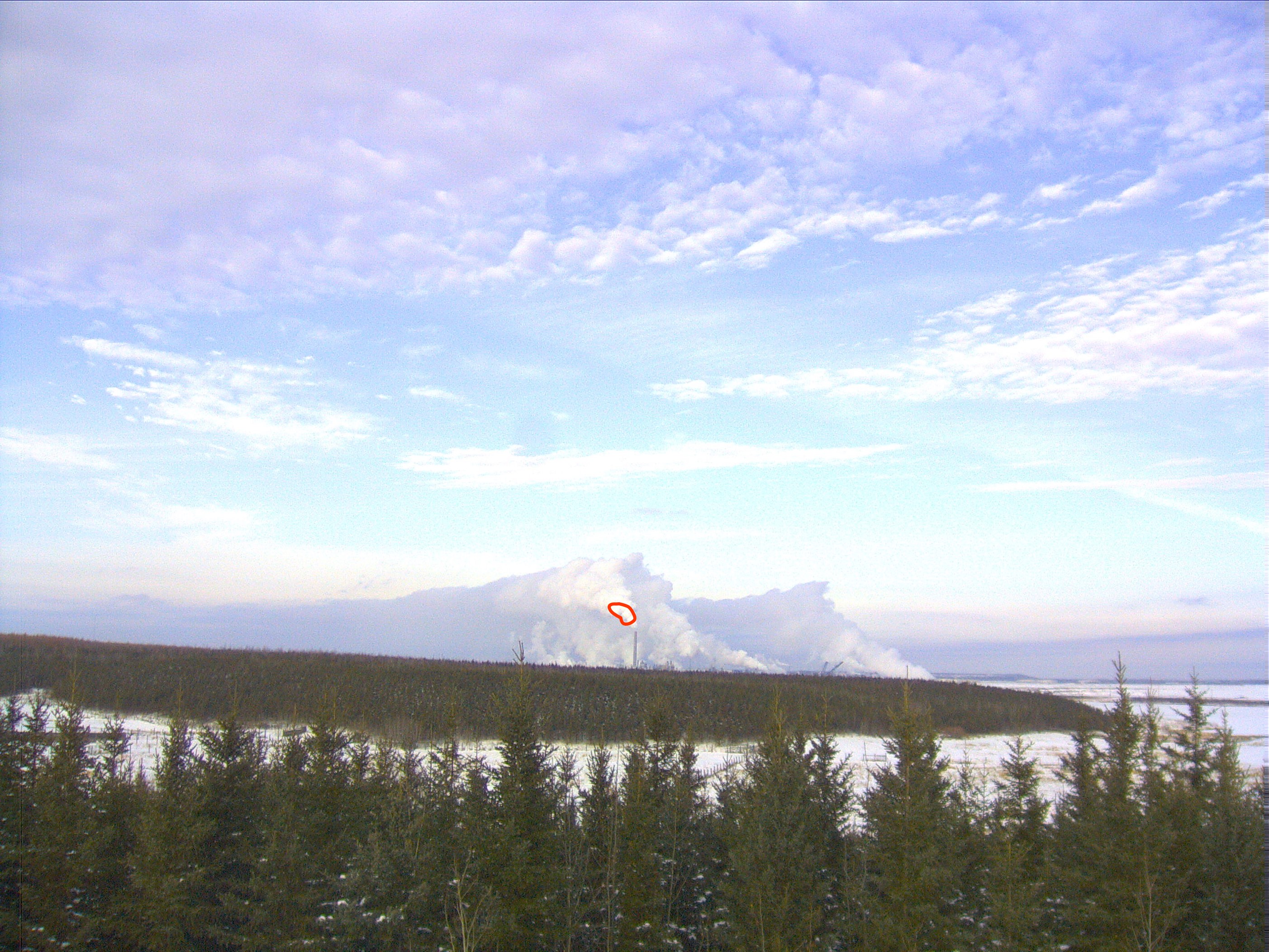}
    \includegraphics[width=\textwidth]
    {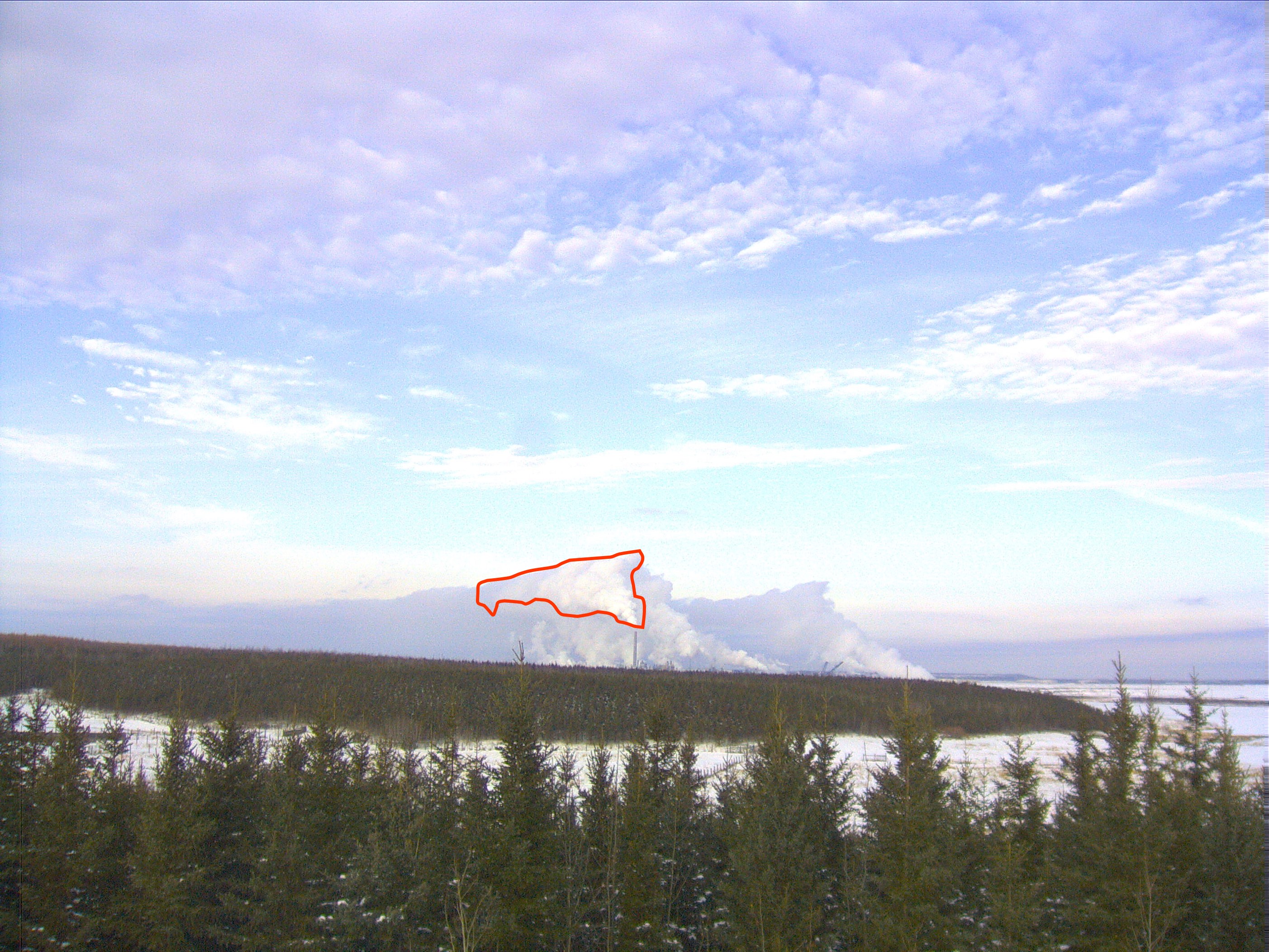}
    \includegraphics[width=\textwidth]
    {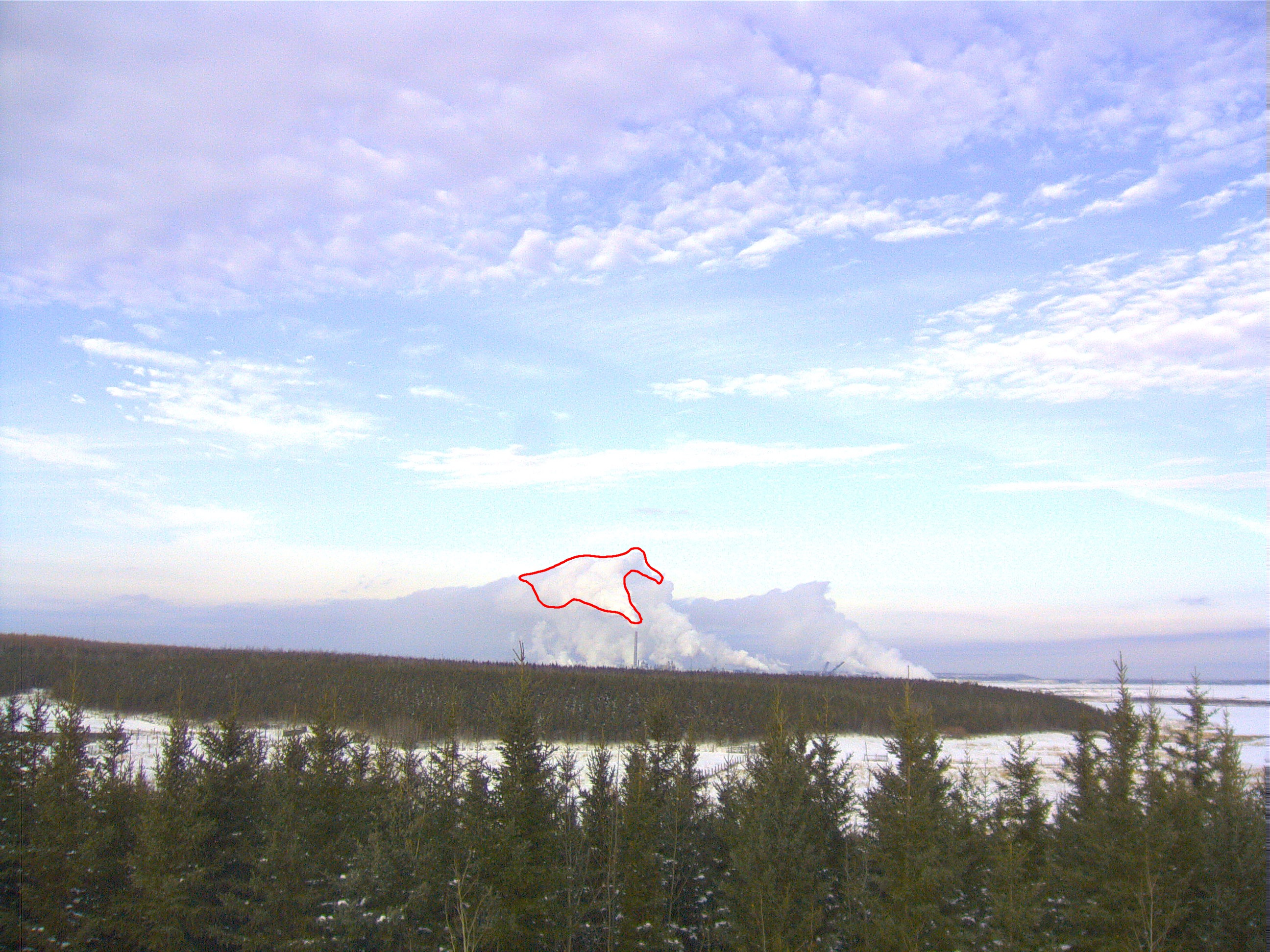}
\end{subfigure}\hfill
\begin{subfigure}[t]{0.15\textwidth}
    \includegraphics[width=\textwidth]
    {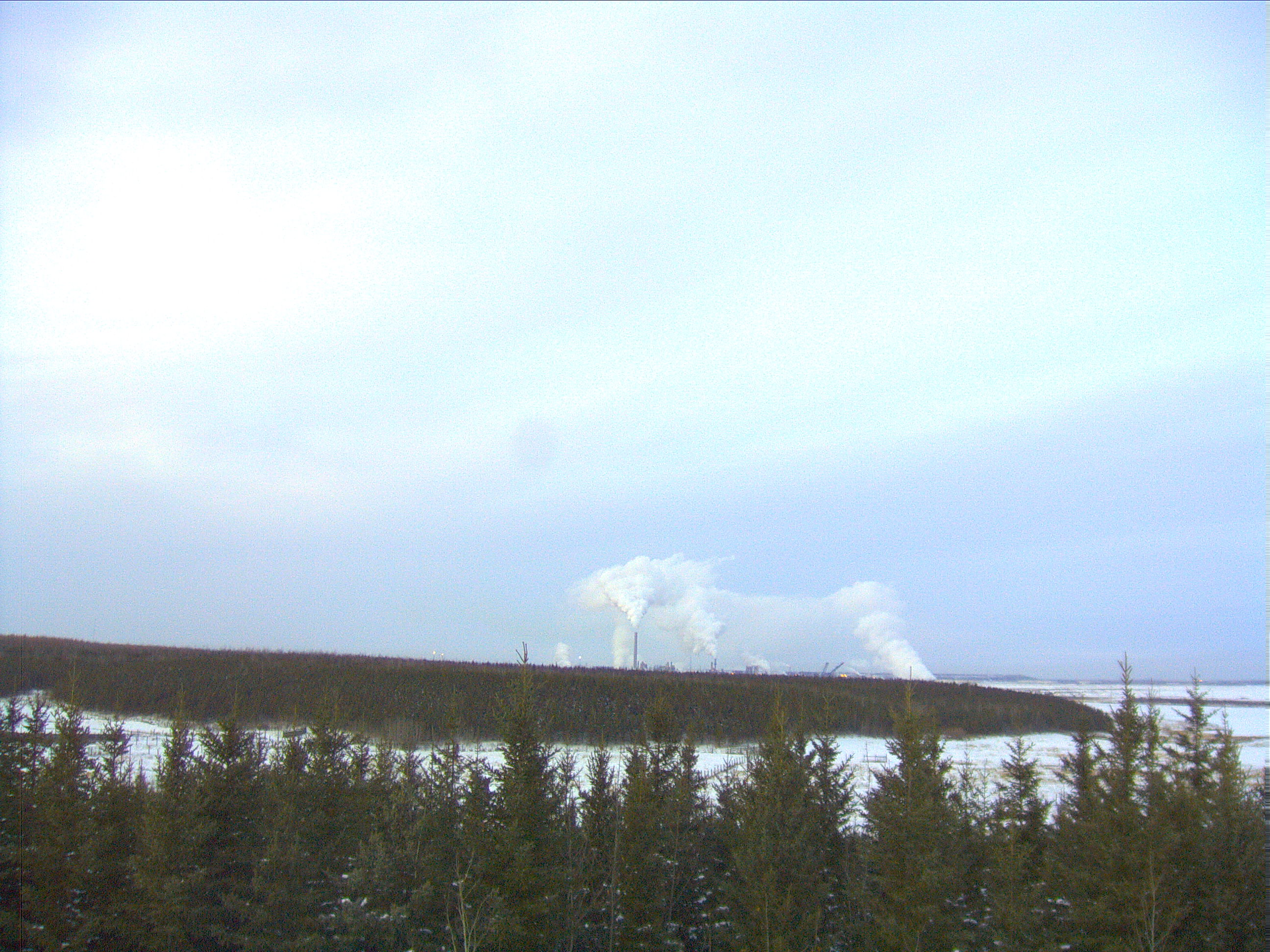}
    \includegraphics[width=\textwidth]
    {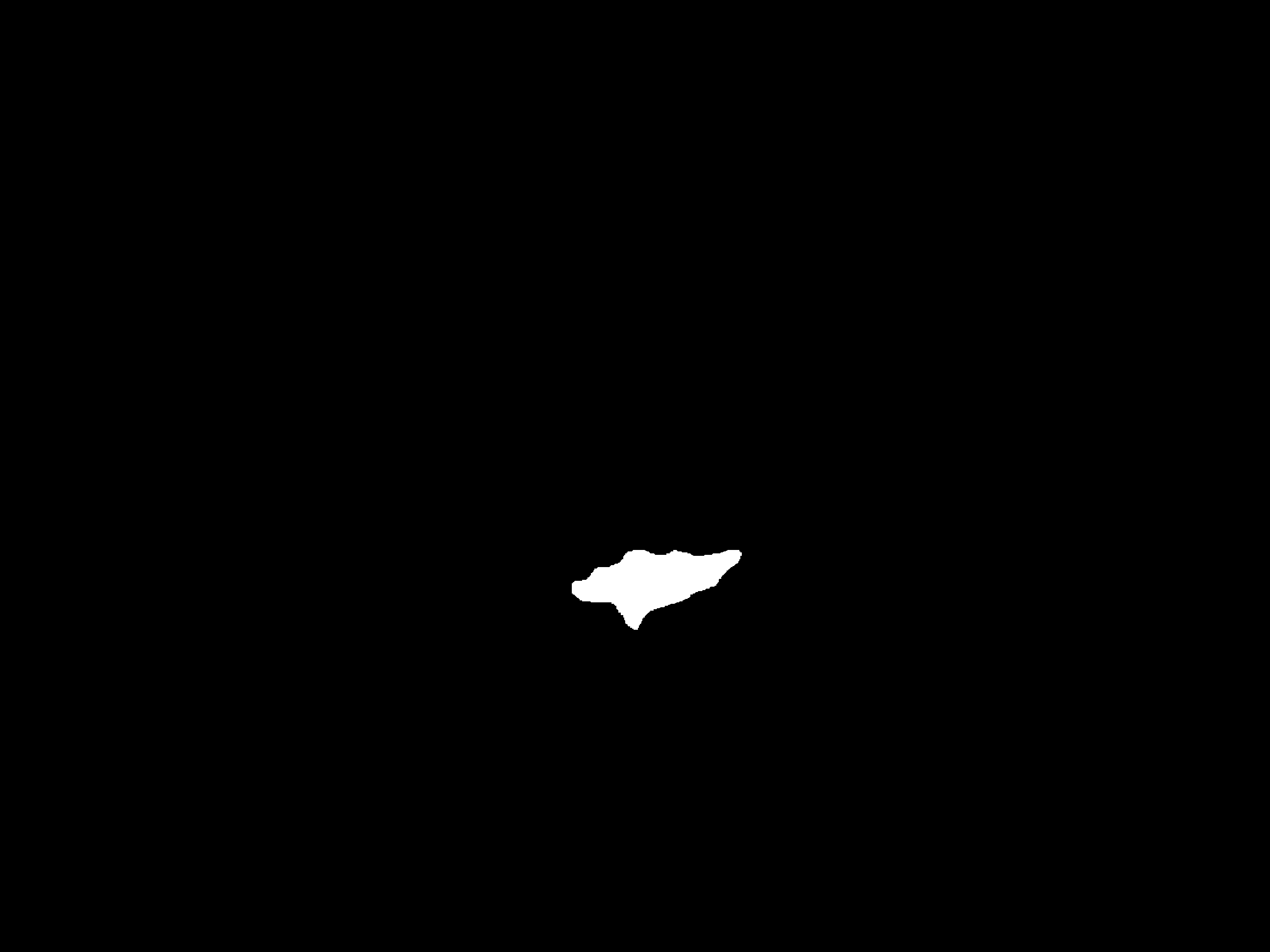}
    \includegraphics[width=\textwidth]
    {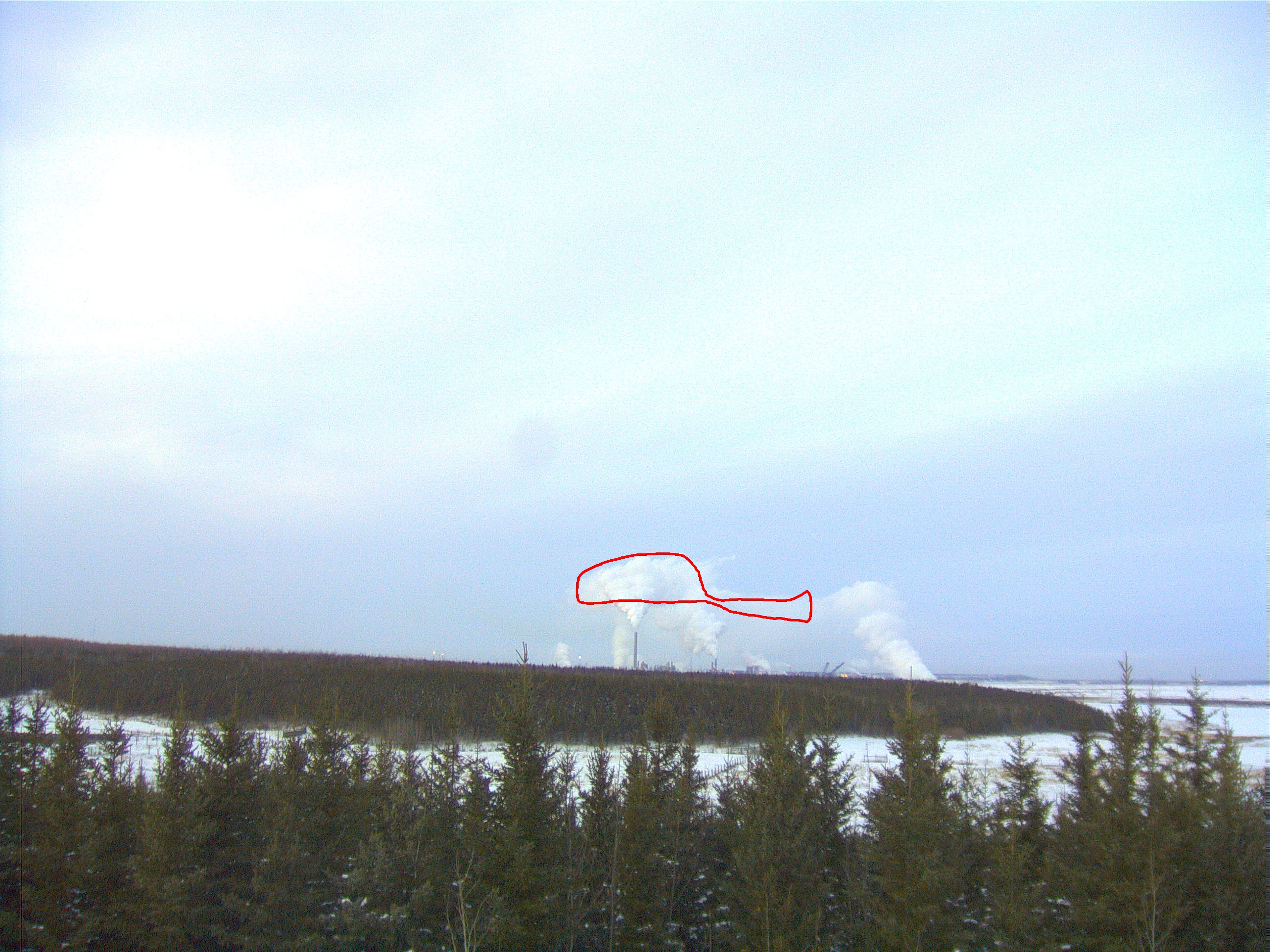}
    \includegraphics[width=\textwidth]
    {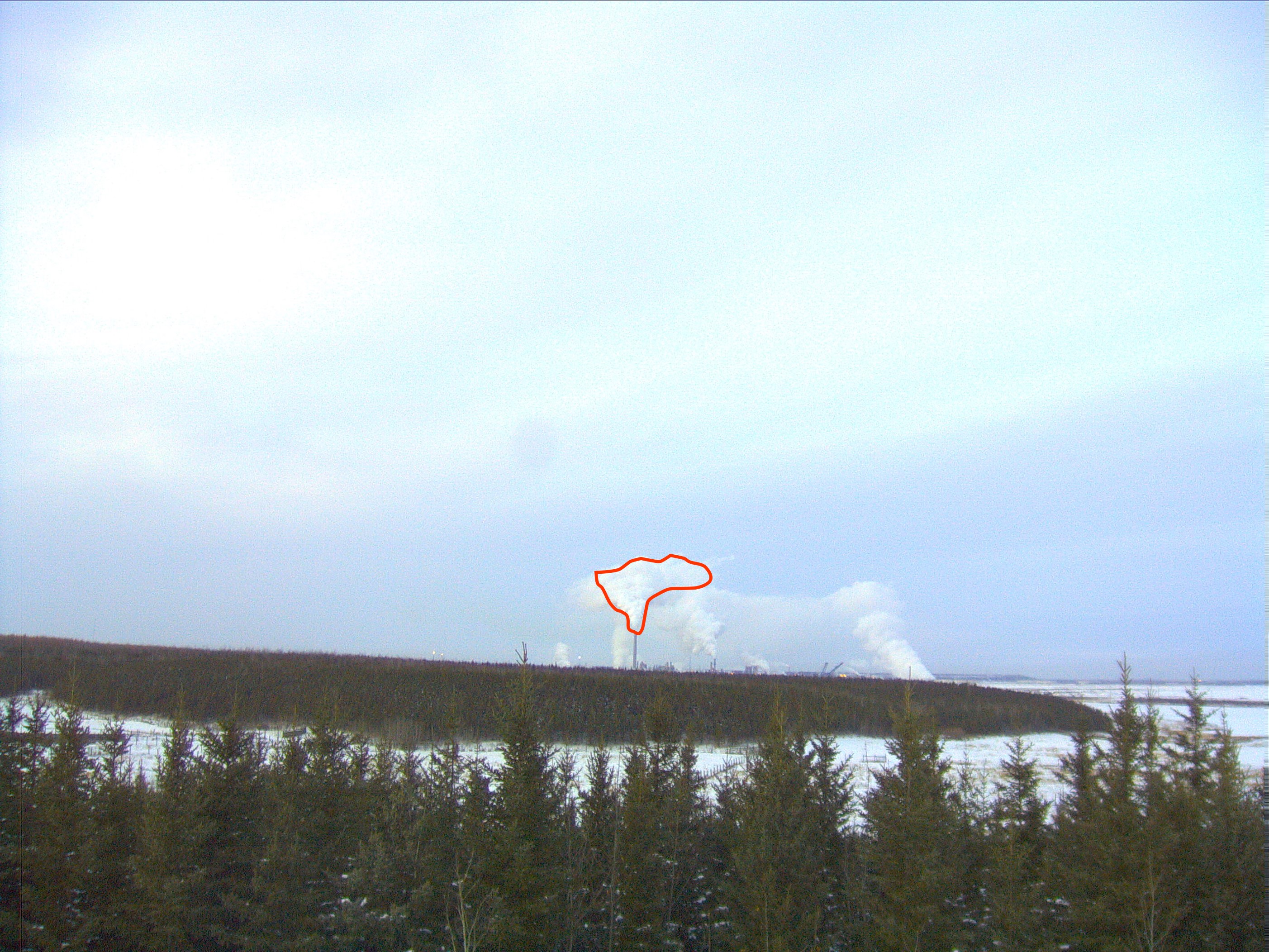}
    \includegraphics[width=\textwidth]
    {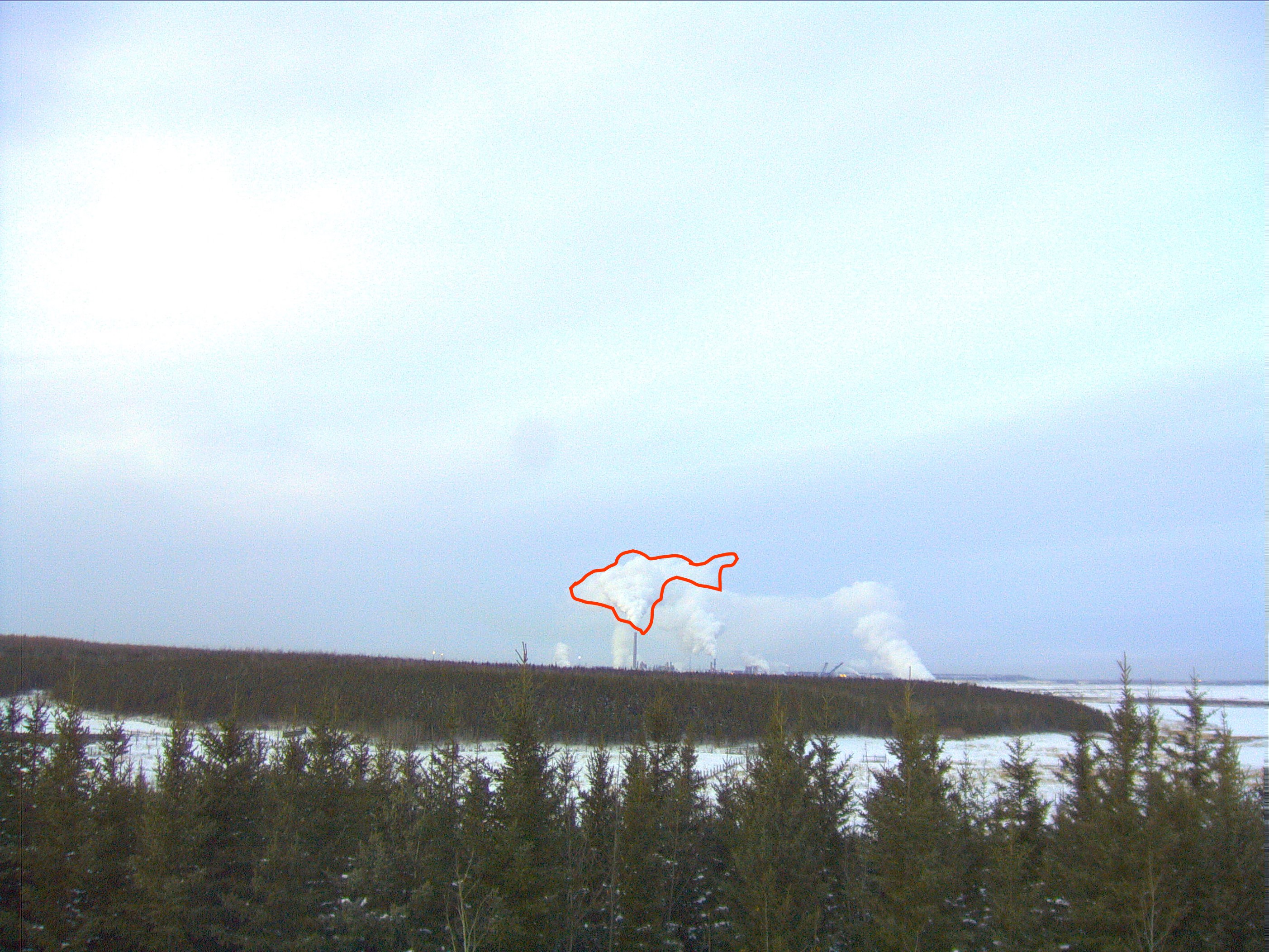}
    \includegraphics[width=\textwidth]
    {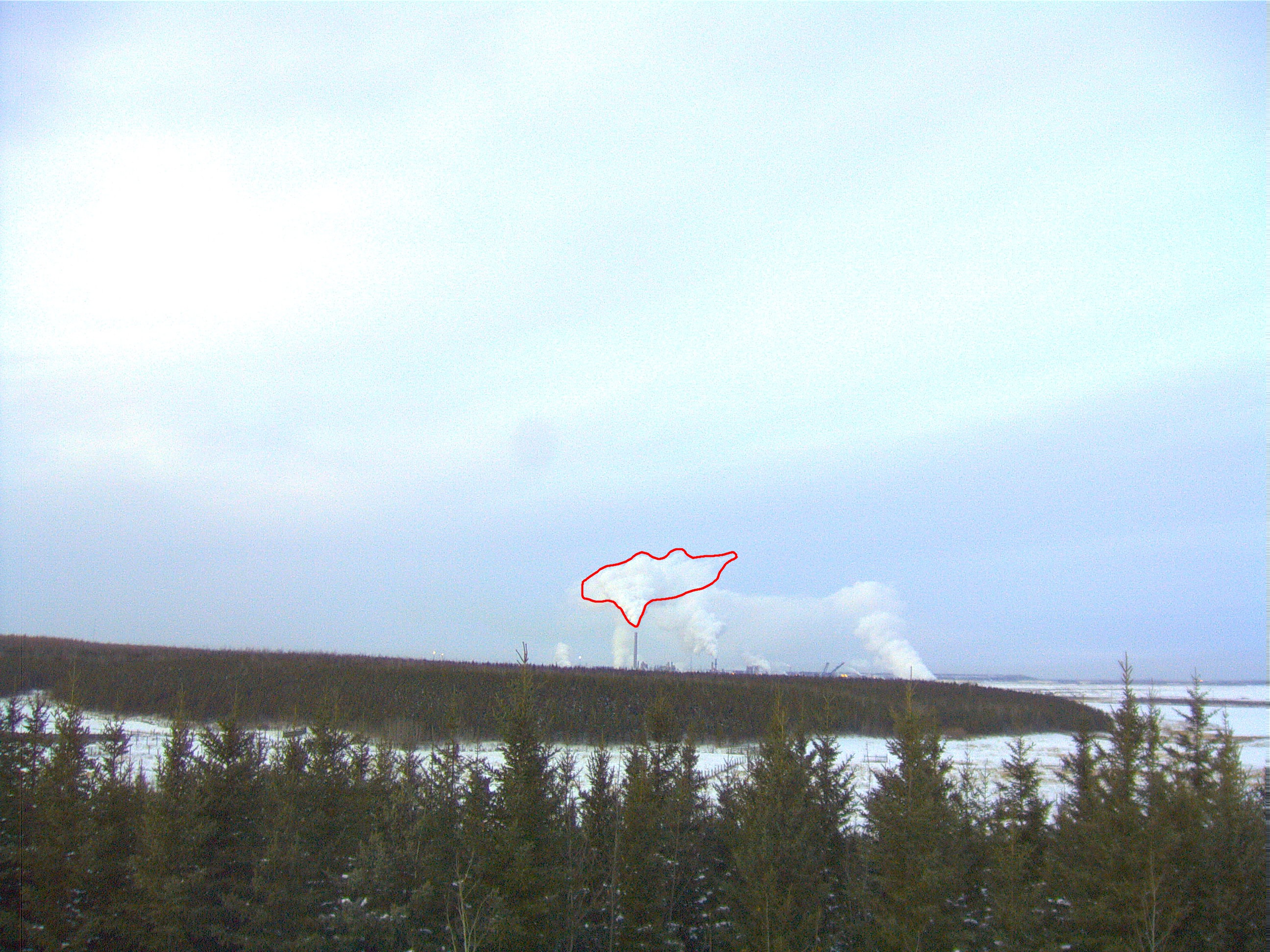}
\end{subfigure}\hfill
\caption{Qualitative results of recognition tasks listed as: (\textbf{a}) Input image, (\textbf{b}) corresponding ground truth, (\textbf{c}) results of Mask R-CNN, (\textbf{d}) FCN, (\textbf{e}) results of DeepLabv3, and (\textbf{f}) results of DPRNet.}
\label{fig.RRResults}
\end{figure}

To demonstrate the qualitative results of the proposed method, we show some visual results to compare competitive methods. Figure \ref{fig.RRResults} depicts these recognition results. The first two rows represent the input images and their corresponding ground truths, respectively, and the other rows give the output of different models. We tried to visualize samples from all classes such that the first two images are related to cloudy/foggy conditions, the second two are from the nighttime dataset, and the last two are obtained from our daytime dataset. It is observed that DPRNet outperformed the other methods by attaining high accuracy of PC localization and, consequently, correctly recognizing the desired smokestack PC.

\subsection{Plume rise measurement}\label{subsection.Plume rise calculation}
\begin{figure}[H]
  \begin{subfigure}{0.23\textwidth}
    \centering
    \includegraphics[width=3 cm]{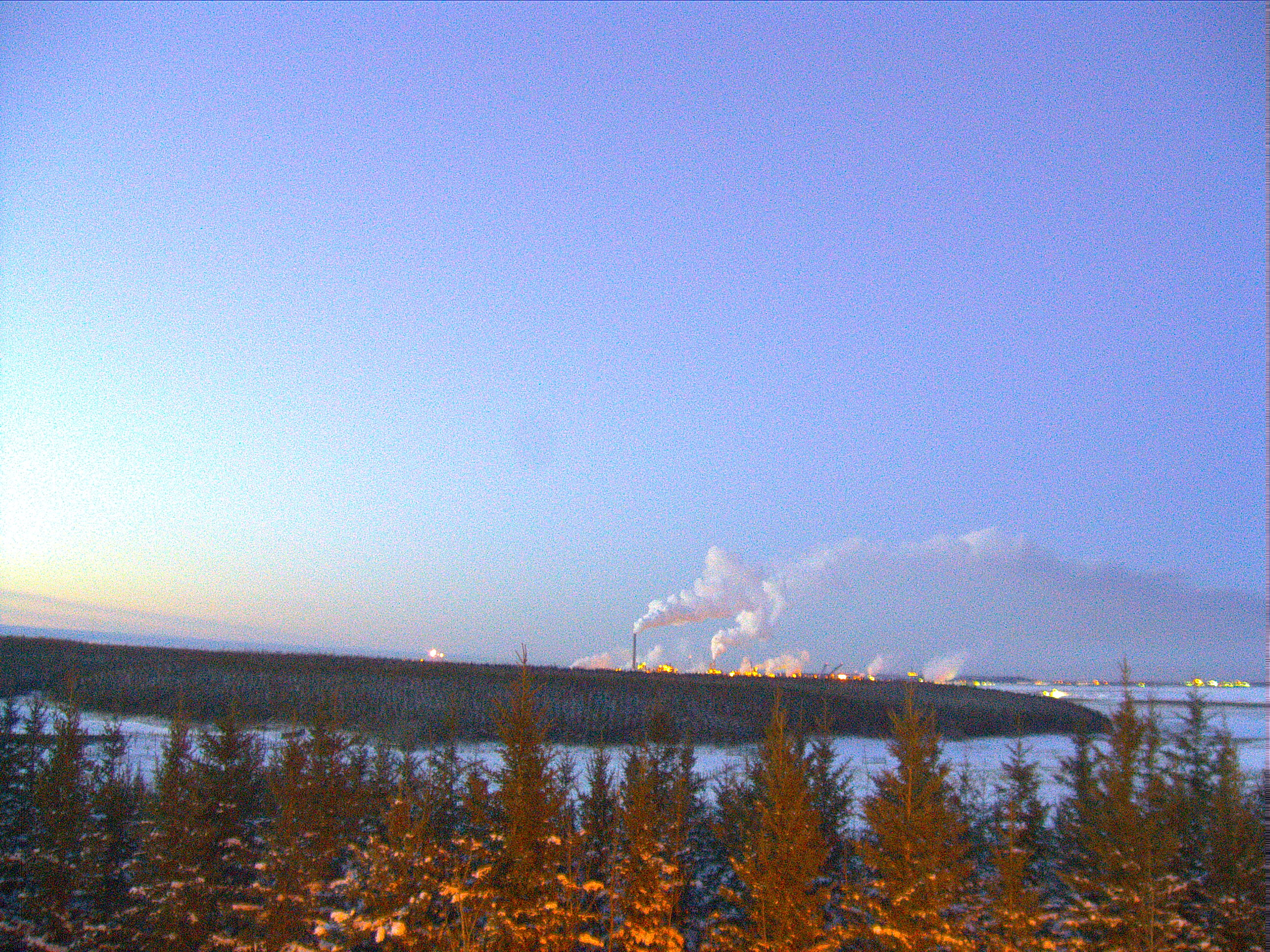}
  \end{subfigure}
  \begin{subfigure}{0.23\textwidth}
    \centering
    \includegraphics[width=3 cm]{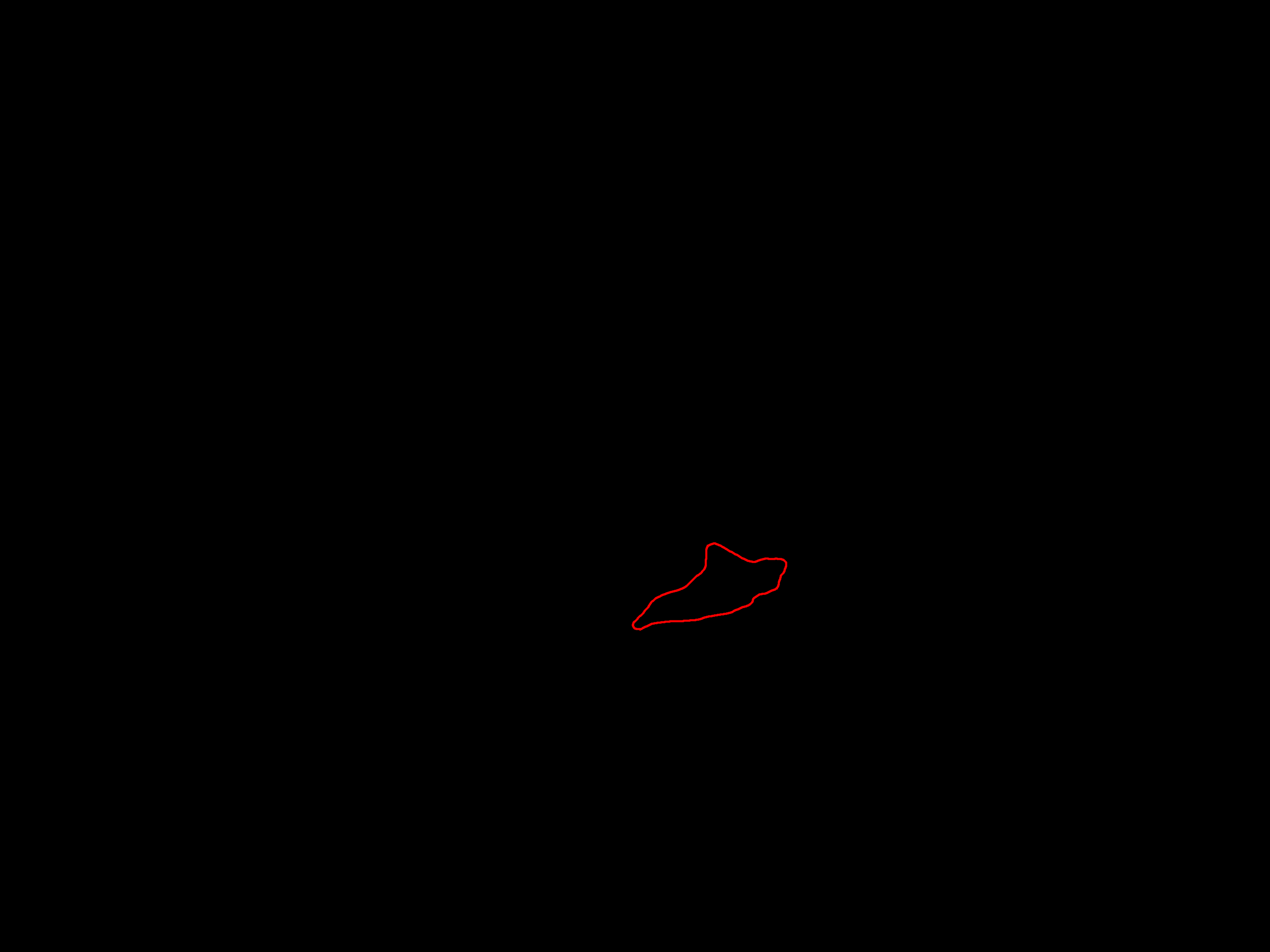}
  \end{subfigure}
    \begin{subfigure}{0.23\textwidth}
    \centering
    \includegraphics[width=3 cm]{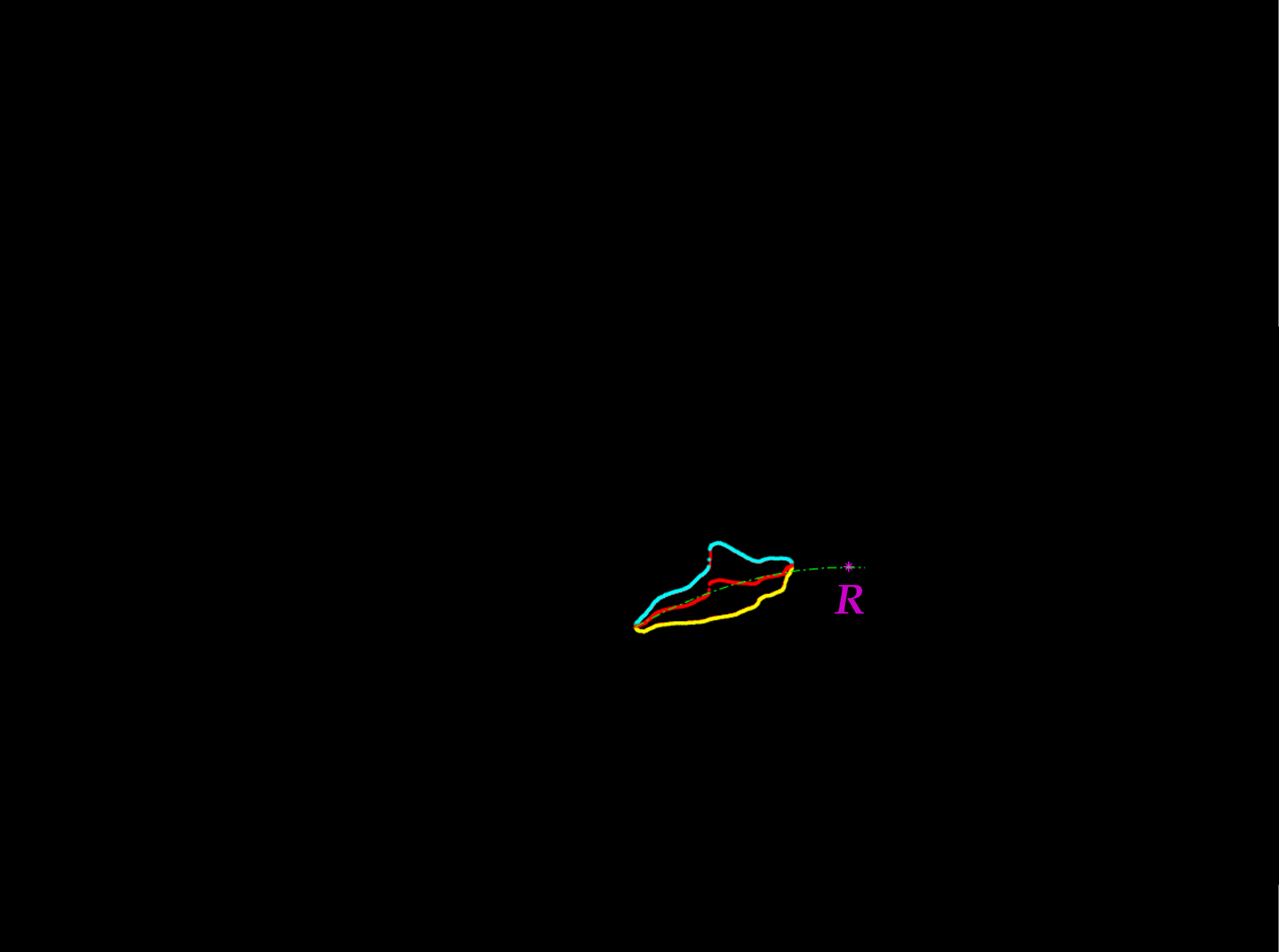}
  \end{subfigure}
  \newline
  \begin{subfigure}{0.23\textwidth}
    \centering
    \includegraphics[width=3 cm]{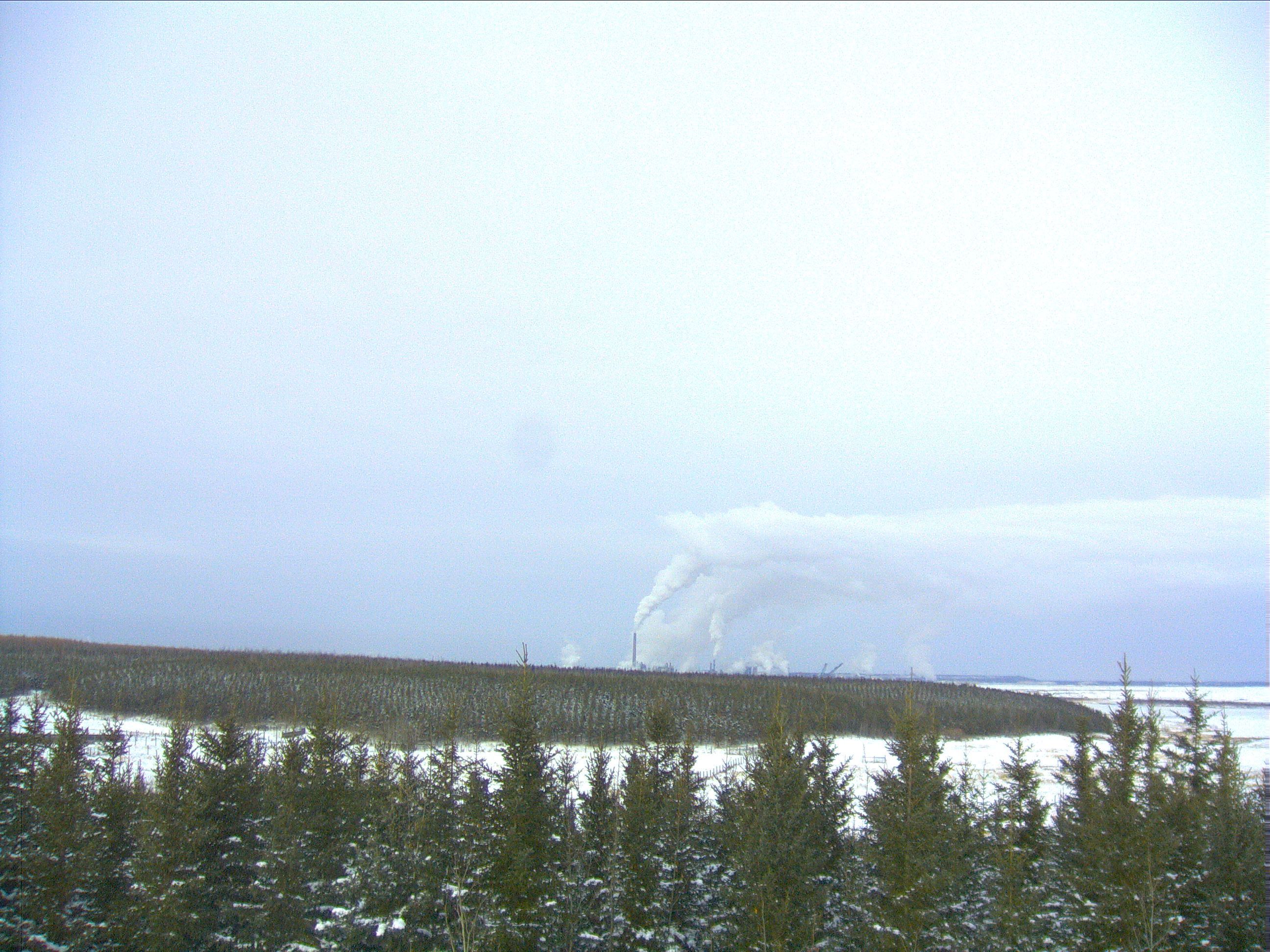}
  \end{subfigure}
  \begin{subfigure}{0.23\textwidth}
    \centering
    \includegraphics[width=3 cm]{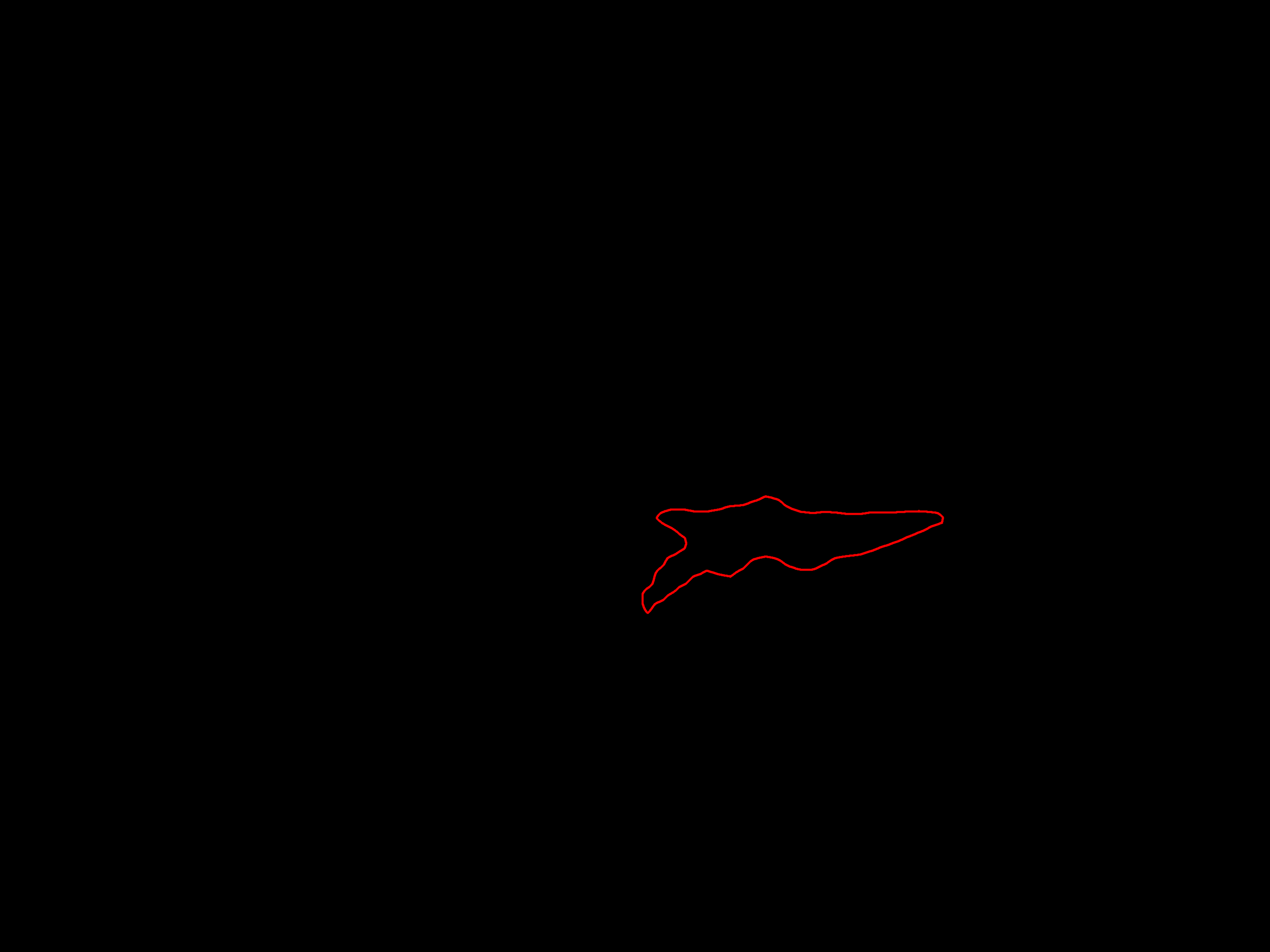}
  \end{subfigure}
    \begin{subfigure}{0.23\textwidth}
    \centering
    \includegraphics[width=3 cm]{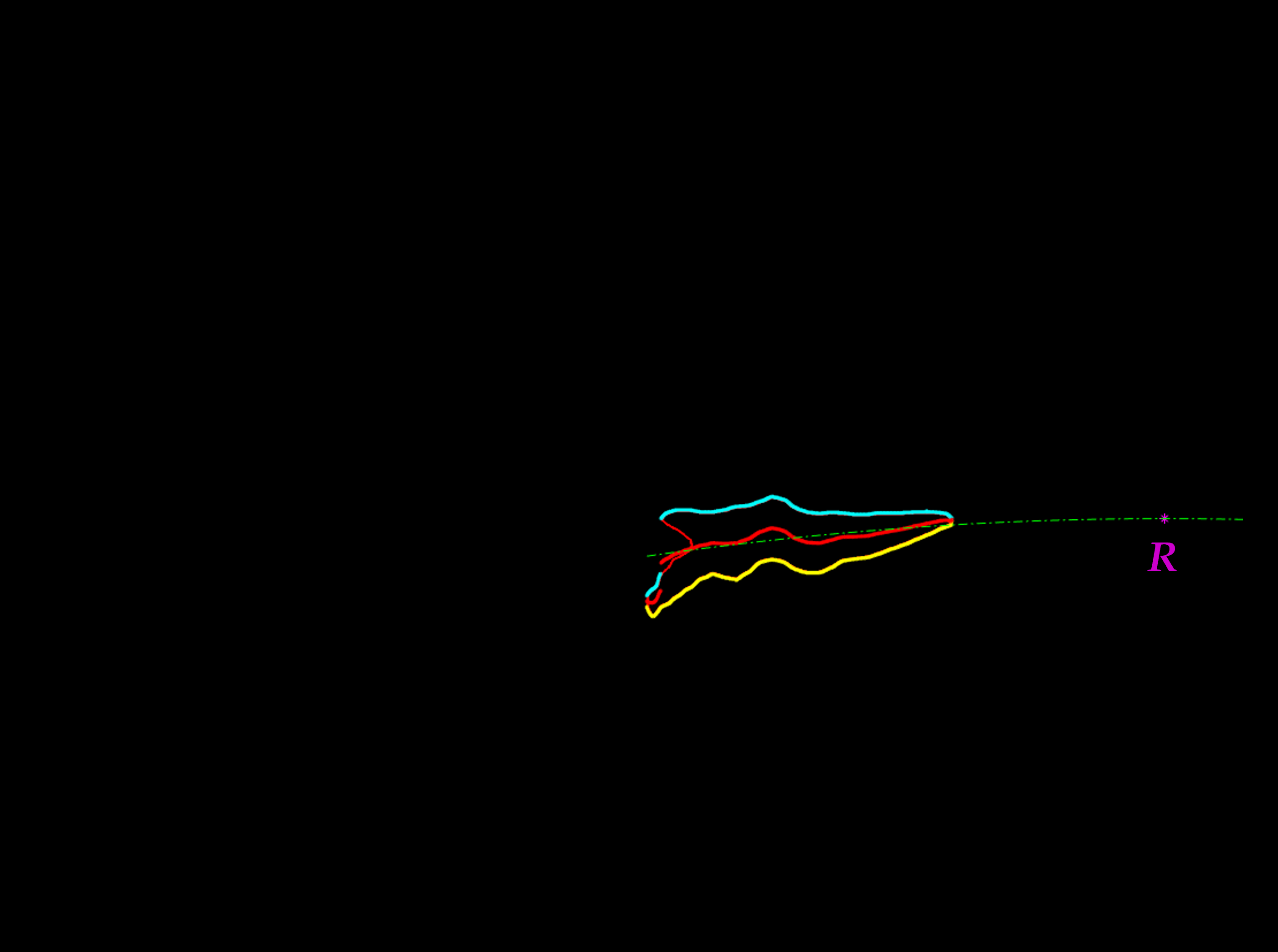}
  \end{subfigure}
  \newline
  \begin{subfigure}{0.23\textwidth}
    \centering
    \includegraphics[width=3 cm]{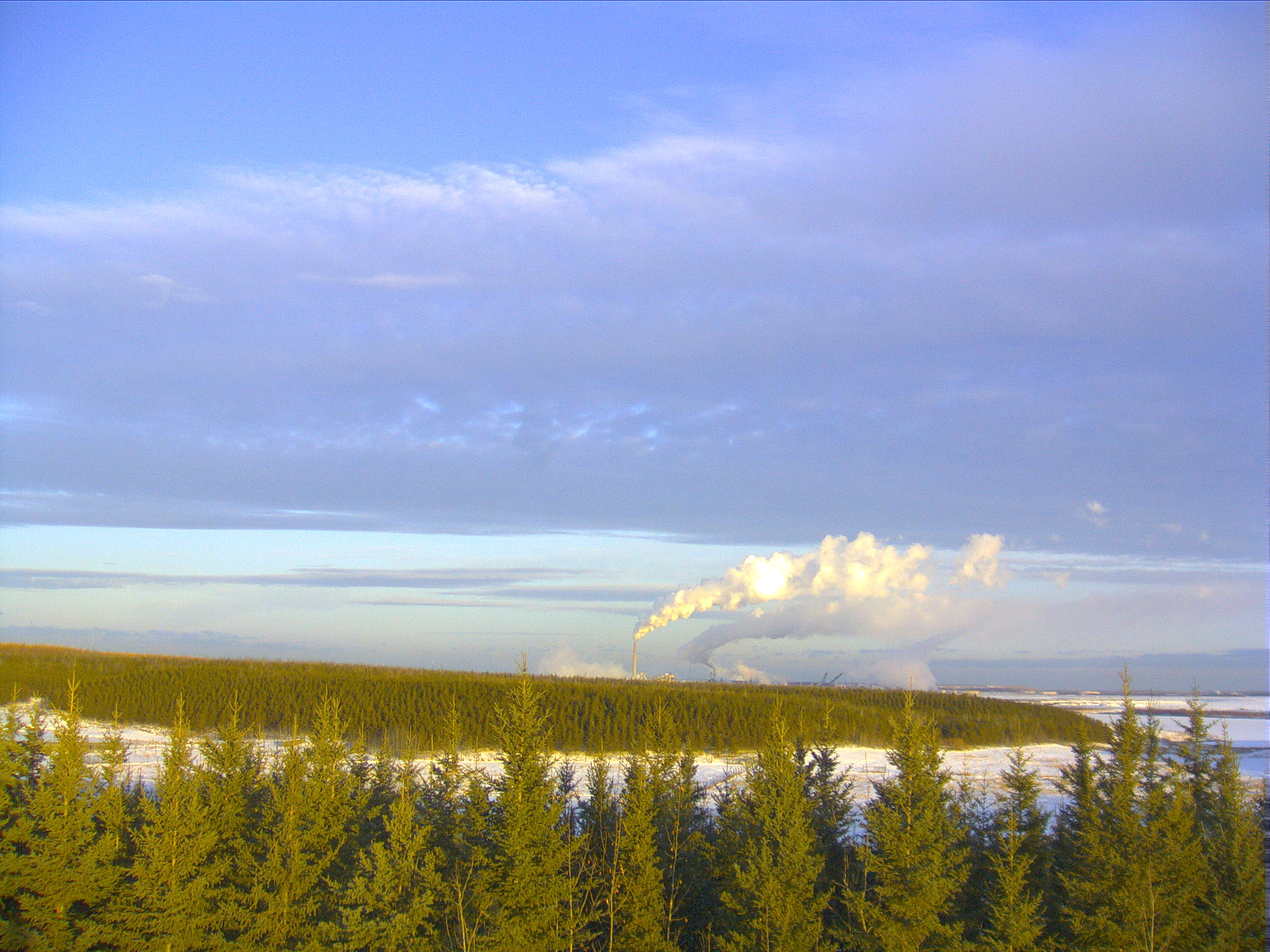}
  \end{subfigure}
  \begin{subfigure}{0.23\textwidth}
    \centering
    \includegraphics[width=3 cm]{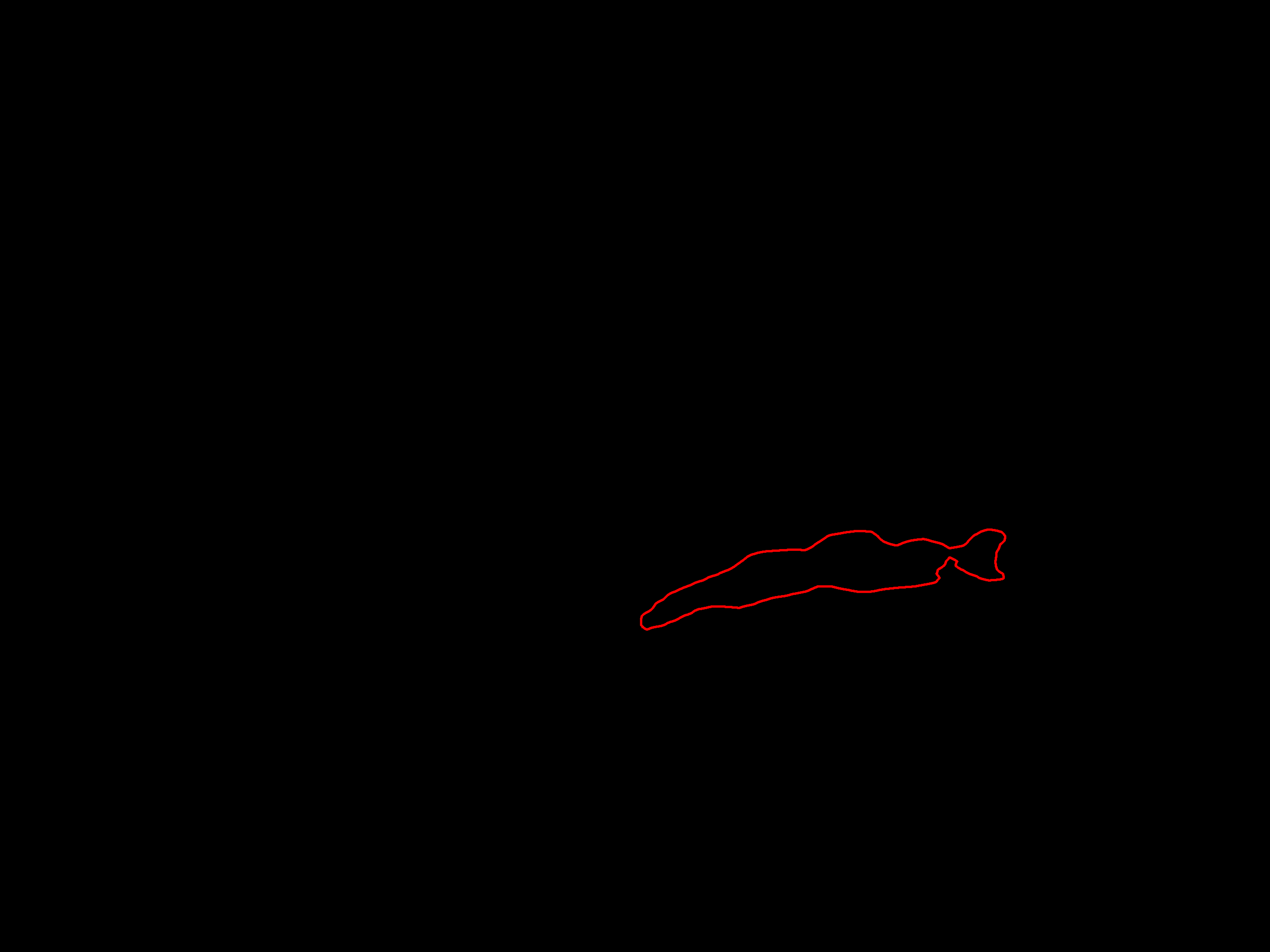}
  \end{subfigure}
    \begin{subfigure}{0.23\textwidth}
    \centering
    \includegraphics[width=3 cm]{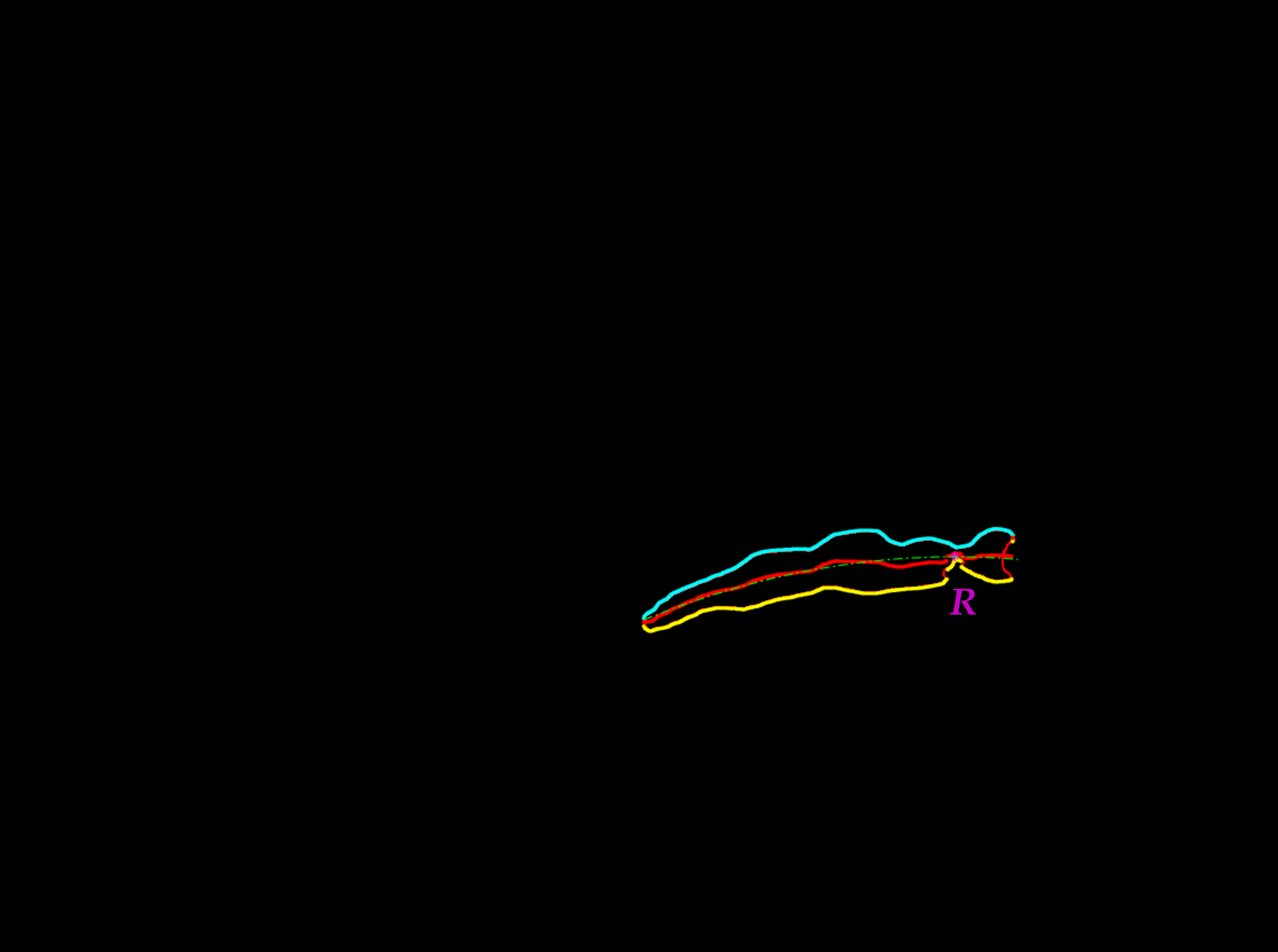}
  \end{subfigure}
  \newline
  \begin{subfigure}{0.23\textwidth}
    \centering
    \includegraphics[width=3 cm]{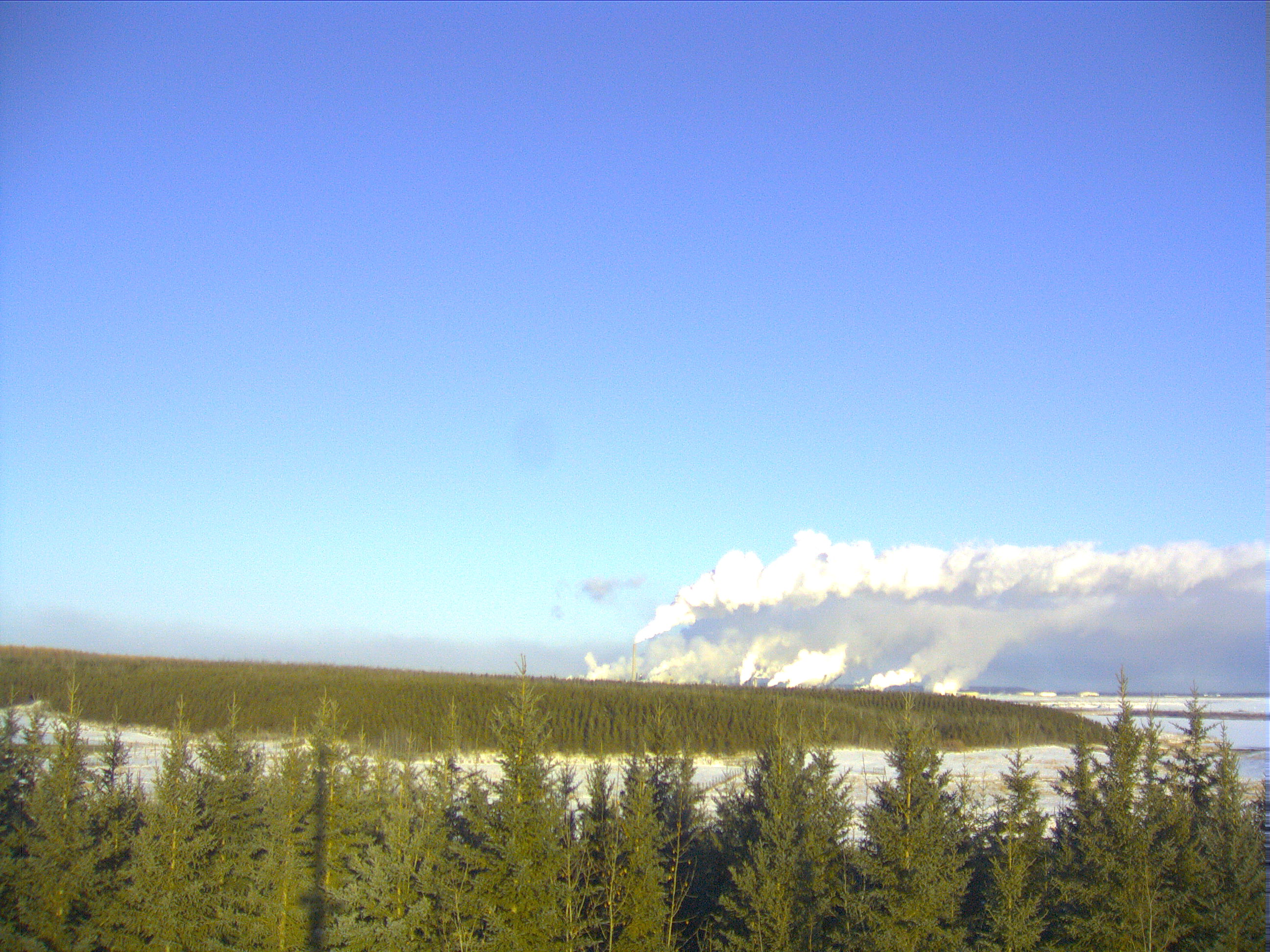}
    \caption{Input image}
    \label{subfig.Top view}
  \end{subfigure}
  \begin{subfigure}{0.23\textwidth}
    \centering
    \includegraphics[width=3 cm]{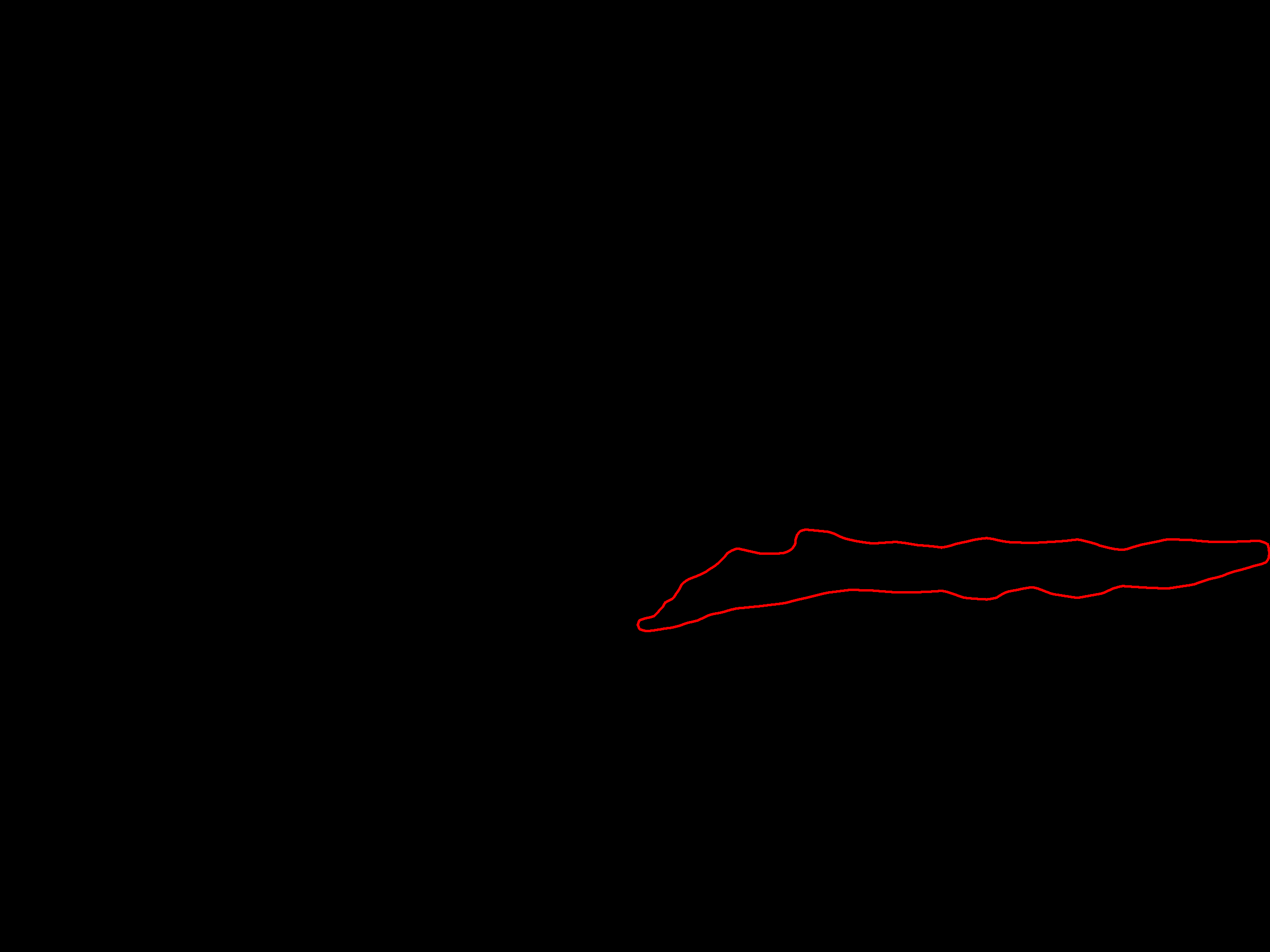}
    \caption{DPRNet result}
    \label{subfig.3D view}
  \end{subfigure}
    \begin{subfigure}{0.23\textwidth}
    \centering
    \includegraphics[width=3 cm]{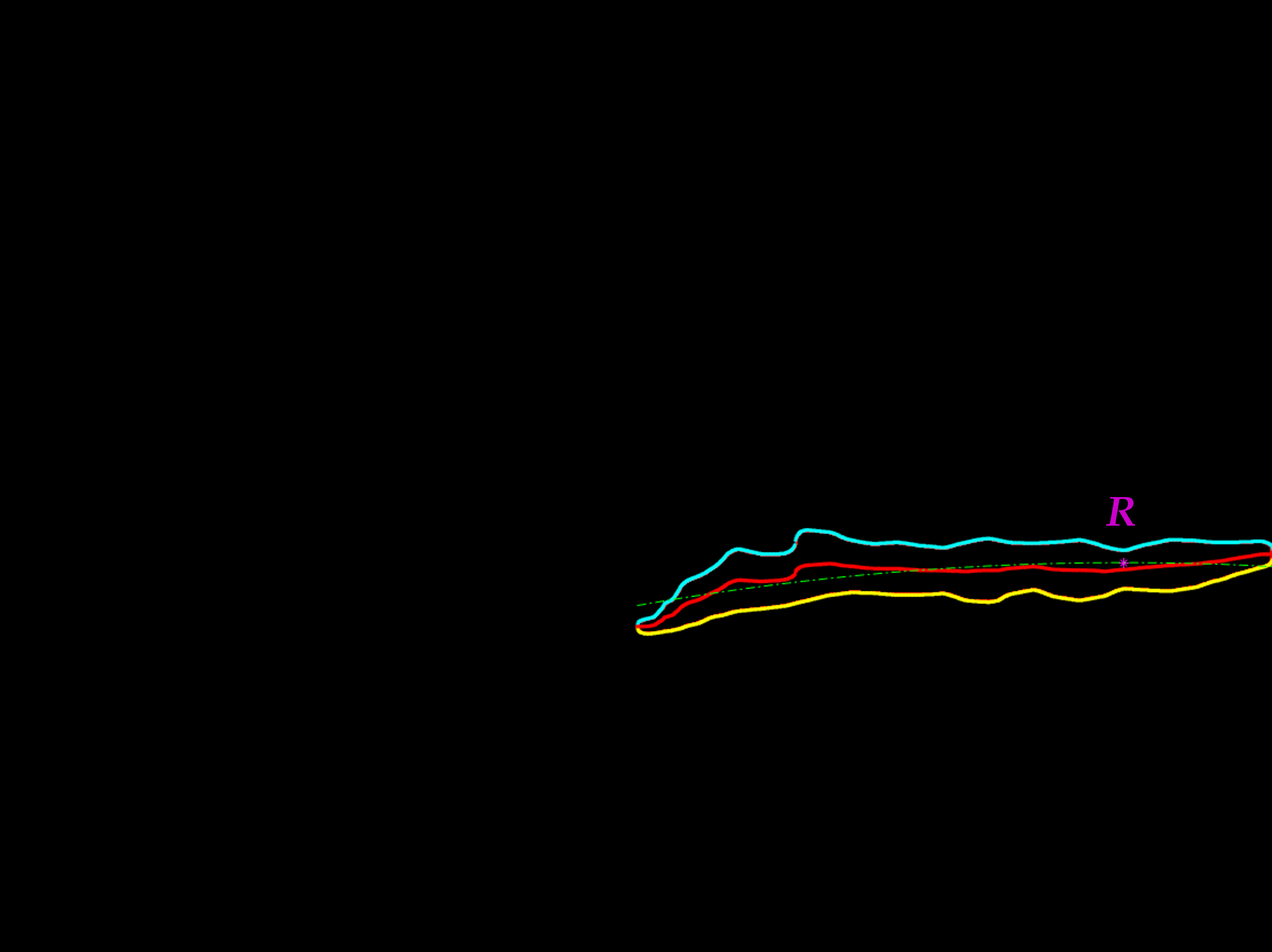}
    \caption{Image measurement}
    \label{subfig.Top view}
  \end{subfigure}
  \caption{DPRNet and image measurement results. In column (c), the red curve represents the meandering of the PC. The cyan and yellow lines, respectively, illustrate the upper and lower boundaries of the PC. Green dashes show the asymptotic curve, and the magenta point represents the point $R$.}
  \label{fig.Asymptotic curve of the plume cloud and selected point $R$}
\end{figure}

As we discussed in Section \ref{section.Methodology}, DPRNet gives PC border detection and recognition. Then, an asymptotic function is fitted to the smokestack PC centerline to measure PR using the critical point $R$ \cite{ge2023unsupervised}. Taking advantage of point $R$ image coordinate and the wind direction information from the meteorological tower, real-life PR measurement is obtained through geometric transformations. Figure \ref{fig.Asymptotic curve of the plume cloud and selected point $R$} illustrates the asymptotic curve for four plume cloud images and the automatically chosen point \emph{R} where the PC reaches neutral buoyancy. Also, the PR and PR distance values of each sample PC are given in Table \ref{table.1145}, as well as the averaged hourly measured wind directions at the image sampling times.

\begin{table}[H]
\caption{PR and PR distance values of each of four PC images and the averaged hourly measured wind directions based on the monitoring station information.\label{table.1145}}
\newcolumntype{C}{>{\centering\arraybackslash}X}
\begin{tabularx}{\textwidth}{CCCCCCC}
\toprule
\textbf{Image} & \textbf{Date (Y-M-D)} & \textbf{Time (H-M-S)} & $\boldsymbol{\varphi}$ \textbf{(deg.)} & $\boldsymbol{\theta}$ \textbf{(deg.)} & $\boldsymbol{\Delta{z}}$ \textbf{(m)} & $\boldsymbol{X_{max}}$ \textbf{(m)}\\
\midrule
I1 & 2019-11-08 & 18-00-13 & 12.16 & -239.8 & 177 & 1685 \\
I2 & 2019-11-09 & 15-00-13 & 3.46 & -248.5 & 450.3 & 3287 \\
I3 & 2019-11-14 & 10-00-16 & 10.41 & -241.6 & 266.8 & 2280 \\
I4 & 2019-11-16 & 11-00-12 & 10.83 & -241.1 & 300.5 & 2905 \\
\bottomrule
\end{tabularx}
\end{table}

\unskip

\section{Conclusion}\label{section.Conclusion}
To measure the PR through remote sensing images, PC border detection and recognition is of the essence as the first step. In this regard, a novel deep learning-based method, inspired by the nature of the problem, is proposed in this paper to detect and recognize the PC accurately. In the next stage, image processing analysis is leveraged to extract the PC centerline. Afterward, the critical point of this curve is estimated, the y-component coordinate of which is equivalent to PR. Lastly, this image measurement is transformed into real-life world under the geometric transformation stage. Experimental results indicate that the proposed method significantly outperformed its rivals. The proposed method face difficulty in the scene where there are several smokestacks. Our future studies focus on multi-source PCs, which frequently occur in industrial environments.

\section{Acknowledgements}\label{section.Acknowledgements}
We want to acknowledge Wood Buffalo Environmental Association (WBEA) for assistance with the camera installation and maintenance at the air-quality monitoring site in the Syncrude facility in northern Alberta, Canada. The project is funded by the "Lassonde School of Engineering Strategic Research Priority Plan" and "Lassonde School of Engineering Innovation Fund," York University, Canada, and "Natural Sciences and Engineering Research Council of Canada – NSERC (grant no. RGPIN 2015-04292)."

%%%%%%%%%%%%%%%%%%%%%%%%%%%%%%%%%%%%%%%%%%
\begin{adjustwidth}{-\extralength}{0cm}

\reftitle{References}

\end{adjustwidth}
\end{document}